\documentclass[]{spie}  

\usepackage{graphicx}
\usepackage{placeins}   

\usepackage{amsmath,amsfonts,amssymb}
\usepackage{graphicx}
\usepackage[colorlinks=true, allcolors=blue]{hyperref}
\renewcommand{\normalsize}{\large}
\title{EAGT: Echocardiography Augmentation for Generalisability and Transferability}

\author[a]{Soroush Elyasi}
\author[a]{Sara Adibzadeh}
\author[a]{Nasim Dadashi Serej*}
\author[a]{Massoud Zolgharni}

\affil[a]{THRIVE Centre, University of West London, London, United Kingdom}

\usepackage{longtable}
\usepackage{array}
\usepackage{xcolor}
\usepackage{float}
\usepackage{subcaption}
\usepackage{geometry}
\usepackage{afterpage}
\usepackage{caption}
\newgeometry{left=1cm,right=1cm,top=1.5cm,bottom=2.5cm}

\begin{document} 

\pagenumbering{arabic}

\setcounter{topnumber}{100}
\setcounter{bottomnumber}{100}
\setcounter{totalnumber}{100}

\renewcommand{\topfraction}{0.95}
\renewcommand{\bottomfraction}{0.95}
\renewcommand{\textfraction}{0.05}
\renewcommand{\floatpagefraction}{0.9}

\maketitle

\begin{abstract}
\begin{center}
\begin{minipage}{0.85\textwidth}
Deep learning models for echocardiography segmentation often struggle to generalise across institutions, scanners, and patient populations, where collecting large, consistently annotated datasets is infeasible. Data augmentation is inexpensive and widely used to improve the robustness of deep learning models; however, its role in enhancing cross-dataset generalisability in echocardiography remains insufficiently understood. This study presents a large-scale multi-dataset evaluation of 29 data augmentation techniques and their pairwise combinations for 2D left ventricular segmentation using a U-Net trained on Unity, CAMUS, and EchoNet Dynamic datasets. Each augmentation was explored under several hyperparameter settings and assessed through repeated runs using Dice and IoU in both in-domain and cross-dataset scenarios, with statistical significance quantified via independent t-tests. In-domain accuracy was near-saturated and insensitive to augmentation, whereas cross-dataset performance varied widely. Geometry-based augmentations including affine, shift-scale-rotate, flip, and perspective produced the largest and most consistent gains, while aggressive intensity- and artefact-based transforms often degraded transfer. Moreover, pairwise combinations outperformed individual augmentations mainly when the two transformations were complementary, particularly by improving some difficult domain-shift cases from poor to acceptable performance. These findings provide empirical guidance for designing augmentation policies that improve the robustness and transferability of echocardiography segmentation models.
\end{minipage}
\end{center}
\end{abstract}

\begin{minipage}{0.90\textwidth}
\end{minipage}
\section{INTRODUCTION}
\label{sec:intro}  

Data augmentation is a widely adopted strategy for improving model performance and enhancing generalisability in deep learning frameworks \cite{islam_systematic_2024,wodzinski_improving_2024,wang_comprehensive_2026,tupper_revisiting_2025}. In computer vision applications, augmentation is typically achieved by applying geometric transformations, intensity modifications, or noise augmentations to images in order to increase task difficulty and expose the model to variations that are not present in the original training set \cite{kumar_image_2024,alomar_data_2023}. Despite its effectiveness, selecting an appropriate subset from a large pool of augmentation techniques, often accompanied by extensive hyperparameter tuning, remains a non-trivial challenge \cite{kumar_image_2024}. This difficulty is further amplified by the fact that the optimal augmentation strategy is highly domain-dependent and varies significantly across image types, such as natural images, remote sensing imagery, civil engineering datasets, and medical images \cite{tupper_revisiting_2025,goceri_medical_2023}. Consequently, defining a universally transferable augmentation policy is impractical, and augmentation is often underexplored or overlooked in specialised domains, particularly in medical imaging and echocardiography \cite{tupper_revisiting_2025}.

An additional factor that substantially complicates augmentation design is the fundamental disparity between natural images and medical images, which precludes the direct transfer of strategies across domains \cite{mazurowski_segment_2023,ma_segment_2024}. Natural images commonly feature RGB colour channels, high spatial resolution, and relatively clear object boundaries. In contrast, medical images are often acquired using modality-specific physical mechanisms and may exhibit greyscale intensity distributions, low contrast, noise, artefacts, and blurred anatomical boundaries. Differences in colour representation, structural appearance, edge sharpness, and noise characteristics are therefore particularly pronounced. As a result, augmentation techniques developed for natural image datasets cannot be directly adopted in medical imaging without careful domain-specific adaptation \cite{tupper_revisiting_2025,saeed_contrastive_2022,holste_efficient_2024,guo_improved_2023}.

These challenges extend across medical imaging modalities. Magnetic resonance imaging (MRI), computed tomography (CT), X-ray, and ultrasound imaging produce fundamentally different contrast mechanisms, spatial resolutions, and noise distributions, leading to modality-specific augmentation requirements \cite{varghese_enhancing_2024}. For example, CT and X-ray images are based on X-ray attenuation, MRI depends on magnetic resonance properties of tissues, and ultrasound relies on acoustic wave propagation and reflection. These differences mean that an augmentation strategy suitable for one modality may not preserve meaningful anatomical or diagnostic information in another. Such variability makes it impractical to define a single augmentation strategy applicable across all imaging modalities, anatomical targets, or clinical scenarios, thereby motivating the need for domain- and task-specific augmentation frameworks \cite{goceri_medical_2023,mazurowski_segment_2023,ma_segment_2024,varghese_enhancing_2024 }.

Within medical imaging, ultrasound presents additional challenges because image formation is highly affected by acoustic propagation, tissue-dependent attenuation, scattering, speckle, and shadowing. For example, proximity to bone structures can significantly increase acoustic reflections and acoustic shadowing, while signal attenuation varies across soft and dense tissues, resulting in spatially heterogeneous image quality \cite{tupper_revisiting_2025,alsinan_bone_2020}. These factors produce images that are often noisy, low in contrast, and strongly dependent on acquisition conditions. Therefore, augmentation strategies for ultrasound must consider modality-specific artefacts \cite{tupper_revisiting_2025,ramakers_ultraaugment_2024}.

Echocardiography is one of the most challenging applications of ultrasound imaging. Echocardiographic data are typically processed and analysed in greyscale and are inherently noisy, often exhibiting blurred cardiac structures and poorly defined edges \cite{moinuddin_medical_2022,boumeridja_enhancing_2025,meyer_ultrasam_2025}. In addition, image appearance can vary due to patient anatomy, probe position, acquisition protocol \cite{sfakianakis_gudu_2023,guo_impact_2024}, operator dependency, and scanner-specific processing. These characteristics make echocardiography substantially different from both natural images and other medical imaging modalities \cite{ramakers_ultraaugment_2024,sfakianakis_gudu_2023}. Consequently, augmentation policies for echocardiographic image analysis require careful design to ensure that transformations preserve clinically meaningful cardiac structures while improving model robustness and generalisability \cite{ramakers_ultraaugment_2024,sfakianakis_gudu_2023}.

In addition to these differences, data annotation is challenging in medical imaging, particularly in echocardiography \cite{tupper_revisiting_2025}. High-quality annotations require expert clinicians with extensive training, formal certification, and substantial clinical experience, making the annotation process both time-consuming and costly \cite{ramakers_ultraaugment_2024,moinuddin_medical_2022,mazurowski_segment_2023,alajrami_semi-supervised_nodate}. This results in a scarcity of labelled data and publicly available datasets when compared to domains such as natural image analysis \cite{ramakers_ultraaugment_2024}. Moreover, medical data are highly sensitive, and many patients are unwilling to have their information shared. Consequently, extensive anonymisation procedures and the acquisition of strict regulatory approvals are required, making data access both time-consuming and costly \cite{goceri_medical_2023}. The high cost and limited availability of labelled data have consequently slowed methodological progress in medical imaging compared to other computer vision fields, primarily due to the need for large and diverse datasets to achieve high-performance models \cite{mazurowski_segment_2023}.

To address these constraints, researchers have explored alternative learning paradigms, including semi-supervised and self-supervised learning approaches \cite{alajrami_semi-supervised_nodate,ferreira_self-supervised_2025,wang_self-supervised_2025,naidoo_consensus-guided_2025}. While these methods can partially reduce dependence on labelled data, some challenges remain, particularly for complex tasks such as segmentation \cite{liu_rethinking_2024,gui_survey_2024}. Moreover, many self-supervised techniques, such as contrastive learning and autoencoder-based methods, implicitly depend on data augmentation to construct meaningful training objectives \cite{tupper_revisiting_2025}. As a result, the effectiveness of these approaches is still strongly influenced by the choice of augmentation strategy \cite{saeed_contrastive_2022,holste_efficient_2024}.

In response, some studies have directly adopted augmentation policies originally designed for natural images, often without systematic validation for medical data \cite{saeed_contrastive_2022,holste_efficient_2024,guo_improved_2023,chartsias_contrastive_2021,lamoureux_segmenting_2023,kim_echofm_2025}. Others have proposed modality-specific solutions for MRI, X-ray, CT, or ultrasound \cite{vepa_integrating_2024,wang_mdal_2025,zhang_multi-condos_2024}. In echocardiography, annotations are commonly provided only for two key cardiac phases, which are end-diastolic (ED) and end-systolic (ES). Several works have therefore utilised intermediate frames within the cardiac cycle as an implicit form of augmentation or as auxiliary data for self-supervised representation learning \cite{holste_efficient_2024,kim_echofm_2025,kondori_echognn_nodate}. While promising, these approaches highlight the continued need for principled, domain-aware augmentation strategies tailored specifically to medical imaging tasks.

These domain discrepancies also limit the effectiveness of pre-trained model weights, such as those derived from large-scale medical image repositories, for example RadImageNet \cite{mei_radimagenet_2022}, when applied to certain medical imaging tasks through transfer learning. Both task-specific models and recent foundation models often experience substantial performance degradation when deployed on unseen datasets or across institutions. This issue is particularly critical in clinical settings, where models trained on carefully curated and costly annotated datasets may fail when transferred between hospitals with different imaging devices, acquisition protocols, and patient populations. Such performance drops undermine the practical utility of models trained under demanding annotation conditions \cite{ma_segment_2024,meyer_ultrasam_2025,eche_toward_2021}.

Although recent efforts, including few-shot learning, one-shot learning, and active learning, aim to mitigate this problem by reducing annotation requirements in the target domain, improving the intrinsic generalisability of trained models remains a more effective and scalable solution \cite{vepa_integrating_2024,alajrami_active_2024,qu_rethinking_2024}. Improving the generalisability of trained models can significantly reduce retraining effort and annotation costs when adapting them to new clinical environments.

While foundation models trained on millions of samples across diverse datasets represent a promising direction toward improved generalisation, their development requires substantial computational resources, extensive labelled and unlabelled data, and specialised hardware, making them impractical in many real-world research and clinical settings. Moreover, even these large-scale models may still exhibit performance degradation when faced with domain shifts, albeit to a lesser extent than task- or dataset-specific models \cite{ma_segment_2024,meyer_ultrasam_2025,kim_echofm_2025}. In contrast, data augmentation represents a long-established, computationally efficient, and domain-adaptable strategy for improving model robustness and generalisability. Despite its proven success in other computer vision domains, the systematic exploration of data augmentation in ultrasound and echocardiography remains limited \cite{tupper_revisiting_2025}. Figure 1 illustrates the overall design and workflow of the proposed research.

\begin{figure}[ht]
\centering
\includegraphics[width=\linewidth]{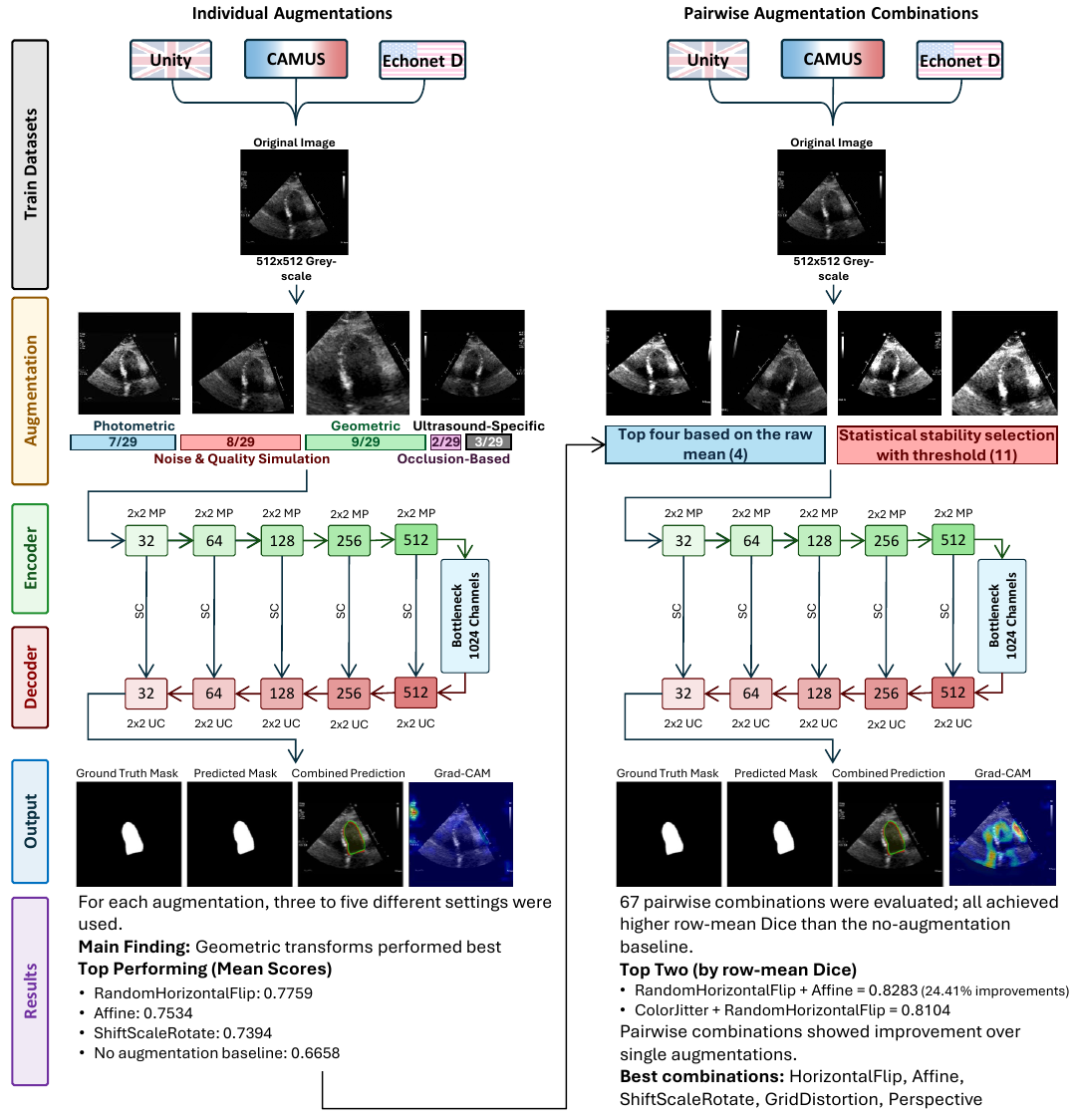}
\caption{
Overall framework illustration 
}
\label{fig:example}
\end{figure}

Although data augmentation is widely used in echocardiographic deep learning, it is usually treated as a routine training choice rather than as a measurable factor influencing cross-dataset robustness. As a result, there is limited evidence on which transformations genuinely improve transfer between datasets acquired using different scanners, vendors, and pre-processing pipelines. The novelty of this study is therefore not simply the number of augmentations evaluated, but the use of augmentation choice as a controlled variable for analysing domain-shift robustness in left ventricular segmentation. By separating in-domain performance from cross-dataset transfer and by evaluating both individual and pairwise transformations, this study identifies which augmentation families are consistently beneficial, redundant, or harmful under dataset shift. To the best of our knowledge, no prior work has systematically investigated individual and pairwise augmentation strategies as the central focus of a large-scale cross-dataset evaluation in echocardiographic segmentation. This study addresses this gap through the following contributions. 

\begin{enumerate}
    \item It provides a systematic evaluation of 29 data augmentation techniques for echocardiographic left ventricular segmentation across three independent datasets.
    \item It distinguishes in-domain performance from cross-dataset generalisation, showing that augmentation choice has limited influence when test data come from the same domain but becomes critical under domain shift.
    \item It evaluates selected pairwise augmentation combinations to identify synergistic, redundant, and antagonistic interactions that are not visible from single-augmentation experiments alone.
\end{enumerate}

\section{Related Work}
Upon examining the popularity of different augmentations, the list is dominated by natural image augmentations commonly found in deep learning frameworks; among them geometric augmentations are the most popular in ultrasound imaging\cite{tupper_revisiting_2025}. For instance, an ejection fraction (EF) prediction model applied random horizontal flips, small rotations $(\pm 10^{\circ})$, Scaling (70-\%130), and translations during training\cite{tabuco_two-view_2022}. In another study, the authors suggested a 10-degree rotation and vertical and horizontal flips for echocardiography images\cite{madani_fast_2018}, although vertical flipping appears less commonly adopted than horizontal flipping across the broader echocardiography literature. These transformations increased data diversity for segmenting LV chambers and improving EF accuracy. However, not all geometric transforms are used unrestrictedly. Studies generally avoid transformations that alter fundamental anatomy, such as certain types of flipping\cite{mazurowski_segment_2023, chen_optimizing_2024}. Some echo studies allow horizontal flips, which preserve anatomical orientation in some views, whereas vertical flips are generally applied more selectively \cite{saeed_contrastive_2022, holste_efficient_2024, guo_improved_2023}. Random cropping, zooming, and scaling are also applied in some echo classifiers\cite{saeed_contrastive_2022, guo_improved_2023, tabuco_two-view_2022, madani_fast_2018, wang_unsupervised_2022}; for example, an AI view-classification model used random resized cropping to augment training frames\cite{naser_artificial_2024}. Moderate rotations and affine transforms are used in many works, but usually constrained to avoid unrealistic view angles\cite{holste_efficient_2024, guo_improved_2023, chartsias_contrastive_2021, tabuco_two-view_2022, wang_unsupervised_2022, van_de_vyver_generative_2025,sfakianakis_gudu_2023}. 

To mimic anatomical variability, some echo segmentation models use elastic transforms\cite{tupper_revisiting_2025, mazurowski_segment_2023}. In echocardiography specifically, elastic transforms can augment limited segmentation datasets by creating new plausible heart shapes. Care is taken that these deformations do not violate the cardiac structure to avoid meaningless shapes. Overall, while less common than simple flips or rotations, elastic-deformation augmentation has been successfully used for echo segmentation tasks to account for anatomical deformation, like myocardial contractions\cite{tabuco_two-view_2022}.

Echo images are greyscale and often vary in brightness and contrast due to machine settings and patient factors. Photometric augmentations are therefore useful to improve generalisation across imaging conditions. Common adjustments include random changes in brightness and contrast, jittering, sometimes gamma corrections, or histogram equalisations\cite{saeed_contrastive_2022, guo_improved_2023, chartsias_contrastive_2021, wang_unsupervised_2022, van_de_vyver_generative_2025}. For example, a cardiac ultrasound segmentation study augmented training images by jittering brightness and contrast levels, simulating different gain settings \cite{van_de_vyver_generative_2025}. By amplifying heart muscle echo intensity, this domain-specific augmentation helped models focus on cardiac structures. Another photometric augmentation is contrast-limited adaptive histogram equalisation (CLAHE) reported in some echo studies to improve structure visibility, though not always explicitly mentioned in recent papers\cite{tupper_revisiting_2025}. Sharpening is another augmentation in this category\cite{wang_unsupervised_2022}.

Ultrasound images inherently contain speckle noise; accordingly,  augmentation can add or vary noise to improve robustness. Some echo deep learning pipelines inject Gaussian noise or a slight blur to simulate poorer image quality or motion\cite{saeed_contrastive_2022, guo_improved_2023}. For example, an echo segmentation study, using the HUNT4 and CAMUS datasets, evaluated augmentations like adding Gaussian noise and Gaussian blur, as well as simulating lower resolution\cite{van_de_vyver_generative_2025}. In ultrasound-specific contexts, researchers have proposed adding speckle noise at varying levels\cite{monkam_annotation_2023}. Wang et al. used a speckle noise augmentation in an echo video analysis\cite{wang_unsupervised_2022}, and Monkam et al. introduced multi-level speckle noise augmentation to progressively degrade image quality during training\cite{monkam_annotation_2023}. Conversely, speckle reduction augmentations artificially denoise images, helping models learn invariance to noise removal algorithms\cite{tupper_revisiting_2025, monkam_annotation_2023, ostvik_myocardial_2021, li_fast_2022}. 

Occlusion-based augmentations have also been explored in echocardiography to force models to rely on global context. Cutout or erasing involves blanking out a random patch of the image. This has been used in some echo training pipelines as a form of regularisation\cite{naidoo_consensus-guided_2025}. In advanced echo models, random erasing, a variant of cutout, has been applied; for example, EchoPrime, a recent echo foundation model, applied random erasing during training alongside other augmentations\cite{vukadinovic_comprehensive_2025}. Additionally, domain-inspired occlusions exist, as Gaussian-shadow and acoustic-shadow augmentation darkens a region with a realistic shadow shape, simulating how rib shadows or lung tissue might obscure the heart \cite{tupper_revisiting_2025, ostvik_myocardial_2021}. This was first proposed by Smistad et al. for ultrasound\cite{smistad_highlighting_2018}. By adding synthetic shadows and random erasing in echo images, the model learns to handle missing data. 

Domain and style augmentations aim to simulate variations in ultrasound machines or imaging conditions. In echocardiography, this includes both stylizing images and using synthetic data generation. A common approach is to simulate ultrasound artefacts and differences; for example, Ostvik et al. developed augmentations for depth-dependent attenuation, dimming distal parts of the image, and haze to mimic poorer penetration\cite{tupper_revisiting_2025, ostvik_myocardial_2021}. These techniques have been applied to echocardiographic images to improve model generalisation across varying imaging depths and gain settings. Another line of work focuses on modifying the ultrasound field of view. Some studies emphasise preserving the intrinsic fan-shaped geometry of ultrasound during augmentation, proposing fan-preserving cropping or zooming operations that narrow or widen the sector while maintaining the original ultrasound acquisition geometry\cite{tupper_revisiting_2025, mazurowski_segment_2023} effectively emulating different probe zoom levels or patient sizes. Similarly, probe angle randomisation was introduced via cone position adjustment and perspective adjustment augmentations in a recent echocardiography study; these geometrically transform the echo fan as if the transducer angle was changed, generating new view angles\cite{sfakianakis_gudu_2023}. 

As reported in a survey study, augmentation techniques in ultrasound deep learning studies are unevenly distributed when ordered by frequency among the 165 studies they consider for their research. The most adopted operations are flip (approx. 60 studies), rotation (approx. 48), zoom (approx. 29), random cropping (approx. 18), and translation (approx. 12). These are followed by moderately used photometric augmentations, including contrast adjustment, brightness adjustment, Gaussian noise, and gamma adjustment, each appearing in roughly 8-10 studies. Less frequently reported general transformations include Gaussian-shadow, haze artifact addition, elastic transform, depth attenuation, random noise, intensity windowing, shear, colour jitter, speckle reduction, image resampling artefacts, intensity adjustment, cutout, normalisation, multiplicative noise, centre crop, and JPEG compression, typically used in fewer than five studies. At the lowest end of the frequency spectrum, speckle noise suppression, low-resolution simulation, speckle noise,  perspective position adjustment, fan-shape-preserving zoom, multi-level speckle noise, unsharp masking, adaptive gamma correction, frequency-domain mixing, manifold mix-up, image-puzzle mixing, class mix, cut mix, hue adjustment, mirroring, random erasing, salt-and-pepper noise, myocardium intensity adjustment, blur, sharpen, field-of-view masking, nonlinear colour map, wrap, speckle parameter mapping, time-gain compensation, acoustic shadow, and cone position adjustment, appear only sporadically, often in one to three studies\cite{tupper_revisiting_2025}.

\section{Methodology}
The study was designed as a multi-dataset experimental analysis to quantify the impact of data augmentation on 2D echocardiography segmentation performance and cross-dataset generalisation. Segmentation accuracy was first evaluated using the mean Dice coefficient and mean Intersection over Union (IoU) on the test sets. Additionally, the main objective was to assess robustness under domain shift. All experiments used a fixed U-Net-based segmentation architecture and identical optimisation settings, varying only the augmentation configuration so that observed performance differences could be attributed to augmentation rather than changes in model capacity or training regime.

\subsection{LV segmentation}
One of the critical tasks in echocardiography is LV segmentation. This involves classifying image pixels into background and LV regions, as shown in Figure 2. By identifying an accurate mask for the LV, clinicians can calculate the EF, a key indicator of cardiac function. This process can be fully automated by relying solely on computer vision models that directly identify LV boundaries or detect the ED and ES phases to assist practitioners in detecting heart-related abnormalities more efficiently. Developing a robust and generalisable model for LV segmentation can significantly reduce the diagnostic workload and decrease the dependence on the availability and experience of expert echocardiologists for routine tasks\cite{naidoo_consensus-guided_2025}. 
\begin{figure}[ht]
\centering
\includegraphics[width=\linewidth]{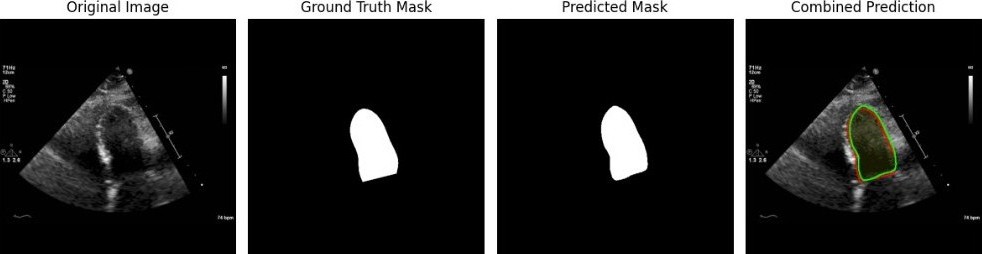}
\caption{
Segmentation and masks by human annotator
}
\label{fig:example}
\end{figure}

\subsection{Dataset}
This study utilised three echocardiography datasets, including Unity, CAMUS (CA), and EchoNet Dynamic (ED). Additionally, we employed the Consensus (CS) dataset (UnityLV-MultiX) as a specialised test set to replace the original Unity test data. The CS dataset consists of 100 apical four-chamber (A4C) videos from the Imperial College Healthcare NHS Trust, where each frame was independently annotated by 11 experts to provide a robust reference standard \cite{naidoo_consensus-guided_2025, ouyang_video-based_2020, ouyang_echonet-dynamic_2020, leclerc_deep_2019}. To ensure fair comparisons across all augmentation experiments, the Unity, CA, and ED datasets were divided into fixed, non-overlapping training, validation, and test subsets, following the official partitioning schemes recommended by their respective publishers. Table 1 summarises the characteristics and splits for each dataset.

For datasets provided as cine loops, the ED and ES frames were pre-defined and accompanied by LV annotations. All datasets were fully anonymised and licensed for research use. Figure 3 displays a representative example from each dataset for visual comparison.

\begin{table}[ht]
\centering
\caption{Overview of the datasets used in this study}
\label{tab:game_personality}
\renewcommand{\arraystretch}{1.5} 
\setlength{\tabcolsep}{6pt} 

\begin{tabular}{|l|l|l|l|}
\hline
\textbf{Features} & \textbf{Unity} & \textbf{CAMUS} & \textbf{EchoNet Dynamic} \\
\hline
Access & Private & Public & Public \\
\hline
Train set & 2057 & 1600 & 14930 \\
\hline
Validation set & 369 & 200 & 2554 \\
\hline
Test set & 200\textsuperscript{*} & 200 & 2576 \\
\hline
Total size & 2626 & 2000 & 20060 \\
\hline
Number of videos & 1224 & 1000 & 10030 \\
\hline
Manual annotations & ED - ES & ED - ES & ED - ES \\
\hline
Views & A4C & A2C\textsuperscript{**} - A4C & A4C \\
\hline
Original size & Variable & Variable & 112×112 pixels \\
\hline
Used size & 512×512 pixels & 512×512 pixels & 512×512 pixels \\
\hline
Scanner brands & Philips - GE \cite{howard_automated_2021} & GE \cite{leclerc_deep_2019} & Philips - Siemens \cite{Ouyang_ed_2019} \\
\hline
\end{tabular}

\vspace{2mm}
\parbox{0.9\linewidth}{
    \footnotesize
    \textsuperscript{*} The 200-image Consensus dataset was used in all tests instead of the original 375-image Unity test set.\\
    \textsuperscript{**} Apical 2-chamber
}
\end{table}

\begin{figure}[H]
\centering
\includegraphics[width=0.7\linewidth]{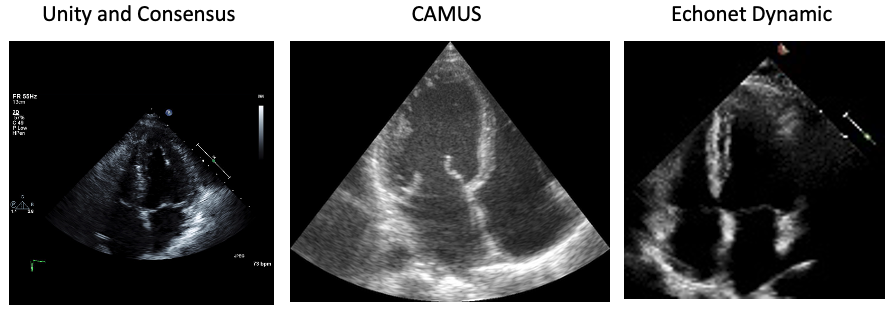}
\caption{
Representative echocardiographic images from the datasets used in this study
}
\label{fig:example}
\end{figure}

\subsection{Domain Shift}

The datasets used in this study originate from multiple ultrasound vendors and scanner platforms, as described in the previous subsection. CA was acquired using GE Vivid E95 ultrasound scanners with an M5S probe and exported from EchoPAC software \cite{leclerc_deep_2019}. ED includes examinations from Philips systems, including iE33, Epiq 5G, and Epiq 7C, as well as a Siemens Acuson SC2000 system \cite{Ouyang_ed_2019}. The Unity and Consensus datasets were collected in UK centres using Philips and GE scanners \cite{howard_automated_2021}.

Although all datasets represent standard 2D transthoracic echocardiography, the imaging systems differ in hardware design and proprietary processing pipelines. Vendor-specific beamforming, dynamic range compression, speckle reduction, gain mapping, sharpening, and post-processing can affect image appearance, producing systematic differences in greyscale distribution, contrast, texture, edge definition, and noise characteristics. Even under standardised acquisition protocols, scanner presets and processing chains can shift pixel intensity distributions and texture statistics. Therefore, manufacturer differences are widely recognised as a source of domain shift in medical imaging \cite{kushol_dsmri_2023, guo_impact_2024, yan_mri_2020}.

From a deep learning perspective, these differences may cause models trained on one vendor or dataset to learn scanner-specific characteristics, such as contrast profiles, texture smoothness, speckle patterns, or edge-enhancement features, alongside anatomical information. When evaluated on images acquired or pre-processed using different pipelines, model performance may decline because of changes in low-level image statistics, not only clinically meaningful anatomical variation. However, vendor and scanner differences are not the only sources of domain variability; dataset-specific acquisition protocols and pre-processing steps can also influence image appearance.

Figure 4 quantitatively reflects some of these differences. Each image characteristic was computed for individual images and then averaged across datasets to highlight systematic distribution shifts. Before computing these characteristics, zero-intensity pixels were excluded because regions outside the fan-shaped ultrasound sector contain only zero values, and the size of these background regions varies between datasets. Removing zero-valued pixels ensured that the comparison focused on meaningful image content rather than differences in black sector margins, allowing the extracted characteristics to better capture image-content-related and structurally meaningful variation.

The plotted characteristics include brightness, contrast, skewness, kurtosis, and dynamic range. Brightness, measured using mean or median pixel intensity, indicates the overall intensity level within the valid imaging region. Contrast was quantified using the standard deviation and the robust 5th-95th percentile intensity range, reflecting intensity variation. Skewness describes distribution asymmetry, with positive values indicating a longer tail toward brighter intensities and negative values indicating a longer tail toward darker intensities. Kurtosis measures peakedness and tail heaviness, where higher values indicate sharper peaks and more pronounced intensity outliers, while lower values indicate flatter distributions. Dynamic range was defined as the difference between maximum and minimum intensity values. For LV core characteristics, segmentation masks were applied and the same characteristics were computed only within the masked LV core region.

Based on Figure 4, clear differences in pixel intensity distributions can be observed across the datasets, suggesting the presence of domain shift. Since segmentation models may learn dataset-specific brightness, contrast, and texture patterns during training, their performance can decline on datasets with different image distributions. This is consistent with the experimental results presented in the following sections and supports the need for domain-robust augmentation strategies to improve generalisation across heterogeneous imaging platforms.

\begin{figure}[H]
\centering

\begin{subfigure}{0.32\linewidth}
    \centering
    \includegraphics[width=\linewidth]{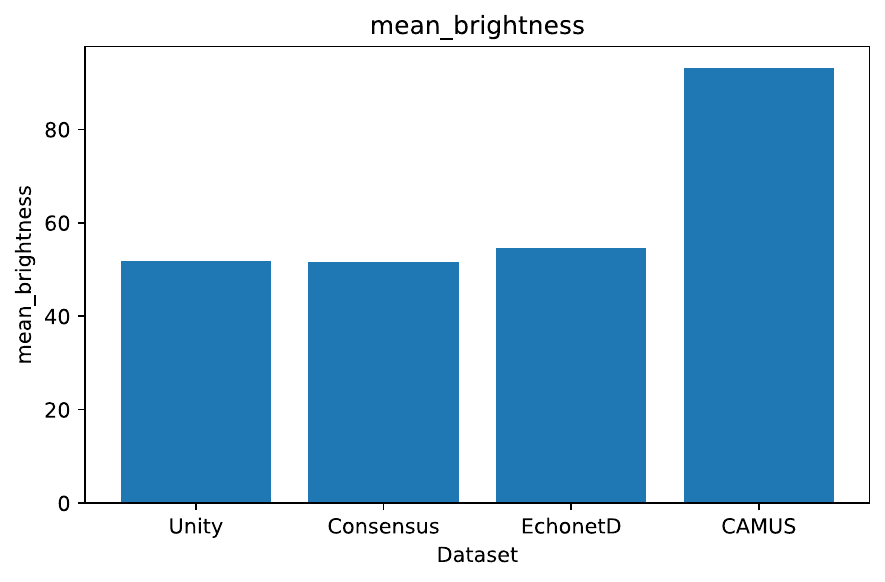}
\end{subfigure}
\hfill
\begin{subfigure}{0.32\linewidth}
    \centering
    \includegraphics[width=\linewidth]{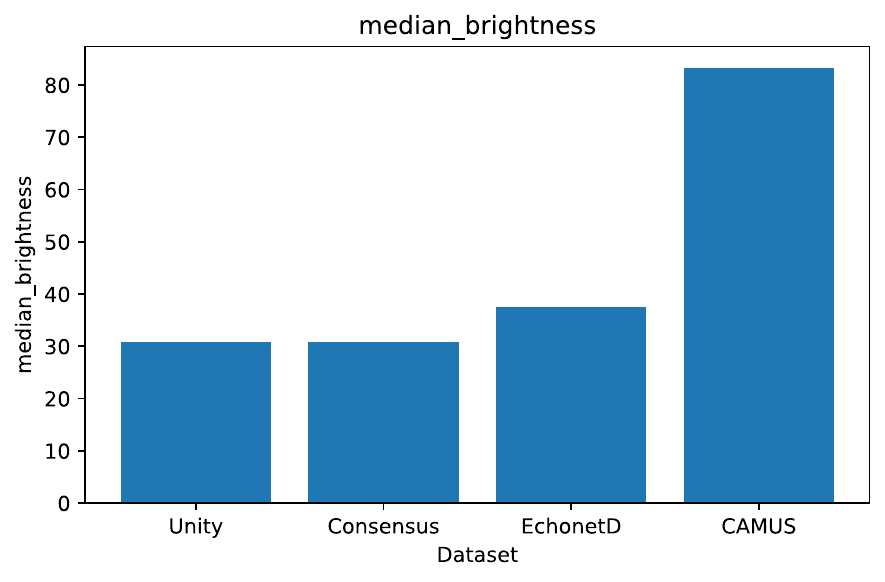}
\end{subfigure}
\hfill
\begin{subfigure}{0.32\linewidth}
    \centering
    \includegraphics[width=\linewidth]{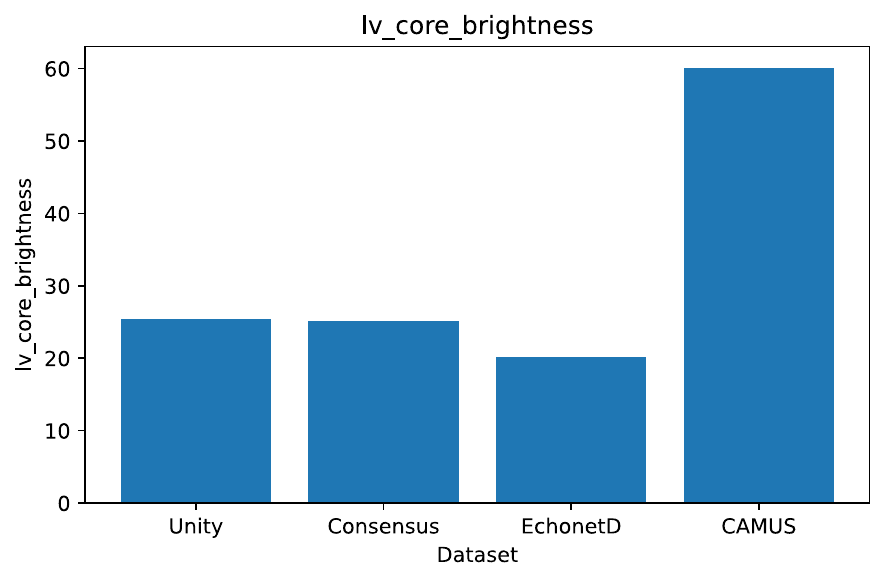}
\end{subfigure}

\vspace{-3mm}

\begin{subfigure}{0.32\linewidth}
    \centering
    \includegraphics[width=\linewidth]{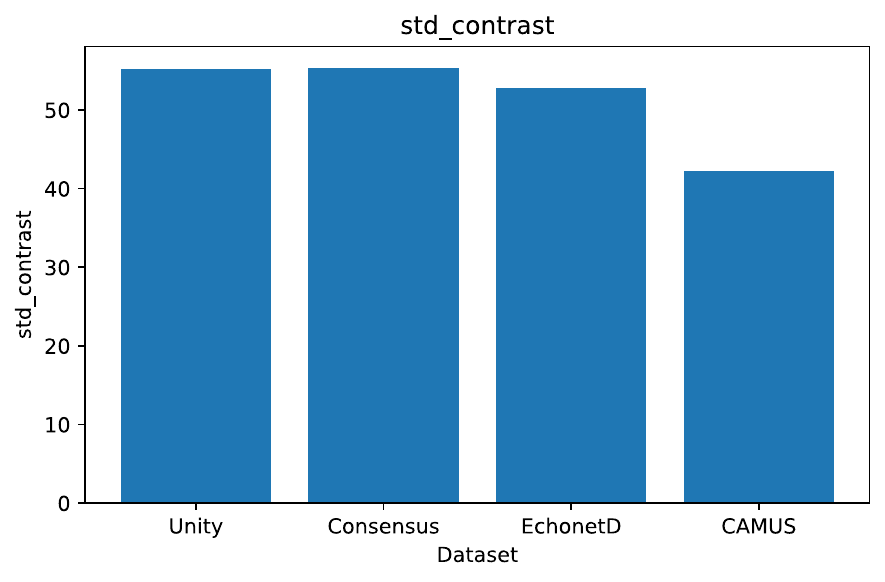}
\end{subfigure}
\hfill
\begin{subfigure}{0.32\linewidth}
    \centering
    \includegraphics[width=\linewidth]{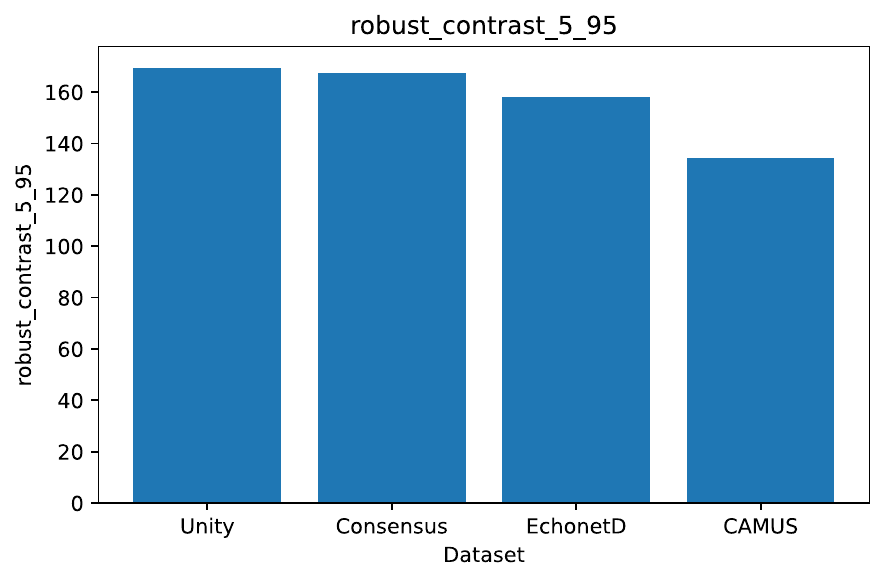}
\end{subfigure}
\hfill
\begin{subfigure}{0.32\linewidth}
    \centering
    \includegraphics[width=\linewidth]{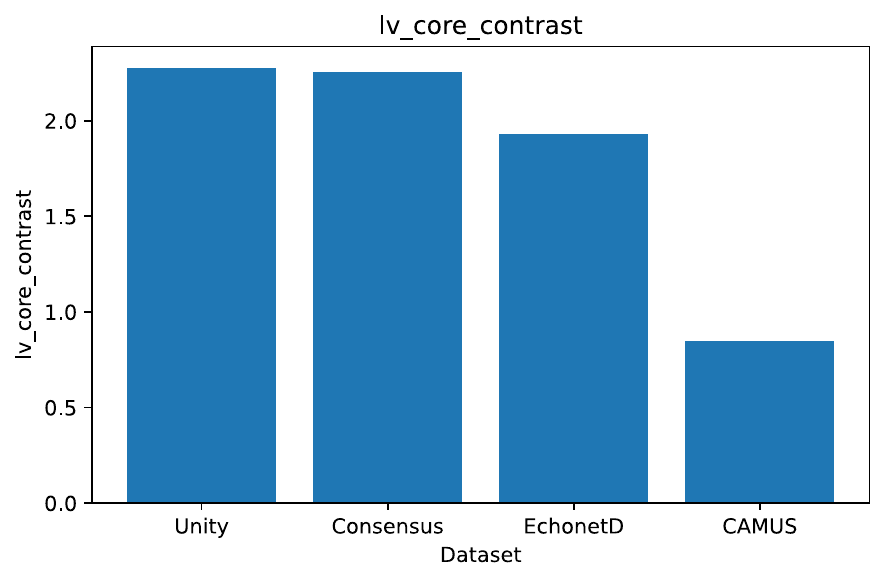}
\end{subfigure}

\vspace{-3mm}

\begin{subfigure}{0.32\linewidth}
    \centering
    \includegraphics[width=\linewidth]{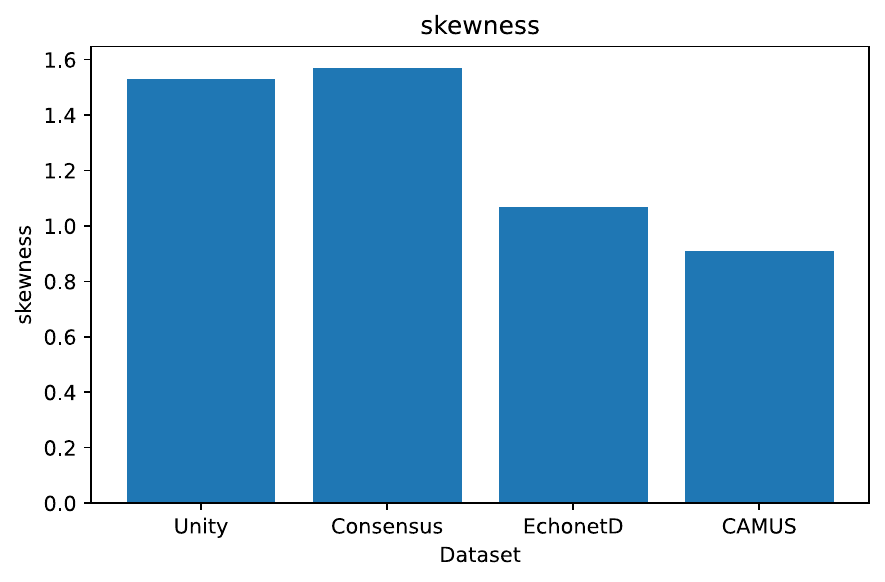}
\end{subfigure}
\hfill
\begin{subfigure}{0.32\linewidth}
    \centering
    \includegraphics[width=\linewidth]{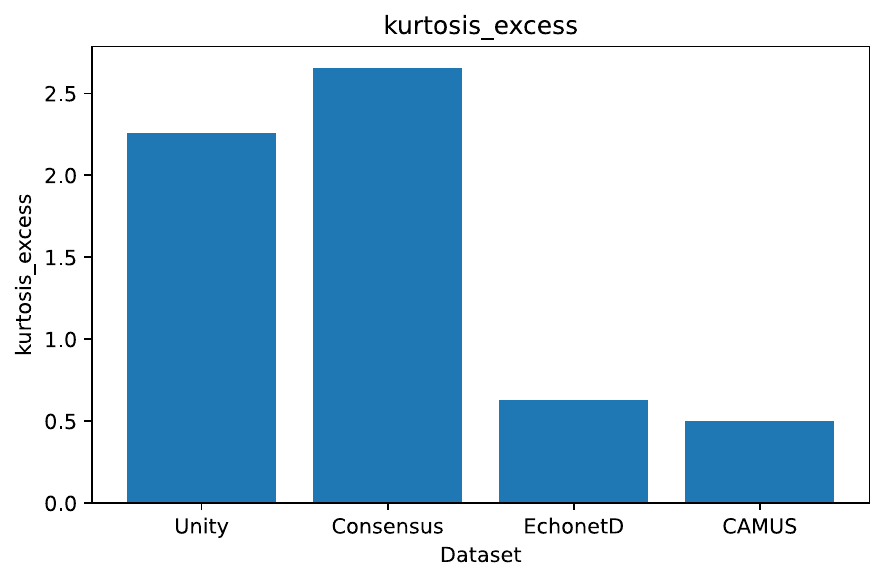}
\end{subfigure}
\hfill
\begin{subfigure}{0.32\linewidth}
    \centering
    \includegraphics[width=\linewidth]{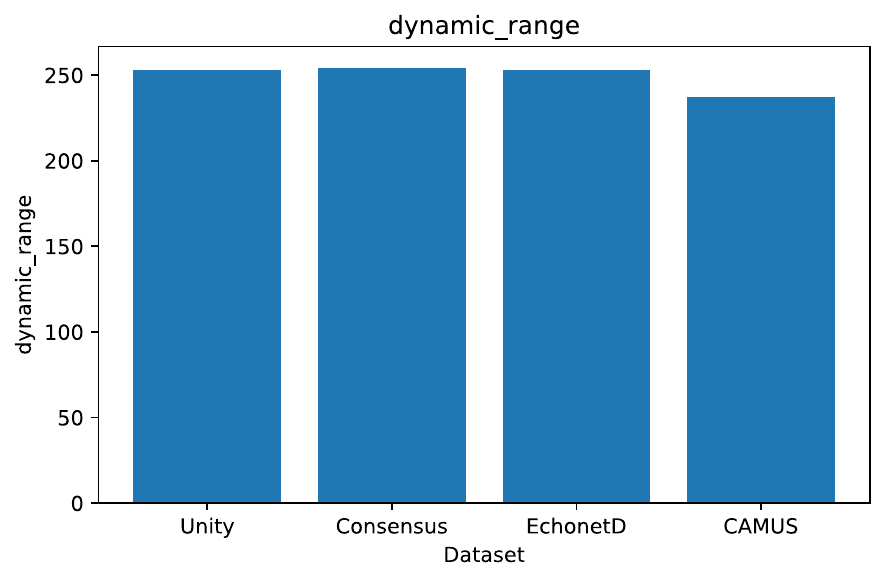}
\end{subfigure}

\caption{Intensity and contrast statistics across the four datasets, computed after excluding zero-valued background pixels}
\label{fig:gradcam_9grid}
\end{figure}

\subsection{Network architecture}
The segmentation network is a 2-D U-Net implemented in PyTorch, designed for echocardiographic cine frames, which takes single-channel greyscale inputs. The encoder consists of five stages with feature dimensions of 32, 64, 128, 256, and 512, each composed of two 3×3 convolutional layers with batch normalisation and ReLU activation, followed by 2×2 max pooling. A bottleneck block expands the feature representation to 1024 channels at the lowest spatial resolution. The decoder mirrors the encoder with five stages of transposed convolutions.

The network was trained in a supervised manner using the Adam optimiser with an initial learning rate of $1 \times 10^{-4}$ and a batch size of 16. BCEWithLogitsLoss was employed for pixel-wise segmentation. Early stopping was applied based on the validation Dice score with a patience of 12 epochs and a minimum improvement threshold of 0.001. A Reduce-on-Plateau learning rate scheduler reduced the learning rate by a factor of 0.5 after five epochs without validation improvement, with a minimum learning rate of $1 \times 10^{-6}$.  Training was conducted for a maximum of 200 epochs. All input frames were directly  resized to 512×512 pixels. The model was trained from scratch, without pre-training, to isolate the effect of data augmentation on learning performance. Additionally, all experiments were implemented in PyTorch v2.9.1 with CUDA 12.8 and trained on a single NVIDIA RTX 4090 GPU (24 GB VRAM). All Jupyter Notebook implementations and the corresponding outputs (exported as CSV files for each experimental run) are publicly available on the linked GitHub repository (\url{https://github.com/soroushelyasi/augmentation_benchmark}).

\subsection{Experiment structure}
This experiment consists of two phases. In the first phase, 29 commonly used data augmentation techniques for ultrasound imaging were selected from the literature and evaluated on echocardiography images. The first four groups include commonly used photometric, noise and quality simulation, geometric, and occlusion-based transformations. The fifth group includes ultrasound-specific augmentations that were recently proposed for ultrasound images \cite{tupper_revisiting_2025}. These augmentations are divided into five groups, as shown in Figure 5. All transformations were applied within mild parameter ranges to preserve anatomical plausibility and the structural integrity of the left ventricle as much as possible.

\begin{figure}[ht]
\centering
\includegraphics[width=\linewidth]{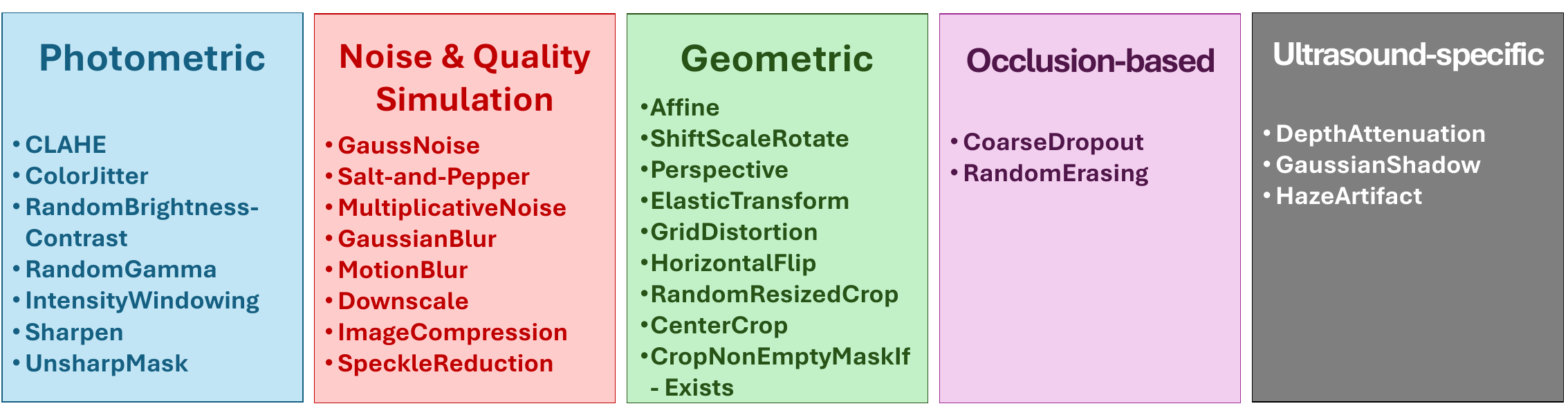}
\caption{
Grouping of the 29 data augmentation techniques into five categories
}
\label{fig:example}
\end{figure}

The objective of this phase was to systematically assess each augmentation to determine its suitability for echocardiography data. Given the importance of hyperparameter optimisation, three to five configurations were tested for each augmentation, as detailed in Appendix D. Each configuration was executed three times, and the mean Dice and IoU scores were reported as the final performance metrics to reduce the impact of model initialisation variability. For cases requiring a fan-shaped mask, we followed the strategy described in Appendix A.

The experiment comprised nine evaluation settings, representing all combinations of training and transfer evaluation across the three datasets, and each model was trained on one dataset and evaluated on both the source dataset and the remaining two datasets without hyperparameter retuning.

The second phase aimed to identify the most effective and statistically significant augmentations. A threshold criterion was introduced to determine which augmentations should proceed to further evaluation. One of the selection criteria was statistical significance, assessed using independent t-tests comparing each augmentation against the baseline model without augmentation, with the significance level set at 0.05. Under this criterion, an augmentation was considered beneficial if it produced improvements in at least six of the nine experimental categories and demonstrated five or more statistically significant improvements. These thresholds were chosen for two main reasons. First, they ensured that only augmentations showing consistent benefits across the majority of evaluations were retained, thereby reducing the inclusion of augmentations with marginal or unstable effects. Second, the thresholds were intentionally designed to restrict the number of candidate augmentations, since a large number of selected augmentations would lead to a rapidly expanding combinatorial search space during the subsequent combination phase, substantially increasing computational cost. 

A second selection criterion was based on the average Dice coefficient obtained by each augmentation across all evaluations. Augmentations were ranked according to their mean Dice performance, and the top four augmentations were selected and combined with those identified through the statistical significance criterion. The choice of the top four was similarly motivated by the need to balance broad representation of high-performing augmentations with the practical requirement of maintaining a manageable search space for Phase 2.

Following the selection of augmentation settings based on the defined criteria, an analysis of pairwise augmentation combinations was conducted to evaluate their combined effects. Some augmentation types appeared more than once in the selected set because multiple hyperparameter settings of the same augmentation satisfied the selection criteria. To avoid combining different parameter settings of the same augmentation type, these variants were not paired with one another. Instead, each selected setting was paired only with selected settings from other augmentation types.

\section{Results}
This study is presented in two phases. The first phase examines each selected augmentation individually to isolate their effects, as detailed in the methodology section. The second phase focuses on the subset of augmentations that yielded superior results in the first phase. 

\subsection{Individual augmentation}
Based on the proposed methodology, 29 augmentation techniques were selected from the literature and grouped into five categories. Each augmentation setting was evaluated independently and repeated three times. For each setting, the mean Dice and IoU scores were calculated across the three runs. The models were evaluated in-domain and across datasets to assess performance and generalisability. To summarise the overall behaviour of each augmentation across these scenarios, we report the unweighted mean of the performance scores across the nine configurations. However, because the differences in in-domain performance were comparatively small across augmentations, most variation in the aggregated score was mainly influenced by cross-dataset performance. Therefore, this summary is interpreted primarily as an indicator of how each augmentation affects model generalisation under domain shift, rather than as a measure of small performance differences within a single dataset.

Bar charts for each augmentation are presented in Figure 6, where the red horizontal line denotes the baseline performance obtained without augmentation. Figure 7 presents the raw Dice score values, while Figure 8 shows the differences relative to the no-augmentation baseline (NONE), calculated by subtracting the corresponding baseline Dice scores. Consistent group-based colour coding is used across the heatmaps and bar charts to indicate the augmentation categories.

\begin{figure}[ht]
\centering
\includegraphics[width=0.87\linewidth]{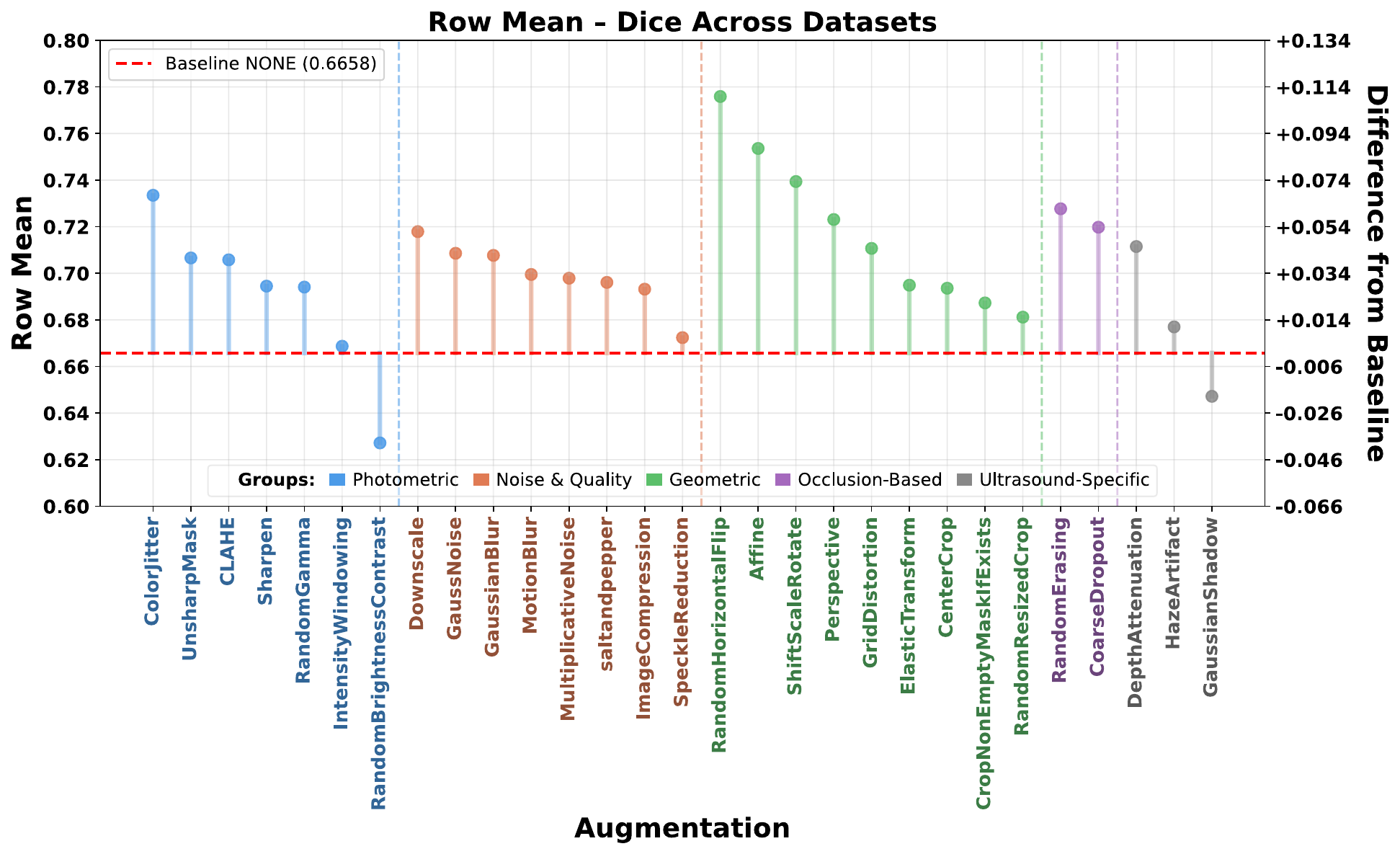}
\caption{
Row-Mean Dice Performance Across Datasets and Relative Improvements Over the NONE
}
\label{fig:example}
\end{figure}

\begin{figure}[ht]
\centering
\includegraphics[width=1\linewidth]{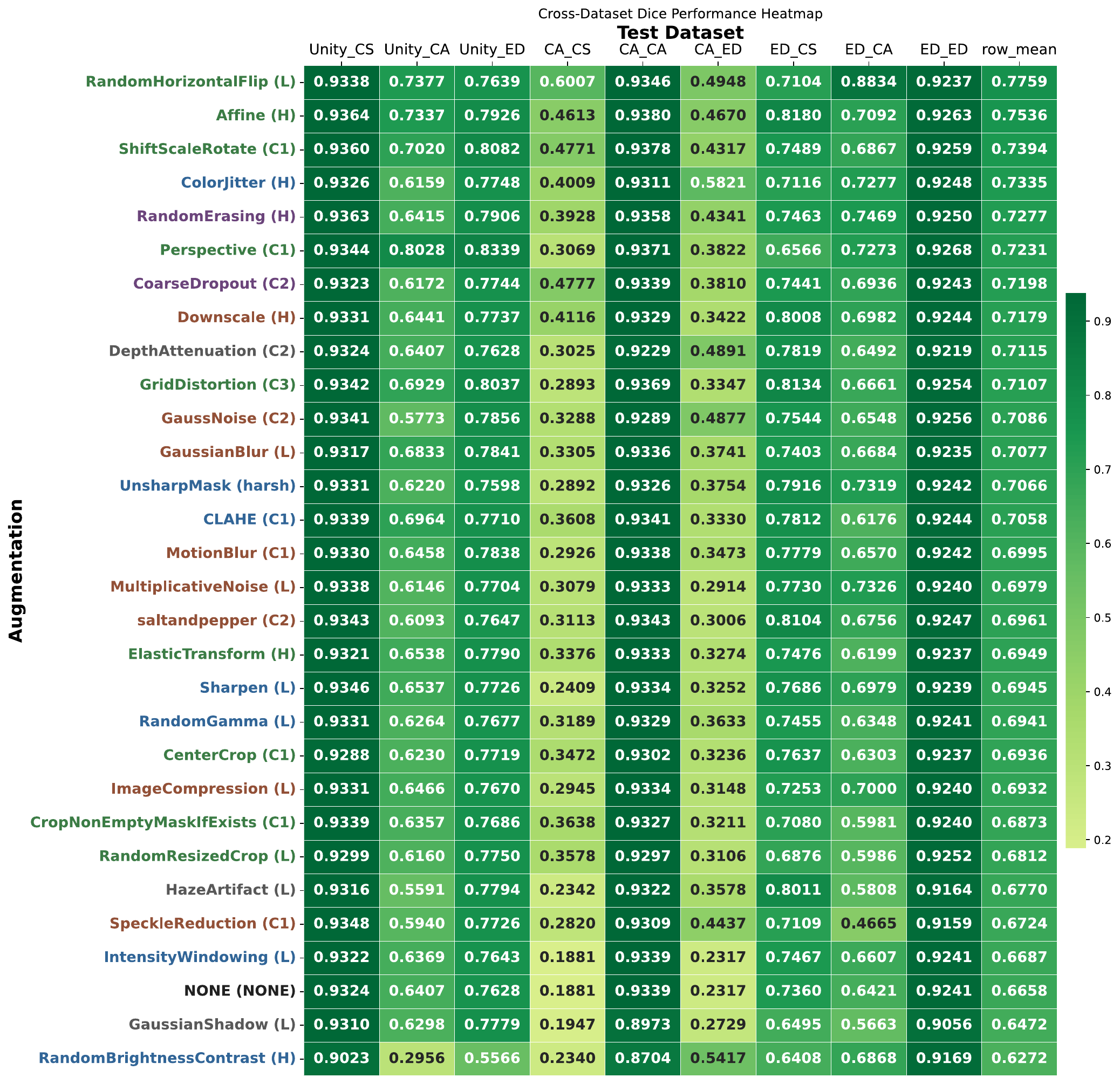}
\caption{
Dice performance heatmap for all augmentation strategies 
}
\label{fig:example}
\end{figure}

For each augmentation, the selected parameter setting is indicated in parentheses after the augmentation name. These selected settings correspond to the best-performing configurations according to the defined selection criteria. To support interpretation, Appendix D provides a detailed description of the selected settings and their associated hyperparameters, together with illustrative examples of their visual effects. Complete results for all other parameter configurations are provided in CSV format in the project’s GitHub repository. 

Although Figures 6-8 provide a clear visual comparison of performance, statistical analysis is needed to confirm whether improvements over the NONE are statistically supported. Following the methodology, we compared three independent runs for each augmentation with their no-augmentation results using an independent t-test, the results of which are reported in Appendix C. In addition to the Dice metric, the results for IoU with the same parameter settings are presented in Figures 9 to 11. These figures follow the same structure and interpretation as in Figures 6 to 8, respectively.

After evaluating each augmentation individually, the next step is to investigate their combined effect when applied in pairs. Testing all possible combinations, or those with more than two augmentations, is computationally infeasible. This is especially true since each configuration must be repeated three times. Therefore, a principled selection strategy is applied to identify a small set of promising augmentations for pairwise evaluation.

\begin{figure}[ht]
\centering
\includegraphics[width=1\linewidth]{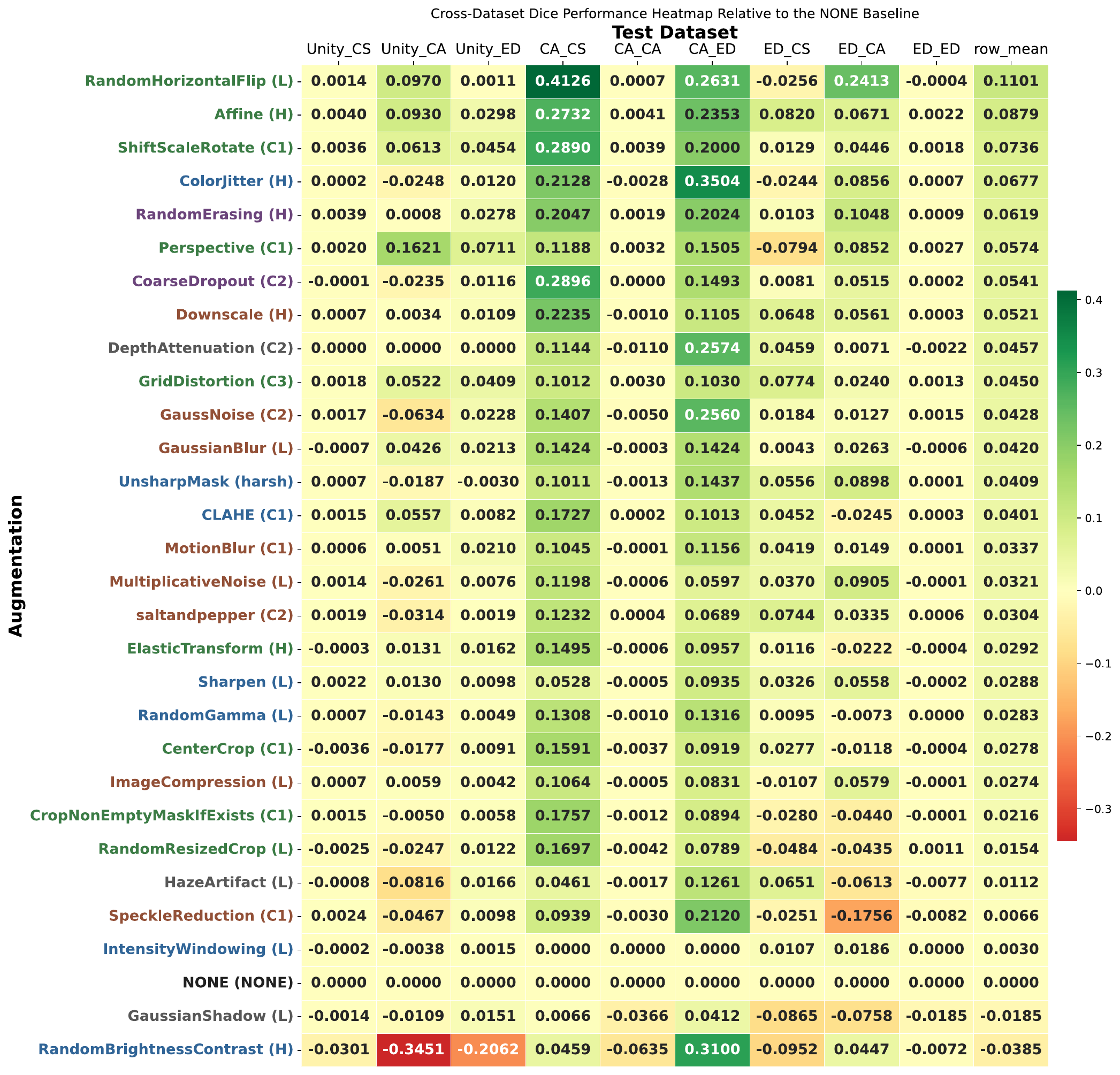}
\caption{
Relative Dice performance with respect to the NONE
}
\label{fig:example}
\end{figure}

Two complementary selection criteria were employed. First, based on the mean-based ranking discussed previously, we selected the top four augmentation settings that demonstrated the most promising overall performance based on the raw mean. These were random horizontal flip (L), affine (H), shift scale rotate(C1), and ColorJitter (H). Second, we considered statistical robustness using the selection thresholds defined in Section 3.5. Augmentation-setting pairs that satisfied these significance and positivity criteria were retained for further pairwise evaluation, as reported in Table 2. Augmentation-setting pairs appearing in the intersection of these two selection strategies were included only once in the subsequent pairwise experiments. Additionally, different parameter settings of the same augmentation were not paired together, as outlined in the methodology.

\begin{figure}[ht]
\centering
\includegraphics[width=1\linewidth]{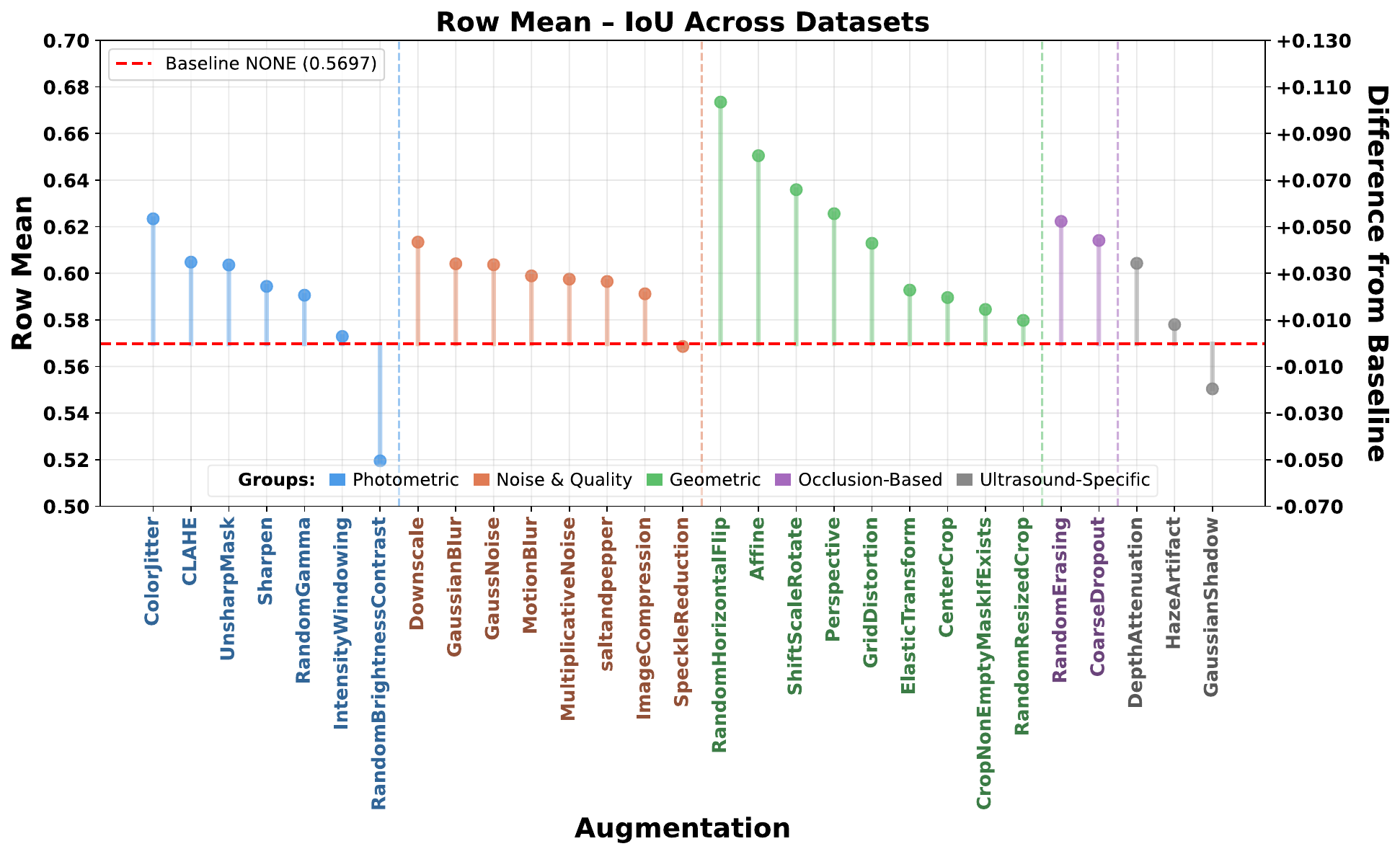}
\caption{
Row-Mean IoU Performance Across Datasets and Relative Improvements Over the NONE
}
\label{fig:example}
\end{figure}

\begin{figure}[ht]
\centering
\includegraphics[width=1\linewidth]{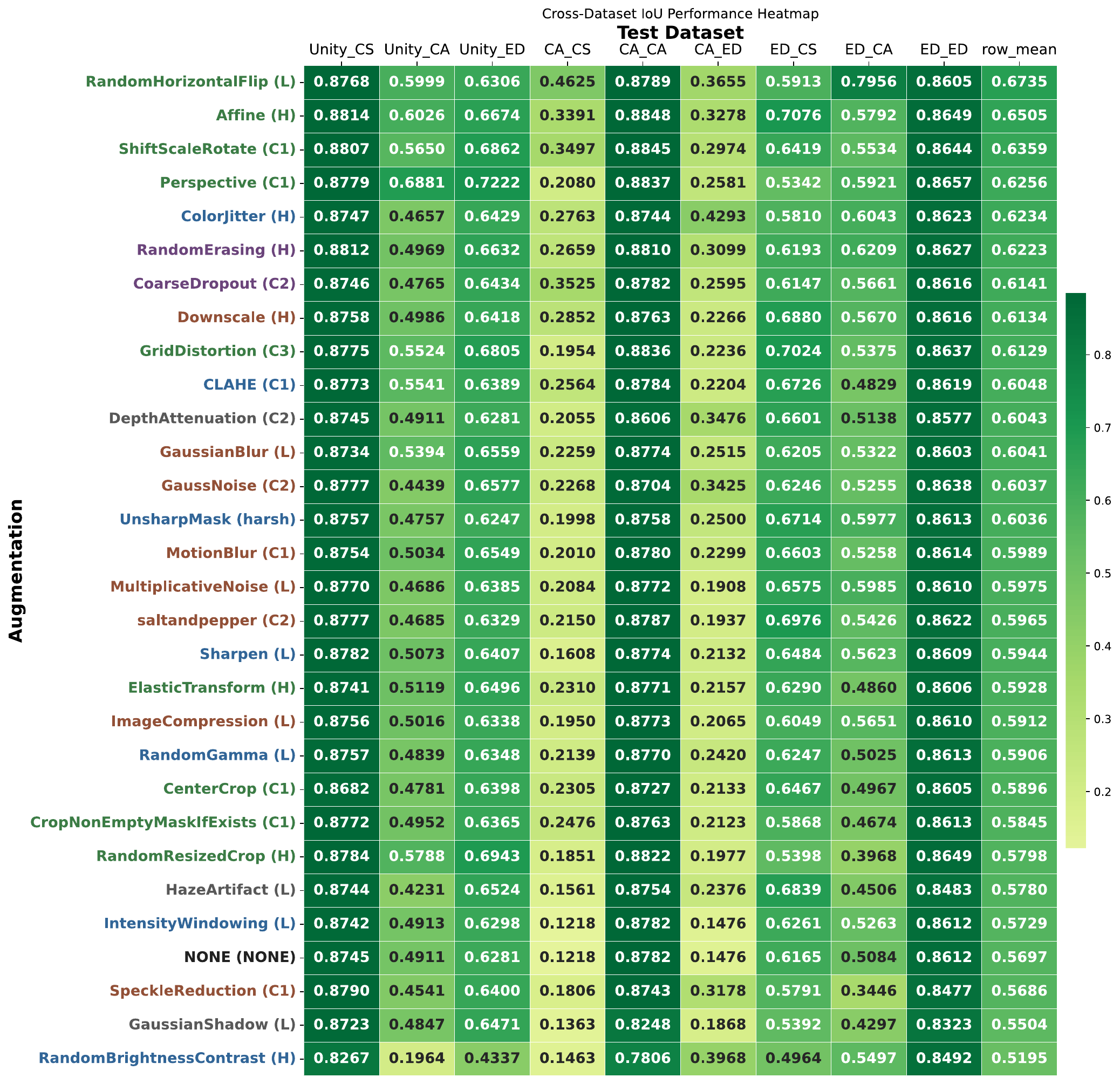}
\caption{
IoU performance heatmap for all augmentation strategies 
}
\label{fig:example}
\end{figure}

\begin{figure}[ht]
\centering
\includegraphics[width=1\linewidth]{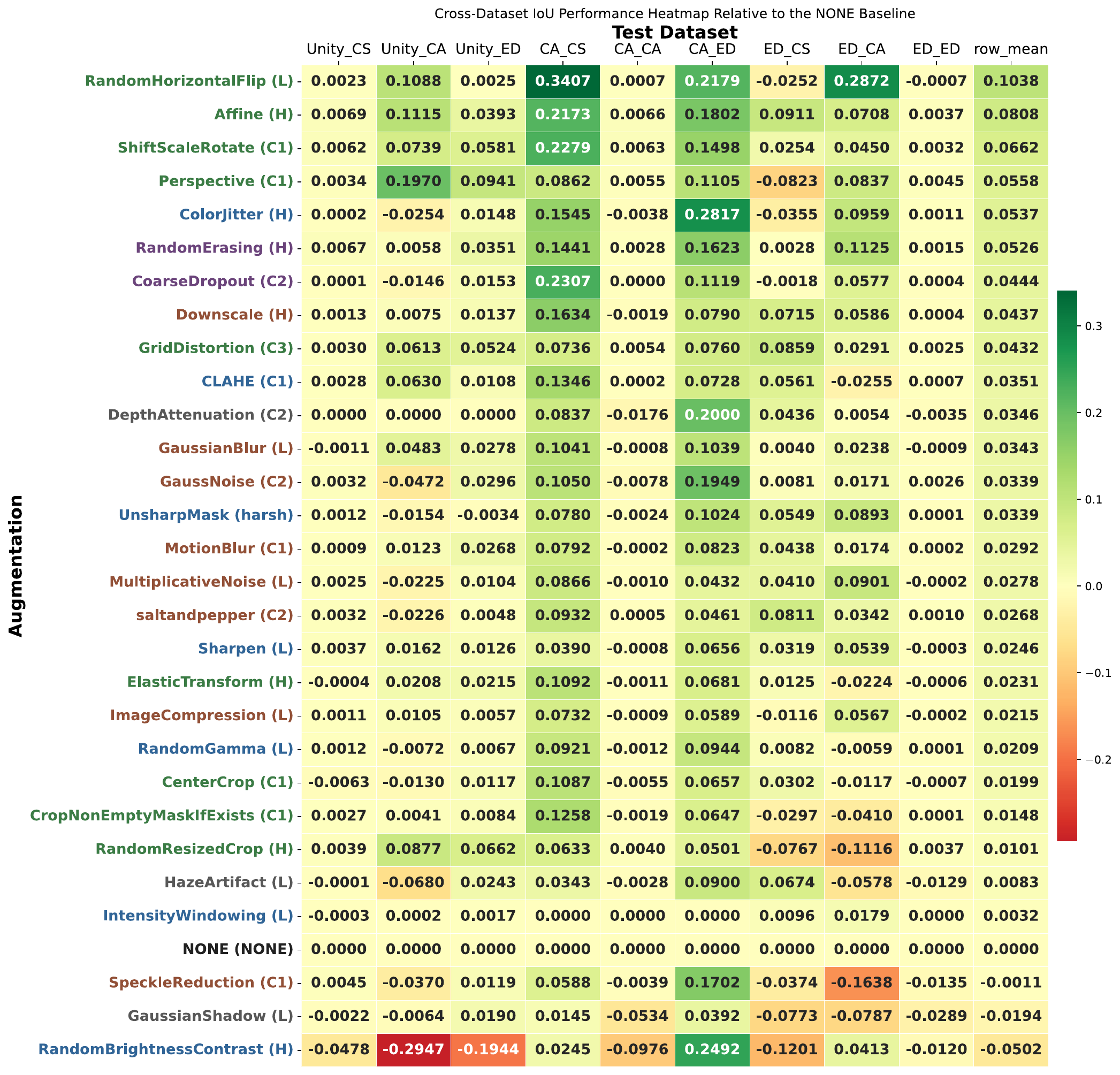}
\caption{
Relative IoU performance with respect to the NONE
}
\label{fig:example}
\end{figure}

\begin{table}[ht]
\caption{Ranked augmentation-setting candidates selected for pairwise augmentation experiments based on significance and positive-improvement counts, both out of nine}
\label{tab:ranked_augmentation}
\begin{center}
\begin{tabular}{|l|l|l|l|}
\hline
\rule[-1ex]{0pt}{3.5ex} Augmentation & Settings & Sig05 Count & Pos. Count \\
\hline
\rule[-1ex]{0pt}{3.5ex} Affine & H & 8 & 9 \\
\hline
\rule[-1ex]{0pt}{3.5ex} ShiftScaleRotate & C1 & 5 & 9 \\
\hline
\rule[-1ex]{0pt}{3.5ex} GridDistortion & C3 & 5 & 9 \\
\hline
\rule[-1ex]{0pt}{3.5ex} Affine & L & 7 & 9 \\
\hline
\rule[-1ex]{0pt}{3.5ex} GridDistortion & C2 & 6 & 9 \\
\hline
\rule[-1ex]{0pt}{3.5ex} GridDistortion & H & 6 & 9 \\
\hline
\rule[-1ex]{0pt}{3.5ex} Perspective & C1 & 5 & 7 \\
\hline
\rule[-1ex]{0pt}{3.5ex} ShiftScaleRotate & H & 6 & 7 \\
\hline
\rule[-1ex]{0pt}{3.5ex} Perspective & H & 6 & 7 \\
\hline
\rule[-1ex]{0pt}{3.5ex} Affine & C1 & 6 & 7 \\
\hline
\rule[-1ex]{0pt}{3.5ex} ShiftScaleRotate & L & 5 & 7 \\
\hline
\end{tabular}
\end{center}
\end{table}

\subsection{Pairwise augmentation combinations }

In this section, we follow the same procedure used for single augmentations, with the difference that pairwise augmentations are evaluated to examine the effect of combining augmentation strategies. Figure 12 shows the Dice bar chart. The detailed heatmap results and statistical significance table are provided in Appendix C.

Finally, Figure 13 presents the corresponding IoU results for the pairwise augmentation experiments. The figures follow the same augmentation pairs as Figure 12, enabling a direct comparison between row mean Dice and IoU metrics.

\begin{figure}[ht]
\centering
\includegraphics[width=1\linewidth]{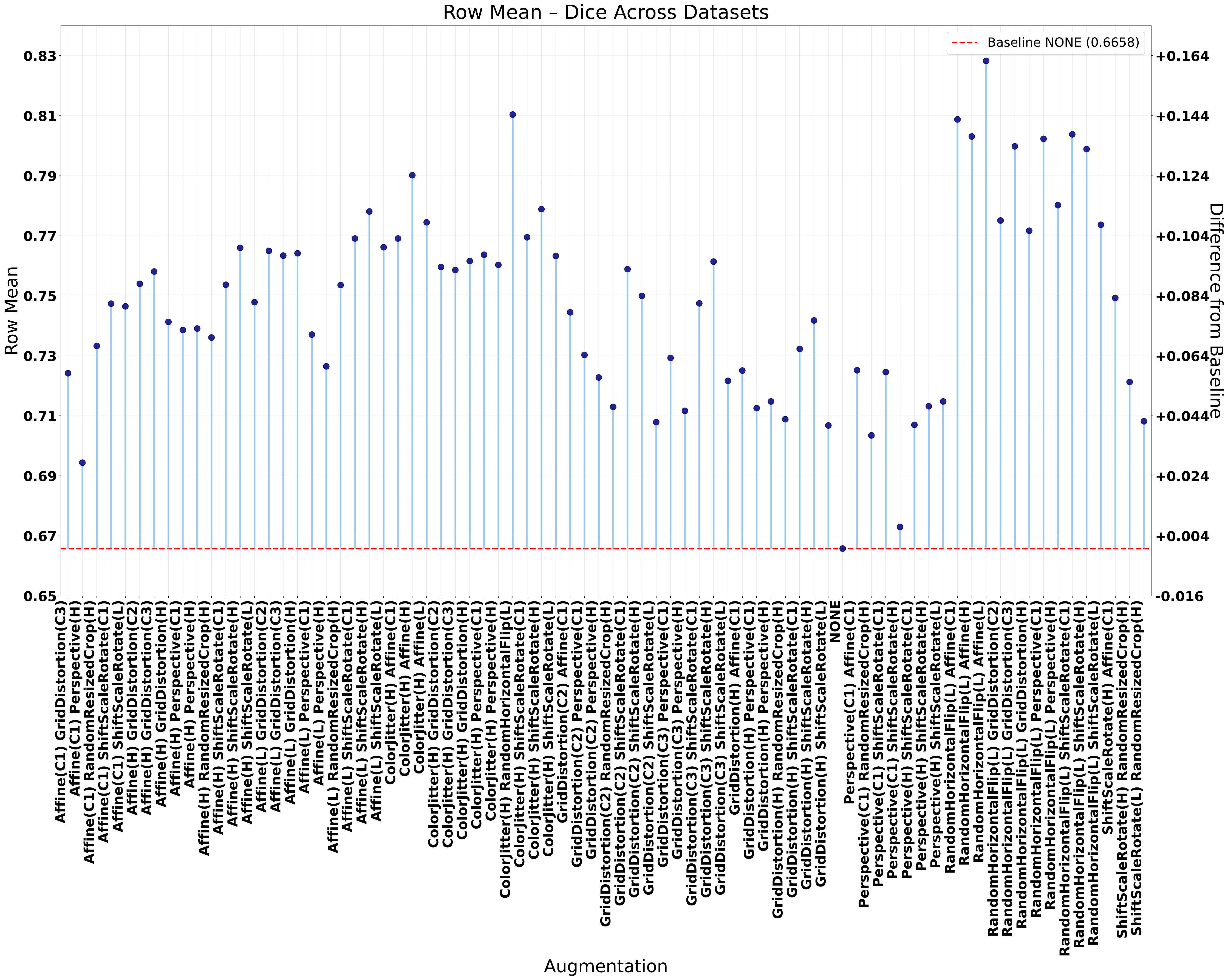}
\caption{
Row-Mean Dice Performance Across Datasets and Relative Improvements Over the NONE for Pairwise Augmentations
}
\label{fig:example}
\end{figure}

\begin{figure}[ht]
\centering
\includegraphics[width=1\linewidth]{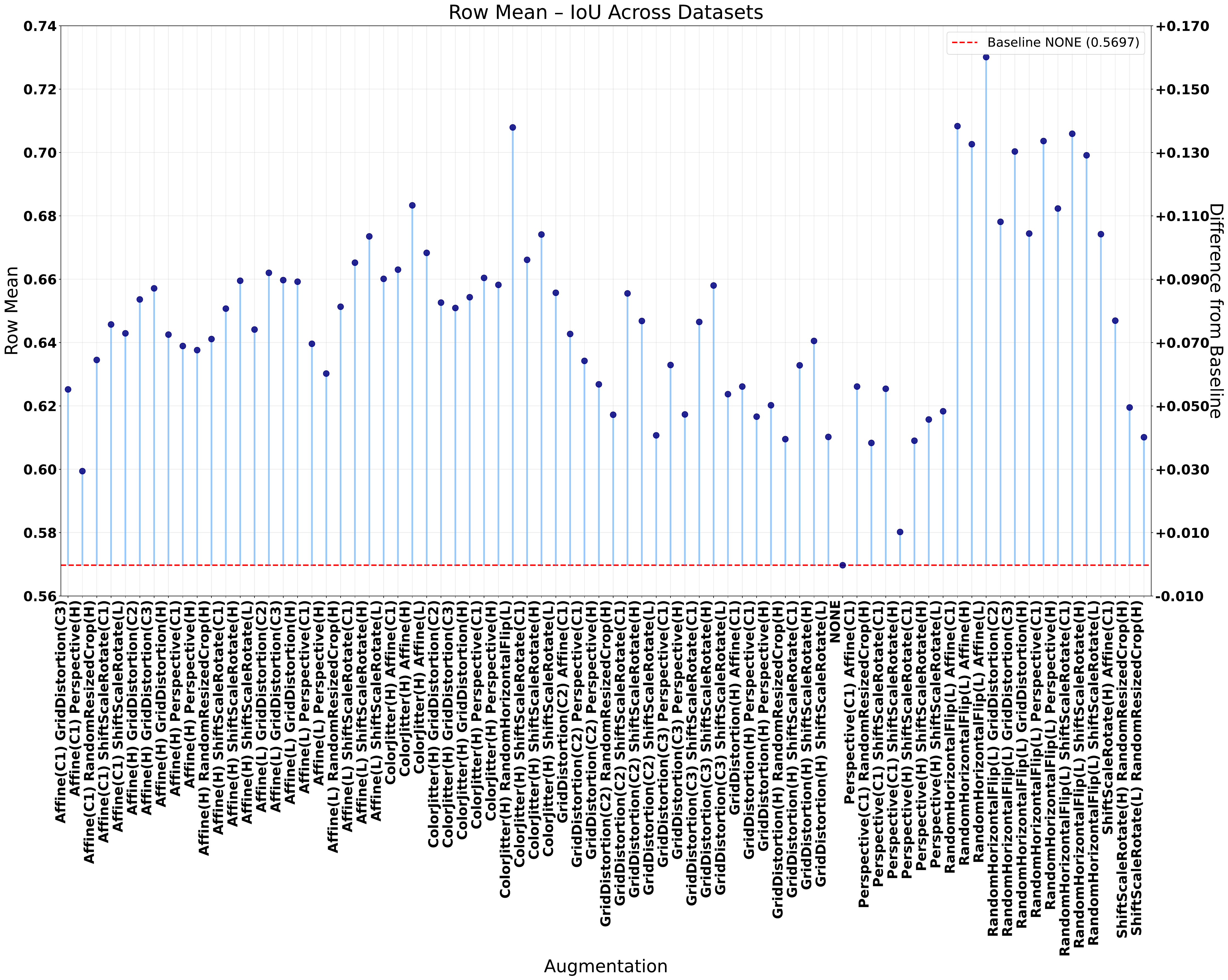}
\caption{
Row-Mean IoU Performance Across Datasets and Relative Improvements Over the NONE for Pairwise Augmentations
}
\label{fig:example}
\end{figure}

\section{Discussion}
Based on the results, in-domain accuracy was near-saturated and largely insensitive to augmentation, whereas cross-dataset performance varied substantially. The augmentations that most improved generalisation were specific geometric augmentations, rather than the photometric or ultrasound-specific transformations that the literature has emphasised; moreover, combining augmentations produced further gains, with a random horizontal flip paired with a low-magnitude affine transform achieving the strongest overall robustness. The same broad ranking was obtained under the IoU metric, indicating that the conclusions are not an artefact of Dice alone. Because IoU penalises over- and under-segmentation more strongly, this agreement suggests that the best augmentations improved core overlap quality rather than only soft boundary agreement.

\subsection{Augmentation acts on generalisation, not on within-dataset accuracy}
Within a single imaging domain, the model is already near saturation, and augmentation has almost nothing to add. Across datasets the picture is entirely different, with the baseline row mean Dice falling to a mean of 0.534 and the augmentations spanning roughly 0.49 to 0.70. The practical consequence is that augmentation studies confined to one dataset, as many echocardiographic studies have done \cite{saeed_contrastive_2022,holste_efficient_2024,guo_improved_2023}, cannot reliably discriminate between augmentations because the headroom only opens under the scanner, vendor, and protocol differences that define clinical deployment\cite{guo_impact_2024,kushol_dsmri_2023,yan_mri_2020} and that are evident in the intensity statistics of Figure 4. The aggregated score should therefore be read as a measure of robustness under domain shift rather than of raw accuracy, and it is a deliberately coarse summary that averages nine heterogeneous transfer settings, three of which are near-constant in-domain values.

\subsection{A compact set of geometric operations carries the benefit, and these operations tolerate strong settings}
The augmentations that most improved generalisation were geometry-based, and the effect belongs to particular operations rather than to a whole category. Horizontal flipping was the strongest single augmentation (mean Dice 0.776), ahead of affine (0.754), shift scale rotate(0.739), and perspective (0.723) warping, consistent with the dominance of flips and rotations in ultrasound practice\cite{tupper_revisiting_2025} and with earlier geometric augmentation for echocardiographic LV analysis\cite{sfakianakis_gudu_2023,tabuco_two-view_2022, madani_fast_2018}. The geometric group is internally uneven. Its cropping members performed near baseline, so the geometric average (0.718) is not even the highest of the five groups, the two occlusion augmentations averaging slightly more (0.724), and a strong colour-jitter setting and random erasing sit among the geometric leaders. The defensible and more useful claim is that reflection and affine or shift scale rotatewarping are the reliable levers, whereas cropping within the same nominal class is not.

The chosen magnitudes support a clear interpretation about how strong an augmentation should be. Each transform was swept across three to five configurations, guided by previous studies and our preliminary experiments, with parameter ranges bounded to keep the ventricle plausible and in frame, and the best configuration was retained on the combined criterion of mean Dice and statistical significance, so the settings were selected from evidence rather than assumed. The geometric winners were kept at moderate to strong magnitudes, with affine reaching rotations of thirty degrees and scaling between 0.6 and 1.5 times, and shift scale rotatereaching shifts of a quarter of the image and rotations of thirty-five degrees. Intensity augmentations show the opposite pattern, helping only at their mildest settings and harming the model at full strength, because the strong random brightness and contrast setting (limits of 0.4) became the worst technique overall and degraded even in-domain accuracy. Geometric perturbation can be pushed hard, since enlarging the range of poses and shapes adds useful variation without destroying the signal, whereas intensity perturbation must stay gentle, because it quickly overwrites the greyscale relationships that separate myocardium from the blood pool.

\subsection{Physical realism of augmentation and its relationship with performance}
The most important interpretive result is that physical plausibility alone was not a reliable predictor of augmentation effectiveness, and three contrasting patterns make this concrete. The geometric winners are anatomically realistic, since rotation, shifting, scaling, and mild perspective reproduce genuine acquisition variability arising from probe angulation, transducer pressure, patient size, and depth or zoom settings, so a model made invariant to them becomes robust to variation introduced during clinical acquisition. 

Deep learning models are highly effective pattern-recognition systems, particularly modern architectures such as U-Net. However, this capability can act as a double-edged sword. On one hand, these models can learn complex and useful representations when sufficient data are available. On the other hand, they may also exploit spurious correlations in the training distribution, focusing on dataset-specific image characteristics rather than the anatomical structures or clinically meaningful features that experts rely on, which may partly explain why segmentation models degrade under domain shift. Therefore, making the learning task more challenging, or reducing the association between such nuisance characteristics and the target label, can encourage the model to focus more strongly on the true region of interest.

This may explain why some mild and physically realistic augmentations are not always sufficient to improve robustness, whereas stronger transformations can sometimes provide a more effective regularising effect. One example is horizontal flipping, which produced an unexpected but strong benefit in our experiments. From a physical perspective, mirroring an apical four-chamber echocardiographic image reverses the left-right arrangement of the cardiac chambers and creates a view that is normally uncommon in standard clinical acquisition. Nevertheless, it improved performance more than any other augmentation. As supported by our Grad-CAM analysis in Appendix B, horizontal flipping may disrupt spurious associations between the left ventricle and its fixed position within the image frame, forcing the model to rely more on the internal echocardiographic fan-shaped region, the LV cavity, and relevant structural features such as shape and wall boundaries, rather than superficial positional cues.

These findings suggest that an augmentation does not necessarily need to be fully physically realistic to be useful; rather, it should preserve the semantic validity of the label while breaking nuisance correlations in the data. However, this interpretation should be treated with caution. Further investigation is needed to determine how such non-physiological augmentations can be used alongside more anatomically plausible transformations. Therefore, these results should be viewed as evidence for further experimental study rather than as direct support for unrestricted use in real-world clinical deployment. The key point is that, although some studies avoid such augmentations due to concerns about physical realism, they remain valuable for experimental investigation. Studying them can provide insight into the shortcut learning and domain-specific representation learning of deep learning models and help identify mechanisms for improving model robustness.

The ultrasound-specific augmentations complete the argument from the opposite direction. Although they are the most faithful to imaging physics, they were among the least useful, and the group produced the lowest mean of the five categories (0.68). This should not be interpreted as meaning that modality-aware augmentation is ineffective, since depth attenuation was in fact the most consistent augmentation of that group, reaching significance in six of nine settings. Therefore, these findings should motivate further investigation into how such transformations work, with the goal of adjusting and developing echocardiography-specific augmentations. Faithfully simulating depth attenuation, acoustic shadowing, haze, and speckle instead adds variation along axes that are not the dominant source of disagreement between these datasets, which is largely geometric and partly resolution related, so the simulated realism may not target the dominant sources of domain shift in these datasets. 

\subsection{Pairwise augmentation combinations add further robustness}
Pairwise augmentation was not simply an extension of the single-augmentation ranking. Instead, it addressed a different question, namely whether two transformations complement each other, duplicate the same effect, or interact negatively. This distinction is important because some augmentations that were useful in isolation became less effective when combined with similar spatial warps, whereas flip-based combinations repeatedly acted as stable augmentation hubs. Pairwise testing therefore revealed synergy, redundancy, and unsafe interactions that would not have been visible from individual augmentation experiments alone.

Combining two operations that cover different sources of variation gave the best results of the study, while combining similar ones did not. A horizontal flip combined with a mild affine transform reached a mean Dice of 0.8283. This improved on the best single augmentation by about 6.75\%, from 0.7759 to 0.8283, and on the baseline by about 24.41\%, from 0.6658 to 0.8283. Nine of the ten strongest pairs contained the flip, marking it as a broadly compatible base operation. However, the optimal partner for the flip was not identical across all source domains. For Unity-trained models, perspective-based pairings were particularly effective, suggesting that Unity transfer benefits strongly from augmentations that simulate probe-angle and viewpoint variation. Pairing two spatial warps that act on the same axis, such as perspective with shift-scale-rotate, gave limited gain and at stronger settings fell as low as 0.673, below either operation applied alone. The benefit therefore comes from complementarity rather than from accumulating transforms, and stacking similar warps is redundant and risks compounding into implausible deformations. This evidence is bounded by the design, since the pairs were built only from the strongest individual augmentations, which were mostly geometric, so the prominence of geometric pairs partly reflects that selection, and the result establishes complementarity within a curated set rather than across augmentations in general.

\subsection{The benefit is largest for small, single-vendor training data, and part of the gap is resolution}
Augmentation provided the greatest benefit in the setting where the training data were most limited. This effect was especially clear for CA, where augmentation did not merely provide a marginal improvement but changed some external transfers from near failure (Dice score of 0.19 and 0.23) to more usable segmentation performance (Dice score of 0.73 and 0.62 with the combined horizontal-flip and affine). In this setting, augmentation choice became a decisive factor in whether cross-dataset transfer succeeded at all, supporting the view that small, single-vendor datasets require stronger augmentation policies than larger and more heterogeneous sources. In contrast, models trained on the larger and more heterogeneous ED dataset already generalised reasonably well without augmentation, leaving less scope for further improvement. This suggests that augmentation can partially compensate for limited data diversity and is particularly valuable when the training set is small, homogeneous, or acquired under narrow imaging conditions. This interpretation is consistent with previous evidence indicating that robustness is mainly improved through increased diversity, whether obtained from additional data, generative augmentation, or large-scale pretraining \cite{tupper_revisiting_2025, kim_echofm_2025, van_de_vyver_generative_2025}.

The strong performance of simulated downscaling, which was the best-performing noise-related augmentation with a mean Dice score of 0.718, also identifies a specific and potentially addressable component of the domain shift. ED images are stored at 112×112 pixels and then upscaled to 512×512 pixels, making them intrinsically softer than the natively higher-resolution Unity and CA images. 

\subsection{Key messages, reliability, and clinical relevance}
For model development, the practical implication is clear. Horizontal flipping combined with a mild affine or shift scale rotatetransformation captures much of the achievable cross-dataset robustness for left ventricular segmentation at negligible computational cost, without requiring artefact simulation or large-scale pretraining. Augmentations should therefore be selected not only by average gain but also by consistency across settings. These criteria can diverge, as strong colour jitter ranked fourth by mean performance but was significant in only two of nine settings, whereas centre cropping was significant in seven of nine despite a lower mean. Methods that perform well on both criteria, such as affine warping, which was significant in eight of nine settings, are therefore safer choices. A stricter way to define safety is to exclude the near-saturated in-domain evaluations and require an augmentation pair to improve all six cross-dataset transfers. Under this no-harm criterion, the most reliable families were again centred on controlled geometric transformations, particularly flip-, affine-, shift-scale-rotate-, and grid-distortion-based combinations. This supports a practical recommendation based on cross-domain consistency rather than on the single highest row-mean score. Conversely, augmentations that improve some transfers while harming others, as observed for strong brightness and contrast adjustment, may be counterproductive despite favourable significance counts. This argues against ranking augmentations by significance tallies alone.

Clinically, the encouraging finding is that single-centre models can be made substantially more reliable on external scanners using simple, anatomically motivated augmentation, reducing but not removing the need for site-specific retraining and re-annotation. Because the most effective operations reflect realistic acquisition variation, the gained robustness is potentially clinically relevant, although prospective validation would be required before clinical deployment. However, even the strongest policy left the weakest transfer performances at approximately 0.62-0.73 Dice, below the in-domain ceiling of around 0.93, which could still produce clinically relevant errors in derived measures such as ejection fraction. Augmentation should therefore be viewed as a necessary and highly cost-effective first step towards cross-vendor reliability, enabling active or self-supervised learning through models with stronger generalisation and transferability.

\section{Limitations and Future Work }
This study presents a large-scale evaluation of data augmentation strategies for echocardiography images, with the aim of improving the generalisability of deep learning models. Although considerable effort was made to minimise experimental limitations, the scale of this study inevitably introduces aspects that warrant further investigation. These are highlighted here as directions for future work.

For each augmentation technique, three to five parameter settings were evaluated. An exhaustive exploration of all hyperparameter combinations is impractical due to the vast search space and the substantial computational resources required. Consequently, not all possible configurations could be examined.

In the augmentation combination experiments, only selected augmentations were paired and evaluated in a pairwise manner. Extending this analysis to higher-order combinations would lead to a combinatorial explosion and is infeasible given the associated computational cost. This limitation should therefore be considered when interpreting the results. Moreover, in most self-supervised learning frameworks, a pair of augmentation transformations is commonly selected to generate different views of the same sample \cite{chen_simple_2020,he_momentum_2020, grill_bootstrap_2020}; accordingly, we limit our analysis to such pairwise combinations, as extending beyond this would incur substantially higher computational costs.

Moreover, this study focused exclusively on the U-Net architecture, which remains one of the most widely adopted models in medical image segmentation. Future work may extend this evaluation to alternative architectures, such as DeepLabV3+, as well as to classification tasks, in order to assess whether the observed augmentation effects are architecture-dependent or whether more generalisable augmentation strategies can be identified for echocardiography applications.

Furthermore, the cross-dataset improvements show that the results suggest that the effective augmentations identified here could be used in self-supervised and active-learning studies, where they can lead to better results for algorithms such as contrastive learning.

A final point to consider is the need to prioritise echocardiography-specific augmentations that meaningfully target anatomical structure rather than relying solely on generic image transformations. Such targeted strategies can encourage the model to learn robust, anatomy-driven representations. At the same time, they highlight the importance of developing appropriate metrics to better characterise domain shift and to guide the design of principled approaches that effectively bridge distribution gaps between datasets. As future work, these augmentation strategies and domain-shift quantification metrics should be investigated further, standardised, and systematically evaluated, enabling a more principled framework for anatomy-aware generalisation in echocardiographic segmentation. Investigating the impact of cropping, padding, and different resizing strategies on generalisation performance is another promising direction for future work, as suggested by our internal experiments.

\section{Conclusion }
This study addresses a persistent challenge in echocardiographic image analysis, namely the poor generalisability of segmentation models across datasets acquired in different clinical environments. Through a large-scale systematic evaluation of 29 augmentation techniques and their pairwise combinations across three datasets, we show that data augmentation is a decisive and low-cost strategy for mitigating domain shift when selected deliberately rather than by convention.

Three main findings emerge. First, augmentation has little effect on in-domain accuracy, which is already near saturation, but substantially improves cross-dataset transfer. Augmentation policies should therefore be judged by robustness under domain shift rather than by within-dataset gains. Second, a compact set of geometric operations, including random horizontal flip, affine transformation, shift-scale-rotate, and perspective, provides the strongest and most statistically consistent improvements, whereas many commonly used intensity- and artefact-based transforms offer limited benefit or even degrade performance. Physical plausibility alone did not predict effectiveness. Horizontal flipping, despite producing an anatomically uncommon mirrored view, was the most beneficial single augmentation, likely because it disrupts spurious associations between the left ventricle and its fixed position in the image frame. Third, well-matched pairwise combinations outperformed individual augmentations. Combining horizontal flip with a mild affine transform increased the overall row-mean Dice from 0.67 to 0.83, representing a 24\% relative gain over the no-augmentation baseline, and converted several weak cross-dataset transfers into usable segmentations.

For practitioners, the recommendation is practical and inexpensive. Pairing horizontal flip with a mild affine or shift scale rotatetransform captures most of the achievable cross-vendor robustness for left ventricular segmentation without artefact simulation or large-scale pretraining. Augmentations should also be selected for consistency across transfer settings rather than peak average performance alone. These gains are clinically encouraging but not sufficient in isolation, as even the strongest policy left the most difficult transfers at approximately 0.62 to 0.73 Dice, below the in-domain ceiling of about 0.93. Prospective validation therefore remains necessary before deployment. By identifying augmentation families that reliably improve transferability, this work provides practical guidance for data-limited multi-institutional settings and establishes a foundation for future self-supervised and annotation-efficient learning in echocardiography.

\section*{Acknowledgements}

The authors express their sincere appreciation to Arya Varastehnezad for his valuable assistance with the image design. The authors also thank Professor Julie Wall for her helpful comments and constructive feedback.

\section*{Funding}

This work was supported in part by the British Heart Foundation (grant no. RG/F/22/110059) and the Vice-Chancellor’s Scholarship at the University of West London.

\section*{Ethics approval and consent to participate}

The South Central-Oxford C Research Ethics Committee gave a favourable ethical opinion for the release of the anonymised UnityLV-MultiX echocardiography dataset (Integrated Research Application System ID: 279328; REC reference: 20/SC/0386). The echocardiograms were collected during routine clinical examinations and individual patient consent was not required because the data were originally acquired for clinical use and anonymised before research release. This study used anonymised secondary echocardiography datasets and did not involve the collection of new human participant data by the authors. CAMUS and EchoNet-Dynamic are publicly available research datasets and were used according to their original data-use conditions. For this study, the ethic code is UWL/UREC/HR-00445.

\section*{Data availability}

CAMUS and EchoNet Dynamic are publicly available from their original sources. Unity and Consensus data are not publicly available due to institutional data-use restrictions, but may be made available from the corresponding author upon reasonable request and subject to the necessary institutional approvals and data-use agreements. 

\section*{Competing interests}

The authors declare no competing interests.



\bibliography{report} 
\bibliographystyle{spiebib} 

\newgeometry{left=1cm,right=1cm,top=1.5cm,bottom=1.5cm}
\clearpage

\appendix    

\section{Echo-specific fan-shape masking}
Some of the targeted augmentations (including depth attenuation, Gaussian-shadow, haze artifact, and speckle reduction) require fan-shaped masks for their application, as specified in the original works that introduced these methods \cite{tupper_revisiting_2025}. This is to avoid unrealistic artefacts caused by applying geometric or intensity transformations to the black background region outside the ultrasound sector, echo-specific pre-processing based on fan-shaped masks was employed. In that study, approximate fan-shaped scan masks were generated for each dataset using morphological image operations. Therefore, we need a means to extract these fan-shaped masks. However, this approach is mainly suitable for datasets with clear fan-shaped boundaries, as illustrated by the CA dataset in Figure 2. In less clearly defined datasets, such as Unity and ED, the sector boundaries are more difficult to detect reliably using only morphological operations. To solve this issue, we need another approach that we employed a CNN-based approach. In this method, fan-shaped regions were estimated by employing a U-Net architecture implemented using a ResNet-50 encoder initialised with ImageNet pre-trained weights. The network was trained on single-channel (greyscale) inputs resized to 512×512 pixels, producing a binary segmentation mask. Training was performed with a batch size of eight for a maximum of 200 epochs. Optimisation was carried out using the AdamW optimiser with a learning rate of $1\text{x}10^{-4}$ and a weight decay of $1\text{x}10^{-4}$. The loss function was a combined Binary Cross-Entropy and Dice loss, equally weighted, to balance region overlap and pixel-wise accuracy. The adaptation of the learning rate was handled via ReduceLROnPlateau, monitoring the validation loss with a patience of two epochs and a reduction factor of 0.5. Early stopping was applied with a patience of five epochs to prevent overfitting. The best model was selected based on the lowest validation loss, and performance was monitored using the Dice similarity coefficient.

\begin{figure}[H]
\centering
\includegraphics[width=0.9\linewidth]{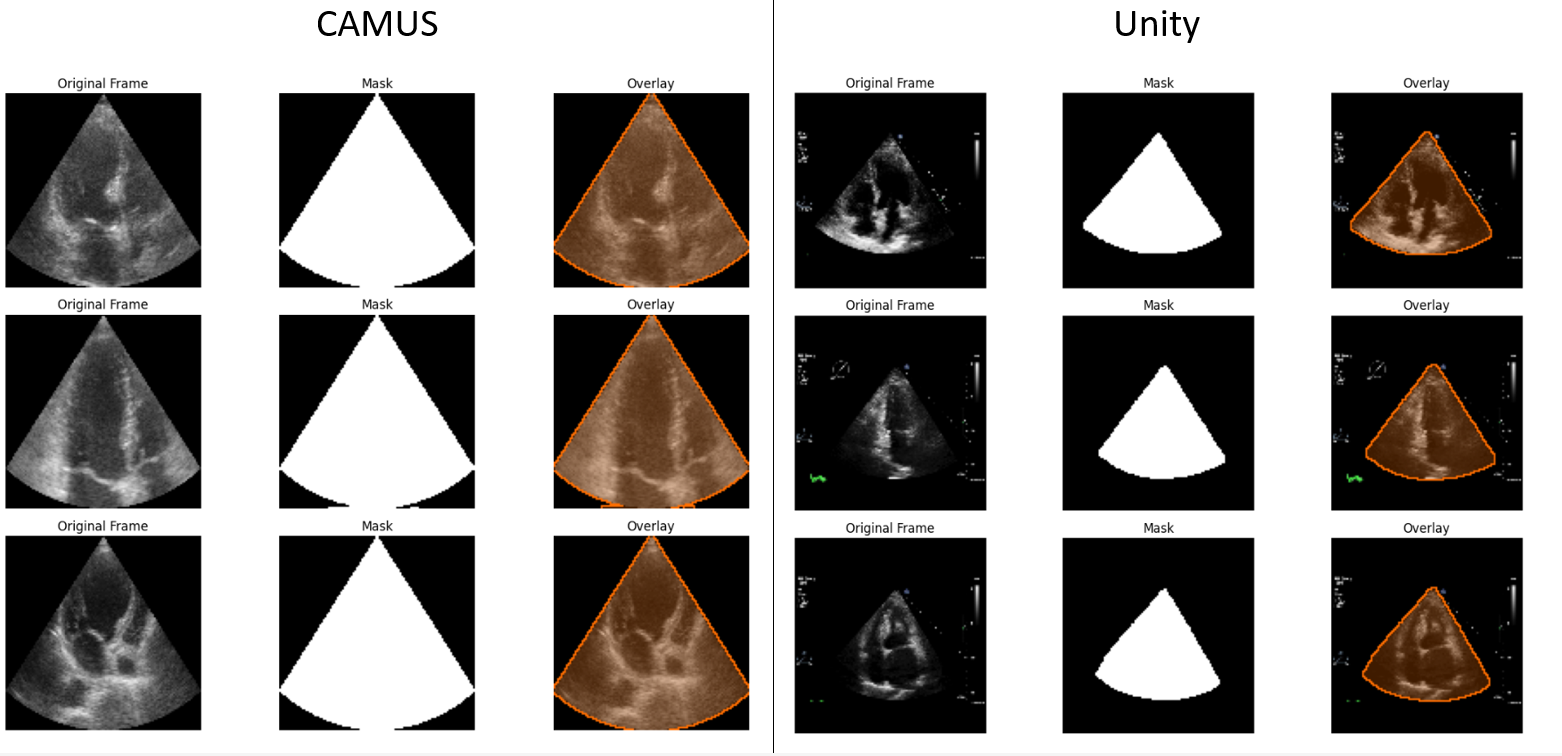}
\caption{
Examples of our CNN-based fan-shaped ultrasound sector mask generation
}
\label{fig:example}
\end{figure}

The resulting model was used to segment the ultrasound cone on small manually annotated subsets of images, which included 25 frames from Unity (20 for training, five for validation), 33 frames from CA (27 for training, six for validation), and 30 images from ED (24 for training, six for validation). Using a pre-trained model is important in this case, as its prior learning of low-level visual features, such as edges and boundaries, can significantly reduce the amount of labelled data required for transfer to a new dataset. Example outputs are shown in Figure 14.

\section{Grad-CAM}

In this section, we compare the Grad-CAM  results obtained without augmentation (None) to those achieved using the best-performing augmentation pair, random horizontal flip (L) combined with affine (L). In each figure, the first image represents the original input image. The next three Grad-CAM  images correspond to models trained without augmentation, in the following order: Unity, CA, and ED. These are followed by three Grad-CAM  results generated using random horizontal flip (L) combined with affine (L), presented in the same order.

Accordingly, each no-augmentation Grad-CAM  result should be compared with the corresponding augmentation-based result located three positions after it. For example, the first Grad-CAM  image (no augmentation, trained on Unity) should be compared with the fourth Grad-CAM image (random horizontal flip (L) combined with affine (L), also trained on Unity). The images shown in Figure 15 are derived from the CS dataset, while the images in Figure 16 are from the CA dataset (first two rows) and ED (last row).

\begin{figure}[H]
\centering

\begin{subfigure}{\linewidth}
    \centering
    \includegraphics[width=\linewidth]{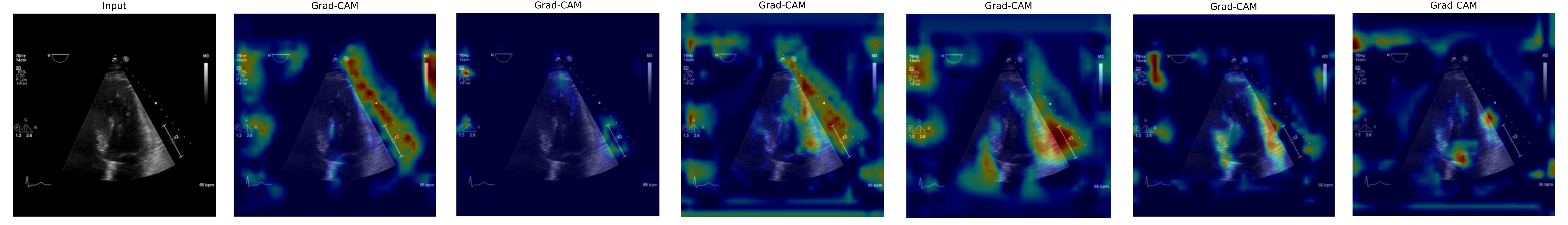}
\end{subfigure}

\vspace{-3mm}

\begin{subfigure}{\linewidth}
    \centering
    \includegraphics[width=\linewidth]{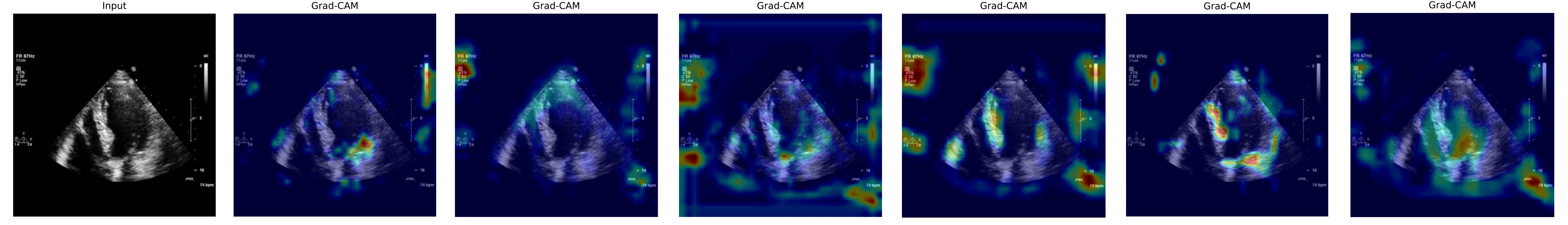}
\end{subfigure}

\vspace{-3mm}

\begin{subfigure}{\linewidth}
    \centering
    \includegraphics[width=\linewidth]{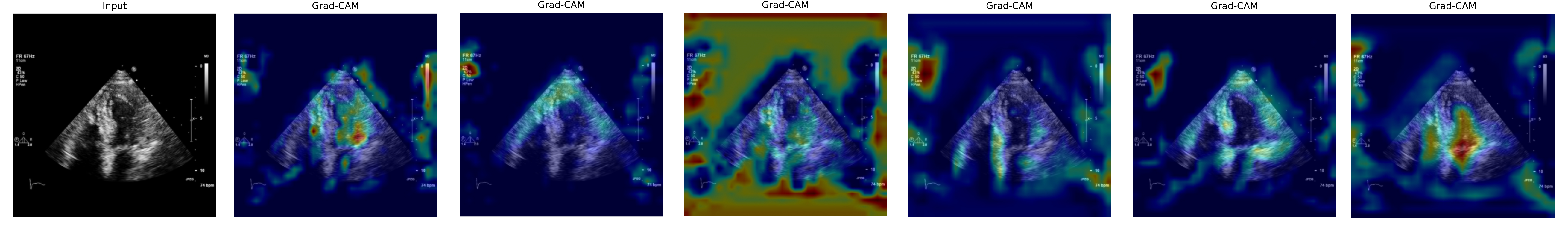}
\end{subfigure}

\caption{Grad-CAM results on CS images}
\label{fig:example}
\end{figure}

\begin{figure}[H]
\centering

\begin{subfigure}{\linewidth}
    \centering
    \includegraphics[width=\linewidth]{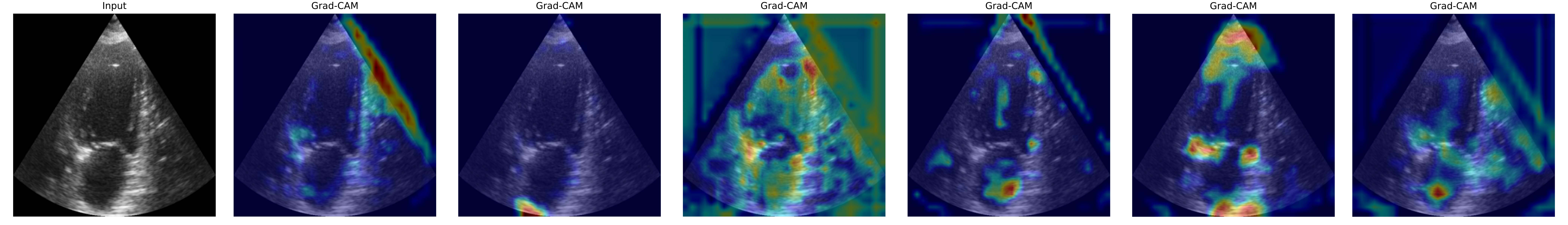}
\end{subfigure}

\vspace{-3mm}

\begin{subfigure}{\linewidth}
    \centering
    \includegraphics[width=\linewidth]{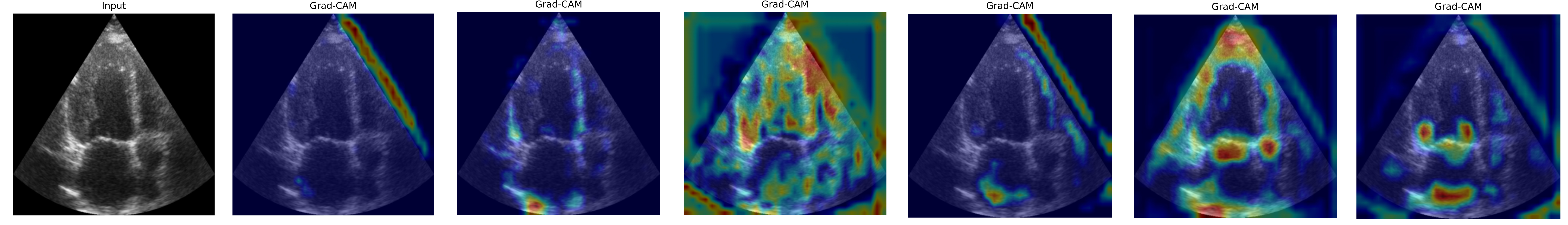}
\end{subfigure}

\vspace{-3mm}

\begin{subfigure}{\linewidth}
    \centering
    \includegraphics[width=\linewidth]{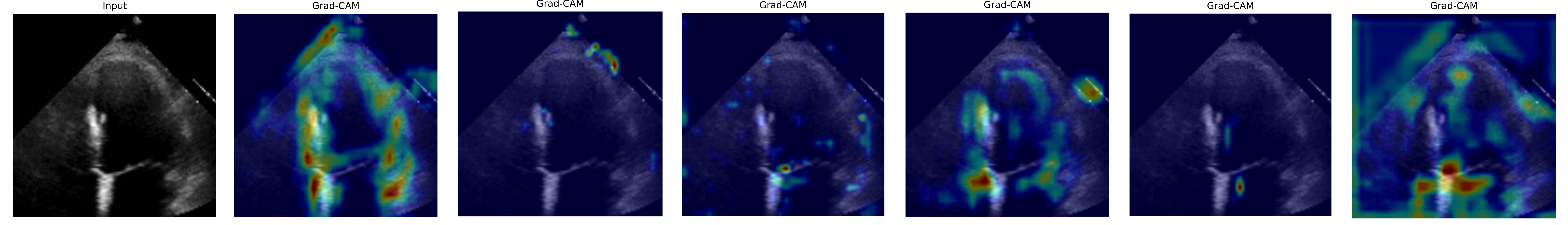}
\end{subfigure}

\caption{Grad-CAM results on CA and ED images}
\label{fig:example}
\end{figure}

\section{Dice Results in detail}

Tables 3 (individual augmentations) and 4 (pairwise augmentation combinations) present the Dice score results in detail. Green cells indicate statistically significant improvements, whereas red cells denote non-significant results. Furthermore, the row-wise mean Dice score and the number of statistically significant outcomes are reported as auxiliary columns to facilitate comparison across different augmentation strategies. In addition to these tables, heatmaps showing detailed results of Phase 2 (pairwise combinations) are presented in Figures 17 and 18.

\definecolor{darkgreen}{RGB}{0,100,0}
\definecolor{darkred}{RGB}{139,0,0}

\begin{table}[ht]
\centering
\caption{Dice Scores and Statistical Significance of Data Augmentation Methods}
\label{tab:game_personality}
\footnotesize
\setlength{\tabcolsep}{1.3pt}
\renewcommand{\arraystretch}{1.2}

\begin{tabular}{|l|l|ccc|ccc|ccc|c|c|}
\hline
\textbf{Augmentation} & \textbf{Setting}
& \multicolumn{3}{c|}{\textbf{Unity}}
& \multicolumn{3}{c|}{\textbf{CAMUS}}
& \multicolumn{3}{c|}{\textbf{EchoNet}}
& \textbf{Row Avg}
& \textbf{Sig 0.05} \\

\cline{3-11}
& 
& CS & CA & ED
& CS & CA & ED
& CS & CA & ED
& & \\ 
\hline

RandomHorizontalFlip & L & \textcolor{darkred}{\textbf{0.9338}} & \textcolor{darkgreen}{\textbf{0.7377}} & \textcolor{darkred}{\textbf{0.7639}} & \textcolor{darkgreen}{\textbf{0.6007}} & \textcolor{darkred}{\textbf{0.9346}} & \textcolor{darkgreen}{\textbf{0.4948}} & \textcolor{darkred}{\textbf{0.7104}} & \textcolor{darkgreen}{\textbf{0.8834}} & \textcolor{darkred}{\textbf{0.9237}} & 0.7759 & 4/9 \\ \hline
Affine & H & \textcolor{darkgreen}{\textbf{0.9364}} & \textcolor{darkgreen}{\textbf{0.7337}} & \textcolor{darkgreen}{\textbf{0.7926}} & \textcolor{darkgreen}{\textbf{0.4613}} & \textcolor{darkgreen}{\textbf{0.9380}} & \textcolor{darkgreen}{\textbf{0.4670}} & \textcolor{darkgreen}{\textbf{0.8180}} & \textcolor{darkred}{\textbf{0.7092}} & \textcolor{darkgreen}{\textbf{0.9263}} & 0.7536 & 8/9 \\ \hline
ShiftScaleRotate & C1 & \textcolor{darkred}{\textbf{0.9360}} & \textcolor{darkred}{\textbf{0.7020}} & \textcolor{darkgreen}{\textbf{0.8082}} & \textcolor{darkgreen}{\textbf{0.4771}} & \textcolor{darkgreen}{\textbf{0.9378}} & \textcolor{darkgreen}{\textbf{0.4317}} & \textcolor{darkred}{\textbf{0.7489}} & \textcolor{darkred}{\textbf{0.6867}} & \textcolor{darkgreen}{\textbf{0.9259}} & 0.7394 & 5/9 \\ \hline
ColorJitter & H & \textcolor{darkred}{\textbf{0.9326}} & \textcolor{darkred}{\textbf{0.6159}} & \textcolor{darkred}{\textbf{0.7748}} & \textcolor{darkgreen}{\textbf{0.4009}} & \textcolor{darkred}{\textbf{0.9311}} & \textcolor{darkgreen}{\textbf{0.5821}} & \textcolor{darkred}{\textbf{0.7116}} & \textcolor{darkred}{\textbf{0.7277}} & \textcolor{darkred}{\textbf{0.9248}} & 0.7335 & 2/9 \\ \hline
RandomErasing & H & \textcolor{darkgreen}{\textbf{0.9363}} & \textcolor{darkred}{\textbf{0.6415}} & \textcolor{darkgreen}{\textbf{0.7906}} & \textcolor{darkred}{\textbf{0.3928}} & \textcolor{darkred}{\textbf{0.9358}} & \textcolor{darkgreen}{\textbf{0.4341}} & \textcolor{darkred}{\textbf{0.7463}} & \textcolor{darkgreen}{\textbf{0.7469}} & \textcolor{darkred}{\textbf{0.9250}} & 0.7277 & 4/9 \\ \hline
Perspective & C1 & \textcolor{darkgreen}{\textbf{0.9344}} & \textcolor{darkgreen}{\textbf{0.8028}} & \textcolor{darkgreen}{\textbf{0.8339}} & \textcolor{darkred}{\textbf{0.3069}} & \textcolor{darkred}{\textbf{0.9371}} & \textcolor{darkgreen}{\textbf{0.3822}} & \textcolor{darkred}{\textbf{0.6566}} & \textcolor{darkred}{\textbf{0.7273}} & \textcolor{darkgreen}{\textbf{0.9268}} & 0.7231 & 5/9 \\ \hline
CoarseDropout & C2 & \textcolor{darkred}{\textbf{0.9323}} & \textcolor{darkred}{\textbf{0.6172}} & \textcolor{darkred}{\textbf{0.7744}} & \textcolor{darkgreen}{\textbf{0.4777}} & \textcolor{darkred}{\textbf{0.9339}} & \textcolor{darkgreen}{\textbf{0.3810}} & \textcolor{darkred}{\textbf{0.7441}} & \textcolor{darkred}{\textbf{0.6936}} & \textcolor{darkred}{\textbf{0.9243}} & 0.7198 & 2/9 \\ \hline
Downscale & H & \textcolor{darkred}{\textbf{0.9331}} & \textcolor{darkred}{\textbf{0.6441}} & \textcolor{darkred}{\textbf{0.7737}} & \textcolor{darkgreen}{\textbf{0.4116}} & \textcolor{darkred}{\textbf{0.9329}} & \textcolor{darkred}{\textbf{0.3422}} & \textcolor{darkgreen}{\textbf{0.8008}} & \textcolor{darkgreen}{\textbf{0.6982}} & \textcolor{darkred}{\textbf{0.9244}} & 0.7179 & 3/9 \\ \hline
DepthAttenuation & C2 & \textcolor{darkgreen}{\textbf{0.9324}} & \textcolor{darkgreen}{\textbf{0.6407}} & \textcolor{darkgreen}{\textbf{0.7628}} & \textcolor{darkred}{\textbf{0.3025}} & \textcolor{darkgreen}{\textbf{0.9229}} & \textcolor{darkgreen}{\textbf{0.4891}} & \textcolor{darkred}{\textbf{0.7819}} & \textcolor{darkred}{\textbf{0.6492}} & \textcolor{darkgreen}{\textbf{0.9219}} & 0.7115 & 6/9 \\ \hline
GridDistortion & C3 & \textcolor{darkred}{\textbf{0.9342}} & \textcolor{darkred}{\textbf{0.6929}} & \textcolor{darkgreen}{\textbf{0.8037}} & \textcolor{darkgreen}{\textbf{0.2893}} & \textcolor{darkgreen}{\textbf{0.9369}} & \textcolor{darkgreen}{\textbf{0.3347}} & \textcolor{darkgreen}{\textbf{0.8134}} & \textcolor{darkred}{\textbf{0.6661}} & \textcolor{darkred}{\textbf{0.9254}} & 0.7107 & 5/9 \\ \hline
GaussNoise & C2 & \textcolor{darkred}{\textbf{0.9341}} & \textcolor{darkred}{\textbf{0.5773}} & \textcolor{darkgreen}{\textbf{0.7856}} & \textcolor{darkgreen}{\textbf{0.3288}} & \textcolor{darkgreen}{\textbf{0.9289}} & \textcolor{darkgreen}{\textbf{0.4877}} & \textcolor{darkred}{\textbf{0.7544}} & \textcolor{darkred}{\textbf{0.6548}} & \textcolor{darkgreen}{\textbf{0.9256}} & 0.7086 & 5/9 \\ \hline
GaussianBlur & L & \textcolor{darkred}{\textbf{0.9317}} & \textcolor{darkgreen}{\textbf{0.6833}} & \textcolor{darkgreen}{\textbf{0.7841}} & \textcolor{darkred}{\textbf{0.3305}} & \textcolor{darkred}{\textbf{0.9336}} & \textcolor{darkgreen}{\textbf{0.3741}} & \textcolor{darkred}{\textbf{0.7403}} & \textcolor{darkred}{\textbf{0.6684}} & \textcolor{darkred}{\textbf{0.9235}} & 0.7077 & 3/9 \\ \hline
UnsharpMask & harsh & \textcolor{darkred}{\textbf{0.9331}} & \textcolor{darkred}{\textbf{0.6220}} & \textcolor{darkred}{\textbf{0.7598}} & \textcolor{darkred}{\textbf{0.2892}} & \textcolor{darkred}{\textbf{0.9326}} & \textcolor{darkgreen}{\textbf{0.3754}} & \textcolor{darkgreen}{\textbf{0.7916}} & \textcolor{darkgreen}{\textbf{0.7319}} & \textcolor{darkred}{\textbf{0.9242}} & 0.7066 & 3/9 \\ \hline
CLAHE & C1 & \textcolor{darkred}{\textbf{0.9339}} & \textcolor{darkgreen}{\textbf{0.6964}} & \textcolor{darkred}{\textbf{0.7710}} & \textcolor{darkgreen}{\textbf{0.3608}} & \textcolor{darkred}{\textbf{0.9341}} & \textcolor{darkgreen}{\textbf{0.3330}} & \textcolor{darkred}{\textbf{0.7812}} & \textcolor{darkred}{\textbf{0.6176}} & \textcolor{darkred}{\textbf{0.9244}} & 0.7058 & 3/9 \\ \hline
MotionBlur & C1 & \textcolor{darkred}{\textbf{0.9330}} & \textcolor{darkred}{\textbf{0.6458}} & \textcolor{darkgreen}{\textbf{0.7838}} & \textcolor{darkred}{\textbf{0.2926}} & \textcolor{darkred}{\textbf{0.9338}} & \textcolor{darkgreen}{\textbf{0.3473}} & \textcolor{darkred}{\textbf{0.7779}} & \textcolor{darkred}{\textbf{0.6570}} & \textcolor{darkred}{\textbf{0.9242}} & 0.6995 & 2/9 \\ \hline
MultiplicativeNoise & L & \textcolor{darkred}{\textbf{0.9338}} & \textcolor{darkred}{\textbf{0.6146}} & \textcolor{darkgreen}{\textbf{0.7704}} & \textcolor{darkred}{\textbf{0.3079}} & \textcolor{darkgreen}{\textbf{0.9333}} & \textcolor{darkred}{\textbf{0.2914}} & \textcolor{darkred}{\textbf{0.7730}} & \textcolor{darkgreen}{\textbf{0.7326}} & \textcolor{darkred}{\textbf{0.9240}} & 0.6979 & 3/9 \\ \hline
saltandpepper & C2 & \textcolor{darkgreen}{\textbf{0.9343}} & \textcolor{darkred}{\textbf{0.6093}} & \textcolor{darkred}{\textbf{0.7647}} & \textcolor{darkred}{\textbf{0.3113}} & \textcolor{darkred}{\textbf{0.9343}} & \textcolor{darkred}{\textbf{0.3006}} & \textcolor{darkgreen}{\textbf{0.8104}} & \textcolor{darkred}{\textbf{0.6756}} & \textcolor{darkgreen}{\textbf{0.9247}} & 0.6961 & 3/9 \\ \hline
ElasticTransform & H & \textcolor{darkred}{\textbf{0.9321}} & \textcolor{darkred}{\textbf{0.6538}} & \textcolor{darkred}{\textbf{0.7790}} & \textcolor{darkgreen}{\textbf{0.3376}} & \textcolor{darkred}{\textbf{0.9333}} & \textcolor{darkgreen}{\textbf{0.3274}} & \textcolor{darkred}{\textbf{0.7476}} & \textcolor{darkred}{\textbf{0.6199}} & \textcolor{darkgreen}{\textbf{0.9237}} & 0.6949 & 3/9 \\ \hline
Sharpen & L & \textcolor{darkgreen}{\textbf{0.9346}} & \textcolor{darkgreen}{\textbf{0.6537}} & \textcolor{darkgreen}{\textbf{0.7726}} & \textcolor{darkred}{\textbf{0.2409}} & \textcolor{darkred}{\textbf{0.9334}} & \textcolor{darkgreen}{\textbf{0.3252}} & \textcolor{darkred}{\textbf{0.7686}} & \textcolor{darkgreen}{\textbf{0.6979}} & \textcolor{darkred}{\textbf{0.9239}} & 0.6945 & 5/9 \\ \hline
RandomGamma & L & \textcolor{darkred}{\textbf{0.9331}} & \textcolor{darkred}{\textbf{0.6264}} & \textcolor{darkred}{\textbf{0.7677}} & \textcolor{darkgreen}{\textbf{0.3189}} & \textcolor{darkred}{\textbf{0.9329}} & \textcolor{darkgreen}{\textbf{0.3633}} & \textcolor{darkred}{\textbf{0.7455}} & \textcolor{darkred}{\textbf{0.6348}} & \textcolor{darkred}{\textbf{0.9241}} & 0.6941 & 2/9 \\ \hline
CenterCrop & C1 & \textcolor{darkgreen}{\textbf{0.9288}} & \textcolor{darkgreen}{\textbf{0.6230}} & \textcolor{darkgreen}{\textbf{0.7719}} & \textcolor{darkgreen}{\textbf{0.3472}} & \textcolor{darkgreen}{\textbf{0.9302}} & \textcolor{darkgreen}{\textbf{0.3236}} & \textcolor{darkred}{\textbf{0.7637}} & \textcolor{darkred}{\textbf{0.6303}} & \textcolor{darkgreen}{\textbf{0.9237}} & 0.6936 & 7/9 \\ \hline
ImageCompression & L & \textcolor{darkred}{\textbf{0.9331}} & \textcolor{darkred}{\textbf{0.6466}} & \textcolor{darkred}{\textbf{0.7670}} & \textcolor{darkgreen}{\textbf{0.2945}} & \textcolor{darkred}{\textbf{0.9334}} & \textcolor{darkred}{\textbf{0.3148}} & \textcolor{darkred}{\textbf{0.7253}} & \textcolor{darkred}{\textbf{0.7000}} & \textcolor{darkred}{\textbf{0.9240}} & 0.6932 & 1/9 \\ \hline
CropNonEmptyMaskIfExists & C1 & \textcolor{darkred}{\textbf{0.9339}} & \textcolor{darkred}{\textbf{0.6357}} & \textcolor{darkred}{\textbf{0.7686}} & \textcolor{darkgreen}{\textbf{0.3638}} & \textcolor{darkred}{\textbf{0.9327}} & \textcolor{darkgreen}{\textbf{0.3211}} & \textcolor{darkred}{\textbf{0.7080}} & \textcolor{darkgreen}{\textbf{0.5981}} & \textcolor{darkred}{\textbf{0.9240}} & 0.6873 & 3/9 \\ \hline
RandomResizedCrop & L & \textcolor{darkred}{\textbf{0.9299}} & \textcolor{darkred}{\textbf{0.6160}} & \textcolor{darkred}{\textbf{0.7750}} & \textcolor{darkred}{\textbf{0.3578}} & \textcolor{darkred}{\textbf{0.9297}} & \textcolor{darkred}{\textbf{0.3106}} & \textcolor{darkred}{\textbf{0.6876}} & \textcolor{darkred}{\textbf{0.5986}} & \textcolor{darkgreen}{\textbf{0.9252}} & 0.6812 & 1/9 \\ \hline
HazeArtifact & L & \textcolor{darkred}{\textbf{0.9316}} & \textcolor{darkred}{\textbf{0.5591}} & \textcolor{darkred}{\textbf{0.7794}} & \textcolor{darkred}{\textbf{0.2342}} & \textcolor{darkred}{\textbf{0.9322}} & \textcolor{darkgreen}{\textbf{0.3578}} & \textcolor{darkgreen}{\textbf{0.8011}} & \textcolor{darkred}{\textbf{0.5808}} & \textcolor{darkgreen}{\textbf{0.9164}} & 0.6770 & 3/9 \\ \hline
SpeckleReduction & C1 & \textcolor{darkred}{\textbf{0.9348}} & \textcolor{darkred}{\textbf{0.5940}} & \textcolor{darkred}{\textbf{0.7726}} & \textcolor{darkred}{\textbf{0.2820}} & \textcolor{darkred}{\textbf{0.9309}} & \textcolor{darkgreen}{\textbf{0.4437}} & \textcolor{darkred}{\textbf{0.7109}} & \textcolor{darkred}{\textbf{0.4665}} & \textcolor{darkred}{\textbf{0.9159}} & 0.6724 & 1/9 \\ \hline
IntensityWindowing & L & \textcolor{darkred}{\textbf{0.9322}} & \textcolor{darkred}{\textbf{0.6369}} & \textcolor{darkred}{\textbf{0.7643}} & \textcolor{darkgreen}{\textbf{0.1881}} & \textcolor{darkgreen}{\textbf{0.9339}} & \textcolor{darkgreen}{\textbf{0.2317}} & \textcolor{darkred}{\textbf{0.7467}} & \textcolor{darkred}{\textbf{0.6607}} & \textcolor{darkred}{\textbf{0.9241}} & 0.6687 & 3/9 \\ \hline
NONE & NONE & \textcolor{darkred}{\textbf{0.9324}} & \textcolor{darkred}{\textbf{0.6407}} & \textcolor{darkred}{\textbf{0.7628}} & \textcolor{darkred}{\textbf{0.1881}} & \textcolor{darkred}{\textbf{0.9339}} & \textcolor{darkred}{\textbf{0.2317}} & \textcolor{darkred}{\textbf{0.7360}} & \textcolor{darkred}{\textbf{0.6421}} & \textcolor{darkred}{\textbf{0.9241}} & 0.6658 & 0/9 \\ \hline
GaussianShadow & L & \textcolor{darkred}{\textbf{0.9310}} & \textcolor{darkred}{\textbf{0.6298}} & \textcolor{darkred}{\textbf{0.7779}} & \textcolor{darkred}{\textbf{0.1947}} & \textcolor{darkred}{\textbf{0.8973}} & \textcolor{darkred}{\textbf{0.2729}} & \textcolor{darkred}{\textbf{0.6495}} & \textcolor{darkred}{\textbf{0.5663}} & \textcolor{darkred}{\textbf{0.9056}} & 0.6472 & 0/9 \\ \hline
RandomBrightnessContrast & H & \textcolor{darkred}{\textbf{0.9023}} & \textcolor{darkgreen}{\textbf{0.2956}} & \textcolor{darkgreen}{\textbf{0.5566}} & \textcolor{darkred}{\textbf{0.2340}} & \textcolor{darkgreen}{\textbf{0.8704}} & \textcolor{darkgreen}{\textbf{0.5417}} & \textcolor{darkgreen}{\textbf{0.6408}} & \textcolor{darkred}{\textbf{0.6868}} & \textcolor{darkgreen}{\textbf{0.9169}} & 0.6272 & 6/9 \\ \hline

\end{tabular}
\end{table}

\newgeometry{left=0.05cm,right=0.1cm,top=1.5cm,bottom=1.5cm}
\footnotesize
\setlength{\tabcolsep}{1.3pt}
\renewcommand{\arraystretch}{1.2}

\begin{longtable}{|>{\fontsize{8.5}{11}\selectfont}l|ccc|ccc|ccc|c|c|}
\caption{Dice Scores and Statistical Significance of Data Pairwise Augmentation Methods}
\label{tab:game_personality} \\

\hline
\textbf{Augmentation} &
\multicolumn{3}{c|}{\textbf{Unity}} &
\multicolumn{3}{c|}{\textbf{CAMUS}} &
\multicolumn{3}{c|}{\textbf{EchoNet}} &
\textbf{Row Avg} & \textbf{Sig 0.05} \\
\cline{2-10}
& CS & CA & ED & CS & CA & ED & CS & CA & ED & & \\
\hline
\endfirsthead

\hline
\textbf{Augmentation} &
\multicolumn{3}{c|}{\textbf{Unity}} &
\multicolumn{3}{c|}{\textbf{CAMUS}} &
\multicolumn{3}{c|}{\textbf{EchoNet}} &
\textbf{Row Avg} & \textbf{Sig 0.05} \\
\cline{2-10}
& CS & CA & ED & CS & CA & ED & CS & CA & ED & & \\
\hline
\endhead

\hline
\multicolumn{12}{r}{\textit{Continued on next page}} \\
\hline
\endfoot

\hline
\endlastfoot
RandomHorizontalFlip(L) Affine(L) & \textcolor{darkred}{\textbf{0.9344}} & \textcolor{darkgreen}{\textbf{0.8178}} & \textcolor{darkgreen}{\textbf{0.8163}} & \textcolor{darkgreen}{\textbf{0.7291}} & \textcolor{darkred}{\textbf{0.9367}} & \textcolor{darkgreen}{\textbf{0.6229}} & \textcolor{darkred}{\textbf{0.7836}} & \textcolor{darkgreen}{\textbf{0.8883}} & \textcolor{darkgreen}{\textbf{0.9255}} & 0.8283 & 6/9 \\ \hline
ColorJitter(H) RandomHorizontalFlip(L) & \textcolor{darkred}{\textbf{0.9339}} & \textcolor{darkgreen}{\textbf{0.7529}} & \textcolor{darkgreen}{\textbf{0.7700}} & \textcolor{darkgreen}{\textbf{0.6202}} & \textcolor{darkred}{\textbf{0.9339}} & \textcolor{darkgreen}{\textbf{0.8077}} & \textcolor{darkgreen}{\textbf{0.6559}} & \textcolor{darkgreen}{\textbf{0.8942}} & \textcolor{darkred}{\textbf{0.9250}} & 0.8104 & 6/9 \\ \hline
RandomHorizontalFlip(L) Affine(C1) & \textcolor{darkred}{\textbf{0.9354}} & \textcolor{darkgreen}{\textbf{0.8124}} & \textcolor{darkgreen}{\textbf{0.8031}} & \textcolor{darkgreen}{\textbf{0.6749}} & \textcolor{darkgreen}{\textbf{0.9360}} & \textcolor{darkgreen}{\textbf{0.5891}} & \textcolor{darkred}{\textbf{0.7328}} & \textcolor{darkgreen}{\textbf{0.8696}} & \textcolor{darkgreen}{\textbf{0.9259}} & 0.8088 & 7/9 \\ \hline
RandomHorizontalFlip(L) ShiftScaleRotate(C1) & \textcolor{darkgreen}{\textbf{0.9351}} & \textcolor{darkgreen}{\textbf{0.8104}} & \textcolor{darkgreen}{\textbf{0.8030}} & \textcolor{darkgreen}{\textbf{0.5762}} & \textcolor{darkred}{\textbf{0.9356}} & \textcolor{darkgreen}{\textbf{0.5591}} & \textcolor{darkgreen}{\textbf{0.8077}} & \textcolor{darkgreen}{\textbf{0.8805}} & \textcolor{darkgreen}{\textbf{0.9264}} & 0.8038 & 8/9 \\ \hline
RandomHorizontalFlip(L) Affine(H) & \textcolor{darkgreen}{\textbf{0.9349}} & \textcolor{darkgreen}{\textbf{0.8132}} & \textcolor{darkred}{\textbf{0.7952}} & \textcolor{darkgreen}{\textbf{0.5951}} & \textcolor{darkred}{\textbf{0.9354}} & \textcolor{darkgreen}{\textbf{0.5668}} & \textcolor{darkred}{\textbf{0.7936}} & \textcolor{darkgreen}{\textbf{0.8688}} & \textcolor{darkred}{\textbf{0.9246}} & 0.8031 & 5/9 \\ \hline
RandomHorizontalFlip(L) Perspective(C1) & \textcolor{darkgreen}{\textbf{0.9334}} & \textcolor{darkgreen}{\textbf{0.8309}} & \textcolor{darkgreen}{\textbf{0.8416}} & \textcolor{darkgreen}{\textbf{0.6155}} & \textcolor{darkred}{\textbf{0.9360}} & \textcolor{darkgreen}{\textbf{0.5638}} & \textcolor{darkred}{\textbf{0.6995}} & \textcolor{darkgreen}{\textbf{0.8745}} & \textcolor{darkred}{\textbf{0.9255}} & 0.8023 & 6/9 \\ \hline
RandomHorizontalFlip(L) GridDistortion(C3) & \textcolor{darkgreen}{\textbf{0.9349}} & \textcolor{darkgreen}{\textbf{0.7865}} & \textcolor{darkgreen}{\textbf{0.8004}} & \textcolor{darkgreen}{\textbf{0.6125}} & \textcolor{darkgreen}{\textbf{0.9371}} & \textcolor{darkgreen}{\textbf{0.5424}} & \textcolor{darkred}{\textbf{0.7664}} & \textcolor{darkgreen}{\textbf{0.8920}} & \textcolor{darkred}{\textbf{0.9260}} & 0.7998 & 7/9 \\ \hline
RandomHorizontalFlip(L) ShiftScaleRotate(H) & \textcolor{darkgreen}{\textbf{0.9350}} & \textcolor{darkgreen}{\textbf{0.7974}} & \textcolor{darkgreen}{\textbf{0.7996}} & \textcolor{darkgreen}{\textbf{0.5814}} & \textcolor{darkgreen}{\textbf{0.9372}} & \textcolor{darkgreen}{\textbf{0.6001}} & \textcolor{darkred}{\textbf{0.7386}} & \textcolor{darkgreen}{\textbf{0.8742}} & \textcolor{darkgreen}{\textbf{0.9262}} & 0.7989 & 8/9 \\ \hline
ColorJitter(H) Affine(H) & \textcolor{darkgreen}{\textbf{0.9356}} & \textcolor{darkgreen}{\textbf{0.7554}} & \textcolor{darkgreen}{\textbf{0.7854}} & \textcolor{darkgreen}{\textbf{0.6030}} & \textcolor{darkred}{\textbf{0.9337}} & \textcolor{darkgreen}{\textbf{0.6191}} & \textcolor{darkgreen}{\textbf{0.8246}} & \textcolor{darkred}{\textbf{0.7298}} & \textcolor{darkgreen}{\textbf{0.9251}} & 0.7902 & 7/9 \\ \hline
RandomHorizontalFlip(L) Perspective(H) & \textcolor{darkred}{\textbf{0.9332}} & \textcolor{darkgreen}{\textbf{0.8288}} & \textcolor{darkgreen}{\textbf{0.8491}} & \textcolor{darkgreen}{\textbf{0.5437}} & \textcolor{darkred}{\textbf{0.9347}} & \textcolor{darkgreen}{\textbf{0.5487}} & \textcolor{darkred}{\textbf{0.5803}} & \textcolor{darkgreen}{\textbf{0.8777}} & \textcolor{darkred}{\textbf{0.9254}} & 0.7802 & 5/9 \\ \hline
ColorJitter(H) ShiftScaleRotate(H) & \textcolor{darkgreen}{\textbf{0.9361}} & \textcolor{darkgreen}{\textbf{0.7255}} & \textcolor{darkgreen}{\textbf{0.8026}} & \textcolor{darkgreen}{\textbf{0.5858}} & \textcolor{darkgreen}{\textbf{0.9357}} & \textcolor{darkgreen}{\textbf{0.6308}} & \textcolor{darkgreen}{\textbf{0.8081}} & \textcolor{darkred}{\textbf{0.6595}} & \textcolor{darkgreen}{\textbf{0.9262}} & 0.7789 & 8/9 \\ \hline
Affine(L) ShiftScaleRotate(H) & \textcolor{darkred}{\textbf{0.9336}} & \textcolor{darkgreen}{\textbf{0.7289}} & \textcolor{darkgreen}{\textbf{0.8125}} & \textcolor{darkgreen}{\textbf{0.5869}} & \textcolor{darkgreen}{\textbf{0.9385}} & \textcolor{darkgreen}{\textbf{0.4972}} & \textcolor{darkgreen}{\textbf{0.8318}} & \textcolor{darkgreen}{\textbf{0.7480}} & \textcolor{darkgreen}{\textbf{0.9254}} & 0.7781 & 8/9 \\ \hline
RandomHorizontalFlip(L) GridDistortion(C2) & \textcolor{darkred}{\textbf{0.9340}} & \textcolor{darkgreen}{\textbf{0.7753}} & \textcolor{darkgreen}{\textbf{0.7984}} & \textcolor{darkred}{\textbf{0.4460}} & \textcolor{darkgreen}{\textbf{0.9371}} & \textcolor{darkgreen}{\textbf{0.5019}} & \textcolor{darkgreen}{\textbf{0.7737}} & \textcolor{darkgreen}{\textbf{0.8844}} & \textcolor{darkgreen}{\textbf{0.9255}} & 0.7751 & 7/9 \\ \hline
ColorJitter(H) Affine(L) & \textcolor{darkgreen}{\textbf{0.9364}} & \textcolor{darkgreen}{\textbf{0.7234}} & \textcolor{darkgreen}{\textbf{0.8031}} & \textcolor{darkgreen}{\textbf{0.5144}} & \textcolor{darkred}{\textbf{0.9341}} & \textcolor{darkgreen}{\textbf{0.5952}} & \textcolor{darkgreen}{\textbf{0.8270}} & \textcolor{darkred}{\textbf{0.7117}} & \textcolor{darkgreen}{\textbf{0.9253}} & 0.7745 & 7/9 \\ \hline
RandomHorizontalFlip(L) ShiftScaleRotate(L) & \textcolor{darkgreen}{\textbf{0.9352}} & \textcolor{darkgreen}{\textbf{0.7678}} & \textcolor{darkgreen}{\textbf{0.7859}} & \textcolor{darkgreen}{\textbf{0.4969}} & \textcolor{darkgreen}{\textbf{0.9375}} & \textcolor{darkgreen}{\textbf{0.5869}} & \textcolor{darkred}{\textbf{0.6406}} & \textcolor{darkgreen}{\textbf{0.8863}} & \textcolor{darkgreen}{\textbf{0.9261}} & 0.7737 & 8/9 \\ \hline
RandomHorizontalFlip(L) GridDistortion(H) & \textcolor{darkgreen}{\textbf{0.9352}} & \textcolor{darkgreen}{\textbf{0.7650}} & \textcolor{darkgreen}{\textbf{0.8000}} & \textcolor{darkgreen}{\textbf{0.4214}} & \textcolor{darkgreen}{\textbf{0.9373}} & \textcolor{darkgreen}{\textbf{0.5213}} & \textcolor{darkred}{\textbf{0.7573}} & \textcolor{darkgreen}{\textbf{0.8824}} & \textcolor{darkgreen}{\textbf{0.9257}} & 0.7717 & 8/9 \\ \hline
ColorJitter(H) ShiftScaleRotate(C1) & \textcolor{darkgreen}{\textbf{0.9359}} & \textcolor{darkgreen}{\textbf{0.7476}} & \textcolor{darkgreen}{\textbf{0.8040}} & \textcolor{darkgreen}{\textbf{0.5594}} & \textcolor{darkred}{\textbf{0.9343}} & \textcolor{darkgreen}{\textbf{0.5672}} & \textcolor{darkgreen}{\textbf{0.8220}} & \textcolor{darkred}{\textbf{0.6293}} & \textcolor{darkgreen}{\textbf{0.9259}} & 0.7695 & 7/9 \\ \hline
ColorJitter(H) Affine(C1) & \textcolor{darkgreen}{\textbf{0.9360}} & \textcolor{darkred}{\textbf{0.7026}} & \textcolor{darkgreen}{\textbf{0.7968}} & \textcolor{darkgreen}{\textbf{0.4943}} & \textcolor{darkred}{\textbf{0.9339}} & \textcolor{darkgreen}{\textbf{0.6072}} & \textcolor{darkgreen}{\textbf{0.8002}} & \textcolor{darkgreen}{\textbf{0.7259}} & \textcolor{darkred}{\textbf{0.9253}} & 0.7691 & 6/9 \\ \hline
Affine(L) ShiftScaleRotate(C1) & \textcolor{darkgreen}{\textbf{0.9343}} & \textcolor{darkgreen}{\textbf{0.7565}} & \textcolor{darkgreen}{\textbf{0.8190}} & \textcolor{darkgreen}{\textbf{0.5609}} & \textcolor{darkgreen}{\textbf{0.9375}} & \textcolor{darkgreen}{\textbf{0.4358}} & \textcolor{darkgreen}{\textbf{0.8345}} & \textcolor{darkgreen}{\textbf{0.7180}} & \textcolor{darkred}{\textbf{0.9253}} & 0.7691 & 8/9 \\ \hline
Affine(L) ShiftScaleRotate(L) & \textcolor{darkred}{\textbf{0.9355}} & \textcolor{darkgreen}{\textbf{0.7108}} & \textcolor{darkgreen}{\textbf{0.8159}} & \textcolor{darkgreen}{\textbf{0.5563}} & \textcolor{darkgreen}{\textbf{0.9390}} & \textcolor{darkgreen}{\textbf{0.4898}} & \textcolor{darkred}{\textbf{0.7417}} & \textcolor{darkgreen}{\textbf{0.7815}} & \textcolor{darkgreen}{\textbf{0.9257}} & 0.7662 & 7/9 \\ \hline
Affine(H) ShiftScaleRotate(H) & \textcolor{darkgreen}{\textbf{0.9347}} & \textcolor{darkgreen}{\textbf{0.7263}} & \textcolor{darkgreen}{\textbf{0.7935}} & \textcolor{darkgreen}{\textbf{0.6127}} & \textcolor{darkred}{\textbf{0.9365}} & \textcolor{darkgreen}{\textbf{0.4640}} & \textcolor{darkgreen}{\textbf{0.8218}} & \textcolor{darkred}{\textbf{0.6797}} & \textcolor{darkgreen}{\textbf{0.9251}} & 0.7660 & 7/9 \\ \hline
Affine(L) GridDistortion(C2) & \textcolor{darkgreen}{\textbf{0.9363}} & \textcolor{darkgreen}{\textbf{0.7304}} & \textcolor{darkgreen}{\textbf{0.8220}} & \textcolor{darkgreen}{\textbf{0.5414}} & \textcolor{darkgreen}{\textbf{0.9389}} & \textcolor{darkgreen}{\textbf{0.4424}} & \textcolor{darkgreen}{\textbf{0.8323}} & \textcolor{darkgreen}{\textbf{0.7157}} & \textcolor{darkgreen}{\textbf{0.9260}} & 0.7650 & 9/9 \\ \hline
Affine(L) GridDistortion(H) & \textcolor{darkred}{\textbf{0.9348}} & \textcolor{darkgreen}{\textbf{0.7304}} & \textcolor{darkgreen}{\textbf{0.8233}} & \textcolor{darkgreen}{\textbf{0.5708}} & \textcolor{darkgreen}{\textbf{0.9391}} & \textcolor{darkgreen}{\textbf{0.4403}} & \textcolor{darkgreen}{\textbf{0.8252}} & \textcolor{darkred}{\textbf{0.6877}} & \textcolor{darkgreen}{\textbf{0.9258}} & 0.7642 & 7/9 \\ \hline
ColorJitter(H) Perspective(C1) & \textcolor{darkgreen}{\textbf{0.9357}} & \textcolor{darkgreen}{\textbf{0.8201}} & \textcolor{darkgreen}{\textbf{0.8307}} & \textcolor{darkgreen}{\textbf{0.4124}} & \textcolor{darkgreen}{\textbf{0.9350}} & \textcolor{darkgreen}{\textbf{0.5476}} & \textcolor{darkred}{\textbf{0.7341}} & \textcolor{darkred}{\textbf{0.7326}} & \textcolor{darkgreen}{\textbf{0.9255}} & 0.7637 & 7/9 \\ \hline
Affine(L) GridDistortion(C3) & \textcolor{darkgreen}{\textbf{0.9359}} & \textcolor{darkgreen}{\textbf{0.7254}} & \textcolor{darkgreen}{\textbf{0.8235}} & \textcolor{darkgreen}{\textbf{0.5806}} & \textcolor{darkgreen}{\textbf{0.9399}} & \textcolor{darkgreen}{\textbf{0.4423}} & \textcolor{darkgreen}{\textbf{0.8244}} & \textcolor{darkred}{\textbf{0.6727}} & \textcolor{darkgreen}{\textbf{0.9260}} & 0.7634 & 8/9 \\ \hline
ColorJitter(H) ShiftScaleRotate(L) & \textcolor{darkred}{\textbf{0.9347}} & \textcolor{darkred}{\textbf{0.6276}} & \textcolor{darkgreen}{\textbf{0.7971}} & \textcolor{darkgreen}{\textbf{0.5123}} & \textcolor{darkred}{\textbf{0.9357}} & \textcolor{darkgreen}{\textbf{0.6158}} & \textcolor{darkgreen}{\textbf{0.7789}} & \textcolor{darkred}{\textbf{0.7413}} & \textcolor{darkgreen}{\textbf{0.9261}} & 0.7633 & 5/9 \\ \hline
ColorJitter(H) GridDistortion(H) & \textcolor{darkgreen}{\textbf{0.9341}} & \textcolor{darkred}{\textbf{0.6394}} & \textcolor{darkgreen}{\textbf{0.8036}} & \textcolor{darkgreen}{\textbf{0.5119}} & \textcolor{darkred}{\textbf{0.9358}} & \textcolor{darkgreen}{\textbf{0.5995}} & \textcolor{darkred}{\textbf{0.7751}} & \textcolor{darkgreen}{\textbf{0.7299}} & \textcolor{darkgreen}{\textbf{0.9255}} & 0.7616 & 6/9 \\ \hline
GridDistortion(C3) ShiftScaleRotate(H) & \textcolor{darkgreen}{\textbf{0.9365}} & \textcolor{darkgreen}{\textbf{0.7018}} & \textcolor{darkgreen}{\textbf{0.8172}} & \textcolor{darkgreen}{\textbf{0.5045}} & \textcolor{darkgreen}{\textbf{0.9382}} & \textcolor{darkgreen}{\textbf{0.4471}} & \textcolor{darkgreen}{\textbf{0.8058}} & \textcolor{darkgreen}{\textbf{0.7752}} & \textcolor{darkgreen}{\textbf{0.9265}} & 0.7614 & 9/9 \\ \hline
ColorJitter(H) Perspective(H) & \textcolor{darkgreen}{\textbf{0.9339}} & \textcolor{darkgreen}{\textbf{0.8248}} & \textcolor{darkgreen}{\textbf{0.8329}} & \textcolor{darkred}{\textbf{0.3922}} & \textcolor{darkred}{\textbf{0.9323}} & \textcolor{darkgreen}{\textbf{0.5318}} & \textcolor{darkred}{\textbf{0.7098}} & \textcolor{darkgreen}{\textbf{0.7607}} & \textcolor{darkred}{\textbf{0.9245}} & 0.7603 & 5/9 \\ \hline
ColorJitter(H) GridDistortion(C2) & \textcolor{darkgreen}{\textbf{0.9338}} & \textcolor{darkred}{\textbf{0.6624}} & \textcolor{darkgreen}{\textbf{0.8045}} & \textcolor{darkgreen}{\textbf{0.4782}} & \textcolor{darkred}{\textbf{0.9362}} & \textcolor{darkgreen}{\textbf{0.5542}} & \textcolor{darkgreen}{\textbf{0.7790}} & \textcolor{darkgreen}{\textbf{0.7621}} & \textcolor{darkred}{\textbf{0.9259}} & 0.7596 & 6/9 \\ \hline
GridDistortion(C2) ShiftScaleRotate(C1) & \textcolor{darkgreen}{\textbf{0.9352}} & \textcolor{darkgreen}{\textbf{0.7390}} & \textcolor{darkgreen}{\textbf{0.8167}} & \textcolor{darkgreen}{\textbf{0.4706}} & \textcolor{darkgreen}{\textbf{0.9382}} & \textcolor{darkgreen}{\textbf{0.4527}} & \textcolor{darkgreen}{\textbf{0.8198}} & \textcolor{darkgreen}{\textbf{0.7326}} & \textcolor{darkred}{\textbf{0.9251}} & 0.7589 & 8/9 \\ \hline
ColorJitter(H) GridDistortion(C3) & \textcolor{darkred}{\textbf{0.9353}} & \textcolor{darkred}{\textbf{0.6715}} & \textcolor{darkgreen}{\textbf{0.8023}} & \textcolor{darkgreen}{\textbf{0.5045}} & \textcolor{darkred}{\textbf{0.9357}} & \textcolor{darkgreen}{\textbf{0.5499}} & \textcolor{darkred}{\textbf{0.7511}} & \textcolor{darkgreen}{\textbf{0.7507}} & \textcolor{darkgreen}{\textbf{0.9261}} & 0.7586 & 5/9 \\ \hline
Affine(H) GridDistortion(C3) & \textcolor{darkgreen}{\textbf{0.9354}} & \textcolor{darkgreen}{\textbf{0.7853}} & \textcolor{darkgreen}{\textbf{0.8218}} & \textcolor{darkred}{\textbf{0.4367}} & \textcolor{darkgreen}{\textbf{0.9367}} & \textcolor{darkgreen}{\textbf{0.4323}} & \textcolor{darkgreen}{\textbf{0.8321}} & \textcolor{darkred}{\textbf{0.7166}} & \textcolor{darkgreen}{\textbf{0.9256}} & 0.7581 & 7/9 \\ \hline
Affine(H) GridDistortion(C2) & \textcolor{darkgreen}{\textbf{0.9355}} & \textcolor{darkgreen}{\textbf{0.7639}} & \textcolor{darkgreen}{\textbf{0.8134}} & \textcolor{darkred}{\textbf{0.3501}} & \textcolor{darkgreen}{\textbf{0.9374}} & \textcolor{darkgreen}{\textbf{0.4667}} & \textcolor{darkgreen}{\textbf{0.8357}} & \textcolor{darkgreen}{\textbf{0.7573}} & \textcolor{darkgreen}{\textbf{0.9260}} & 0.7540 & 8/9 \\ \hline
Affine(H) ShiftScaleRotate(C1) & \textcolor{darkgreen}{\textbf{0.9350}} & \textcolor{darkgreen}{\textbf{0.7427}} & \textcolor{darkgreen}{\textbf{0.8023}} & \textcolor{darkgreen}{\textbf{0.5194}} & \textcolor{darkgreen}{\textbf{0.9334}} & \textcolor{darkgreen}{\textbf{0.4203}} & \textcolor{darkred}{\textbf{0.8333}} & \textcolor{darkred}{\textbf{0.6713}} & \textcolor{darkgreen}{\textbf{0.9254}} & 0.7537 & 7/9 \\ \hline
GridDistortion(C2) ShiftScaleRotate(H) & \textcolor{darkgreen}{\textbf{0.9363}} & \textcolor{darkgreen}{\textbf{0.7423}} & \textcolor{darkgreen}{\textbf{0.8160}} & \textcolor{darkgreen}{\textbf{0.4587}} & \textcolor{darkgreen}{\textbf{0.9380}} & \textcolor{darkgreen}{\textbf{0.4413}} & \textcolor{darkgreen}{\textbf{0.7929}} & \textcolor{darkred}{\textbf{0.6983}} & \textcolor{darkred}{\textbf{0.9261}} & 0.7500 & 7/9 \\ \hline
ShiftScaleRotate(H) Affine(C1) & \textcolor{darkred}{\textbf{0.9342}} & \textcolor{darkgreen}{\textbf{0.7609}} & \textcolor{darkgreen}{\textbf{0.8159}} & \textcolor{darkgreen}{\textbf{0.4288}} & \textcolor{darkgreen}{\textbf{0.9377}} & \textcolor{darkgreen}{\textbf{0.4566}} & \textcolor{darkred}{\textbf{0.7861}} & \textcolor{darkred}{\textbf{0.6981}} & \textcolor{darkgreen}{\textbf{0.9256}} & 0.7493 & 6/9 \\ \hline
Affine(H) ShiftScaleRotate(L) & \textcolor{darkgreen}{\textbf{0.9355}} & \textcolor{darkgreen}{\textbf{0.7208}} & \textcolor{darkred}{\textbf{0.7816}} & \textcolor{darkred}{\textbf{0.4795}} & \textcolor{darkgreen}{\textbf{0.9383}} & \textcolor{darkgreen}{\textbf{0.4666}} & \textcolor{darkgreen}{\textbf{0.8269}} & \textcolor{darkred}{\textbf{0.6561}} & \textcolor{darkgreen}{\textbf{0.9259}} & 0.7479 & 6/9 \\ \hline
GridDistortion(C3) ShiftScaleRotate(C1) & \textcolor{darkgreen}{\textbf{0.9362}} & \textcolor{darkgreen}{\textbf{0.7575}} & \textcolor{darkgreen}{\textbf{0.8163}} & \textcolor{darkgreen}{\textbf{0.4363}} & \textcolor{darkgreen}{\textbf{0.9383}} & \textcolor{darkgreen}{\textbf{0.4188}} & \textcolor{darkgreen}{\textbf{0.8232}} & \textcolor{darkred}{\textbf{0.6752}} & \textcolor{darkgreen}{\textbf{0.9259}} & 0.7475 & 8/9 \\ \hline
Affine(C1) ShiftScaleRotate(C1) & \textcolor{darkgreen}{\textbf{0.9355}} & \textcolor{darkgreen}{\textbf{0.7381}} & \textcolor{darkgreen}{\textbf{0.8185}} & \textcolor{darkgreen}{\textbf{0.4934}} & \textcolor{darkgreen}{\textbf{0.9370}} & \textcolor{darkgreen}{\textbf{0.4062}} & \textcolor{darkgreen}{\textbf{0.8362}} & \textcolor{darkred}{\textbf{0.6366}} & \textcolor{darkgreen}{\textbf{0.9254}} & 0.7474 & 8/9 \\ \hline
Affine(C1) ShiftScaleRotate(L) & \textcolor{darkgreen}{\textbf{0.9354}} & \textcolor{darkgreen}{\textbf{0.7339}} & \textcolor{darkgreen}{\textbf{0.8025}} & \textcolor{darkgreen}{\textbf{0.4652}} & \textcolor{darkgreen}{\textbf{0.9386}} & \textcolor{darkgreen}{\textbf{0.4390}} & \textcolor{darkred}{\textbf{0.7519}} & \textcolor{darkred}{\textbf{0.7255}} & \textcolor{darkgreen}{\textbf{0.9261}} & 0.7465 & 7/9 \\ \hline
GridDistortion(C2) Affine(C1) & \textcolor{darkgreen}{\textbf{0.9359}} & \textcolor{darkgreen}{\textbf{0.7324}} & \textcolor{darkgreen}{\textbf{0.8190}} & \textcolor{darkgreen}{\textbf{0.3953}} & \textcolor{darkgreen}{\textbf{0.9387}} & \textcolor{darkgreen}{\textbf{0.4232}} & \textcolor{darkgreen}{\textbf{0.8155}} & \textcolor{darkred}{\textbf{0.7140}} & \textcolor{darkgreen}{\textbf{0.9262}} & 0.7445 & 8/9 \\ \hline
GridDistortion(H) ShiftScaleRotate(H) & \textcolor{darkgreen}{\textbf{0.9361}} & \textcolor{darkgreen}{\textbf{0.7036}} & \textcolor{darkgreen}{\textbf{0.8163}} & \textcolor{darkred}{\textbf{0.3901}} & \textcolor{darkgreen}{\textbf{0.9388}} & \textcolor{darkgreen}{\textbf{0.4328}} & \textcolor{darkred}{\textbf{0.7763}} & \textcolor{darkgreen}{\textbf{0.7567}} & \textcolor{darkgreen}{\textbf{0.9255}} & 0.7418 & 7/9 \\ \hline
Affine(H) GridDistortion(H) & \textcolor{darkgreen}{\textbf{0.9355}} & \textcolor{darkgreen}{\textbf{0.7680}} & \textcolor{darkgreen}{\textbf{0.8092}} & \textcolor{darkgreen}{\textbf{0.3391}} & \textcolor{darkgreen}{\textbf{0.9373}} & \textcolor{darkgreen}{\textbf{0.4375}} & \textcolor{darkgreen}{\textbf{0.8407}} & \textcolor{darkred}{\textbf{0.6790}} & \textcolor{darkgreen}{\textbf{0.9257}} & 0.7413 & 8/9 \\ \hline
Affine(H) Perspective(H) & \textcolor{darkgreen}{\textbf{0.9344}} & \textcolor{darkgreen}{\textbf{0.7762}} & \textcolor{darkgreen}{\textbf{0.8132}} & \textcolor{darkred}{\textbf{0.3951}} & \textcolor{darkred}{\textbf{0.9353}} & \textcolor{darkgreen}{\textbf{0.3941}} & \textcolor{darkred}{\textbf{0.7840}} & \textcolor{darkred}{\textbf{0.6948}} & \textcolor{darkred}{\textbf{0.9249}} & 0.7391 & 4/9 \\ \hline
Affine(H) Perspective(C1) & \textcolor{darkgreen}{\textbf{0.9361}} & \textcolor{darkgreen}{\textbf{0.7830}} & \textcolor{darkgreen}{\textbf{0.8072}} & \textcolor{darkgreen}{\textbf{0.3954}} & \textcolor{darkred}{\textbf{0.9334}} & \textcolor{darkgreen}{\textbf{0.3966}} & \textcolor{darkred}{\textbf{0.7895}} & \textcolor{darkred}{\textbf{0.6811}} & \textcolor{darkred}{\textbf{0.9249}} & 0.7386 & 5/9 \\ \hline
Affine(L) Perspective(C1) & \textcolor{darkgreen}{\textbf{0.9352}} & \textcolor{darkgreen}{\textbf{0.7978}} & \textcolor{darkgreen}{\textbf{0.8327}} & \textcolor{darkgreen}{\textbf{0.3936}} & \textcolor{darkgreen}{\textbf{0.9386}} & \textcolor{darkgreen}{\textbf{0.3690}} & \textcolor{darkred}{\textbf{0.7613}} & \textcolor{darkred}{\textbf{0.6808}} & \textcolor{darkred}{\textbf{0.9252}} & 0.7371 & 6/9 \\ \hline
GridDistortion(H) ShiftScaleRotate(C1) & \textcolor{darkgreen}{\textbf{0.9367}} & \textcolor{darkgreen}{\textbf{0.7520}} & \textcolor{darkgreen}{\textbf{0.8173}} & \textcolor{darkgreen}{\textbf{0.3670}} & \textcolor{darkgreen}{\textbf{0.9378}} & \textcolor{darkgreen}{\textbf{0.4016}} & \textcolor{darkred}{\textbf{0.7942}} & \textcolor{darkred}{\textbf{0.6584}} & \textcolor{darkgreen}{\textbf{0.9258}} & 0.7323 & 7/9 \\ \hline
GridDistortion(C2) Perspective(C1) & \textcolor{darkgreen}{\textbf{0.9359}} & \textcolor{darkgreen}{\textbf{0.8208}} & \textcolor{darkgreen}{\textbf{0.8428}} & \textcolor{darkred}{\textbf{0.3193}} & \textcolor{darkgreen}{\textbf{0.9382}} & \textcolor{darkgreen}{\textbf{0.3791}} & \textcolor{darkred}{\textbf{0.7538}} & \textcolor{darkred}{\textbf{0.6564}} & \textcolor{darkgreen}{\textbf{0.9265}} & 0.7303 & 6/9 \\ \hline
GridDistortion(C3) Perspective(C1) & \textcolor{darkgreen}{\textbf{0.9351}} & \textcolor{darkgreen}{\textbf{0.8031}} & \textcolor{darkgreen}{\textbf{0.8378}} & \textcolor{darkred}{\textbf{0.3269}} & \textcolor{darkgreen}{\textbf{0.9374}} & \textcolor{darkgreen}{\textbf{0.3707}} & \textcolor{darkred}{\textbf{0.7508}} & \textcolor{darkred}{\textbf{0.6754}} & \textcolor{darkgreen}{\textbf{0.9266}} & 0.7293 & 6/9 \\ \hline
Affine(L) Perspective(H) & \textcolor{darkred}{\textbf{0.9348}} & \textcolor{darkgreen}{\textbf{0.7989}} & \textcolor{darkgreen}{\textbf{0.8404}} & \textcolor{darkgreen}{\textbf{0.3198}} & \textcolor{darkgreen}{\textbf{0.9381}} & \textcolor{darkgreen}{\textbf{0.3833}} & \textcolor{darkred}{\textbf{0.7199}} & \textcolor{darkred}{\textbf{0.6788}} & \textcolor{darkred}{\textbf{0.9241}} & 0.7265 & 5/9 \\ \hline
Perspective(C1) Affine(C1) & \textcolor{darkgreen}{\textbf{0.9338}} & \textcolor{darkgreen}{\textbf{0.7658}} & \textcolor{darkgreen}{\textbf{0.8335}} & \textcolor{darkgreen}{\textbf{0.4234}} & \textcolor{darkred}{\textbf{0.9373}} & \textcolor{darkgreen}{\textbf{0.3725}} & \textcolor{darkred}{\textbf{0.7260}} & \textcolor{darkred}{\textbf{0.6082}} & \textcolor{darkred}{\textbf{0.9263}} & 0.7252 & 5/9 \\ \hline
GridDistortion(H) Affine(C1) & \textcolor{darkred}{\textbf{0.9371}} & \textcolor{darkgreen}{\textbf{0.7320}} & \textcolor{darkgreen}{\textbf{0.8228}} & \textcolor{darkred}{\textbf{0.3737}} & \textcolor{darkred}{\textbf{0.9371}} & \textcolor{darkgreen}{\textbf{0.3827}} & \textcolor{darkred}{\textbf{0.7697}} & \textcolor{darkred}{\textbf{0.6449}} & \textcolor{darkgreen}{\textbf{0.9254}} & 0.7251 & 4/9 \\ \hline
Perspective(C1) ShiftScaleRotate(C1) & \textcolor{darkgreen}{\textbf{0.9348}} & \textcolor{darkgreen}{\textbf{0.7979}} & \textcolor{darkgreen}{\textbf{0.8403}} & \textcolor{darkgreen}{\textbf{0.4950}} & \textcolor{darkgreen}{\textbf{0.9373}} & \textcolor{darkgreen}{\textbf{0.3958}} & \textcolor{darkred}{\textbf{0.6785}} & \textcolor{darkred}{\textbf{0.5158}} & \textcolor{darkred}{\textbf{0.9257}} & 0.7246 & 6/9 \\ \hline
Affine(C1) GridDistortion(C3) & \textcolor{darkgreen}{\textbf{0.9362}} & \textcolor{darkgreen}{\textbf{0.7414}} & \textcolor{darkgreen}{\textbf{0.8239}} & \textcolor{darkred}{\textbf{0.3642}} & \textcolor{darkred}{\textbf{0.9378}} & \textcolor{darkred}{\textbf{0.3592}} & \textcolor{darkgreen}{\textbf{0.7962}} & \textcolor{darkred}{\textbf{0.6333}} & \textcolor{darkgreen}{\textbf{0.9258}} & 0.7242 & 5/9 \\ \hline
GridDistortion(C2) Perspective(H) & \textcolor{darkred}{\textbf{0.9341}} & \textcolor{darkgreen}{\textbf{0.8251}} & \textcolor{darkgreen}{\textbf{0.8460}} & \textcolor{darkgreen}{\textbf{0.3069}} & \textcolor{darkgreen}{\textbf{0.9381}} & \textcolor{darkgreen}{\textbf{0.3769}} & \textcolor{darkred}{\textbf{0.6829}} & \textcolor{darkred}{\textbf{0.6691}} & \textcolor{darkgreen}{\textbf{0.9257}} & 0.7228 & 6/9 \\ \hline
GridDistortion(C3) ShiftScaleRotate(L) & \textcolor{darkred}{\textbf{0.9345}} & \textcolor{darkred}{\textbf{0.6757}} & \textcolor{darkgreen}{\textbf{0.8057}} & \textcolor{darkred}{\textbf{0.2541}} & \textcolor{darkgreen}{\textbf{0.9376}} & \textcolor{darkgreen}{\textbf{0.4087}} & \textcolor{darkgreen}{\textbf{0.7802}} & \textcolor{darkgreen}{\textbf{0.7718}} & \textcolor{darkgreen}{\textbf{0.9267}} & 0.7217 & 6/9 \\ \hline
GridDistortion(H) Perspective(H) & \textcolor{darkred}{\textbf{0.9348}} & \textcolor{darkgreen}{\textbf{0.8261}} & \textcolor{darkgreen}{\textbf{0.8464}} & \textcolor{darkred}{\textbf{0.2582}} & \textcolor{darkgreen}{\textbf{0.9378}} & \textcolor{darkgreen}{\textbf{0.3543}} & \textcolor{darkred}{\textbf{0.6718}} & \textcolor{darkred}{\textbf{0.6783}} & \textcolor{darkred}{\textbf{0.9254}} & 0.7148 & 4/9 \\ \hline
Perspective(H) ShiftScaleRotate(L) & \textcolor{darkred}{\textbf{0.9342}} & \textcolor{darkgreen}{\textbf{0.8218}} & \textcolor{darkgreen}{\textbf{0.8432}} & \textcolor{darkgreen}{\textbf{0.3965}} & \textcolor{darkgreen}{\textbf{0.9374}} & \textcolor{darkred}{\textbf{0.3480}} & \textcolor{darkred}{\textbf{0.5777}} & \textcolor{darkred}{\textbf{0.6492}} & \textcolor{darkgreen}{\textbf{0.9257}} & 0.7148 & 5/9 \\ \hline
Perspective(H) ShiftScaleRotate(H) & \textcolor{darkred}{\textbf{0.9351}} & \textcolor{darkgreen}{\textbf{0.7980}} & \textcolor{darkgreen}{\textbf{0.8422}} & \textcolor{darkgreen}{\textbf{0.4272}} & \textcolor{darkgreen}{\textbf{0.9367}} & \textcolor{darkred}{\textbf{0.3539}} & \textcolor{darkgreen}{\textbf{0.5998}} & \textcolor{darkred}{\textbf{0.6005}} & \textcolor{darkred}{\textbf{0.9257}} & 0.7132 & 5/9 \\ \hline
GridDistortion(H) Perspective(C1) & \textcolor{darkred}{\textbf{0.9349}} & \textcolor{darkgreen}{\textbf{0.8010}} & \textcolor{darkgreen}{\textbf{0.8364}} & \textcolor{darkred}{\textbf{0.2837}} & \textcolor{darkred}{\textbf{0.9380}} & \textcolor{darkgreen}{\textbf{0.3906}} & \textcolor{darkred}{\textbf{0.6943}} & \textcolor{darkred}{\textbf{0.6084}} & \textcolor{darkgreen}{\textbf{0.9263}} & 0.7126 & 4/9 \\ \hline
GridDistortion(C3) Perspective(H) & \textcolor{darkred}{\textbf{0.9351}} & \textcolor{darkgreen}{\textbf{0.8139}} & \textcolor{darkgreen}{\textbf{0.8396}} & \textcolor{darkgreen}{\textbf{0.2485}} & \textcolor{darkgreen}{\textbf{0.9375}} & \textcolor{darkgreen}{\textbf{0.3405}} & \textcolor{darkred}{\textbf{0.6907}} & \textcolor{darkred}{\textbf{0.6738}} & \textcolor{darkred}{\textbf{0.9257}} & 0.7117 & 5/9 \\ \hline
GridDistortion(C2) ShiftScaleRotate(L) & \textcolor{darkred}{\textbf{0.9347}} & \textcolor{darkred}{\textbf{0.6700}} & \textcolor{darkgreen}{\textbf{0.8103}} & \textcolor{darkred}{\textbf{0.3269}} & \textcolor{darkgreen}{\textbf{0.9379}} & \textcolor{darkgreen}{\textbf{0.4268}} & \textcolor{darkred}{\textbf{0.7155}} & \textcolor{darkred}{\textbf{0.6221}} & \textcolor{darkgreen}{\textbf{0.9266}} & 0.7079 & 4/9 \\ \hline
Perspective(H) ShiftScaleRotate(C1) & \textcolor{darkgreen}{\textbf{0.9356}} & \textcolor{darkgreen}{\textbf{0.7852}} & \textcolor{darkgreen}{\textbf{0.8378}} & \textcolor{darkgreen}{\textbf{0.4658}} & \textcolor{darkred}{\textbf{0.9355}} & \textcolor{darkgreen}{\textbf{0.3804}} & \textcolor{darkred}{\textbf{0.6041}} & \textcolor{darkred}{\textbf{0.4929}} & \textcolor{darkgreen}{\textbf{0.9254}} & 0.7070 & 6/9 \\ \hline
GridDistortion(H) ShiftScaleRotate(L) & \textcolor{darkgreen}{\textbf{0.9351}} & \textcolor{darkred}{\textbf{0.6581}} & \textcolor{darkgreen}{\textbf{0.8077}} & \textcolor{darkred}{\textbf{0.2030}} & \textcolor{darkgreen}{\textbf{0.9386}} & \textcolor{darkgreen}{\textbf{0.4098}} & \textcolor{darkred}{\textbf{0.7527}} & \textcolor{darkgreen}{\textbf{0.7297}} & \textcolor{darkgreen}{\textbf{0.9263}} & 0.7068 & 6/9 \\ \hline
Affine(C1) Perspective(H) & \textcolor{darkred}{\textbf{0.9333}} & \textcolor{darkgreen}{\textbf{0.7972}} & \textcolor{darkgreen}{\textbf{0.8371}} & \textcolor{darkred}{\textbf{0.3179}} & \textcolor{darkred}{\textbf{0.9356}} & \textcolor{darkred}{\textbf{0.3642}} & \textcolor{darkred}{\textbf{0.6464}} & \textcolor{darkred}{\textbf{0.4930}} & \textcolor{darkgreen}{\textbf{0.9252}} & 0.6944 & 3/9 \\ \hline
Perspective(C1) ShiftScaleRotate(H) & \textcolor{darkgreen}{\textbf{0.9356}} & \textcolor{darkgreen}{\textbf{0.7830}} & \textcolor{darkgreen}{\textbf{0.8422}} & \textcolor{darkgreen}{\textbf{0.3389}} & \textcolor{darkgreen}{\textbf{0.9375}} & \textcolor{darkgreen}{\textbf{0.3472}} & \textcolor{darkgreen}{\textbf{0.4347}} & \textcolor{darkred}{\textbf{0.5119}} & \textcolor{darkgreen}{\textbf{0.9256}} & 0.6730 & 8/9 \\ \hline
NONE & \textcolor{darkred}{\textbf{0.9324}} & \textcolor{darkred}{\textbf{0.6407}} & \textcolor{darkred}{\textbf{0.7628}} & \textcolor{darkred}{\textbf{0.1881}} & \textcolor{darkred}{\textbf{0.9339}} & \textcolor{darkred}{\textbf{0.2317}} & \textcolor{darkred}{\textbf{0.7360}} & \textcolor{darkred}{\textbf{0.6421}} & \textcolor{darkred}{\textbf{0.9241}} & 0.6658 & 0/9 \\ \hline

\end{longtable}

\begin{figure}[H]
    \centering

    \begin{subfigure}[b]{0.49\textwidth}
        \centering
        \includegraphics[
            height=0.95\textheight,
            keepaspectratio
        ]{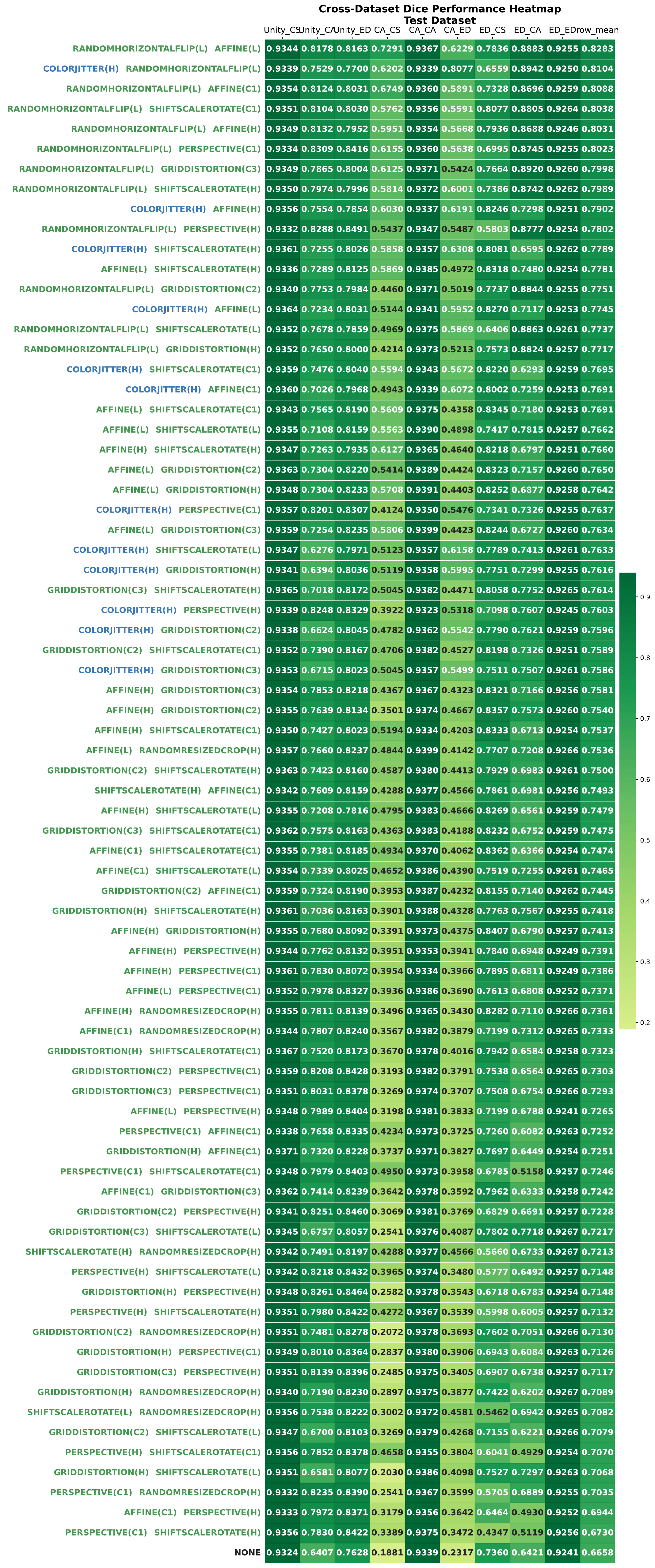}
        \caption{Raw}
        \label{fig:left_plot}
    \end{subfigure}
    \hfill
    \begin{subfigure}[b]{0.49\textwidth}
        \centering
        \includegraphics[
            height=0.95\textheight,
            keepaspectratio
        ]{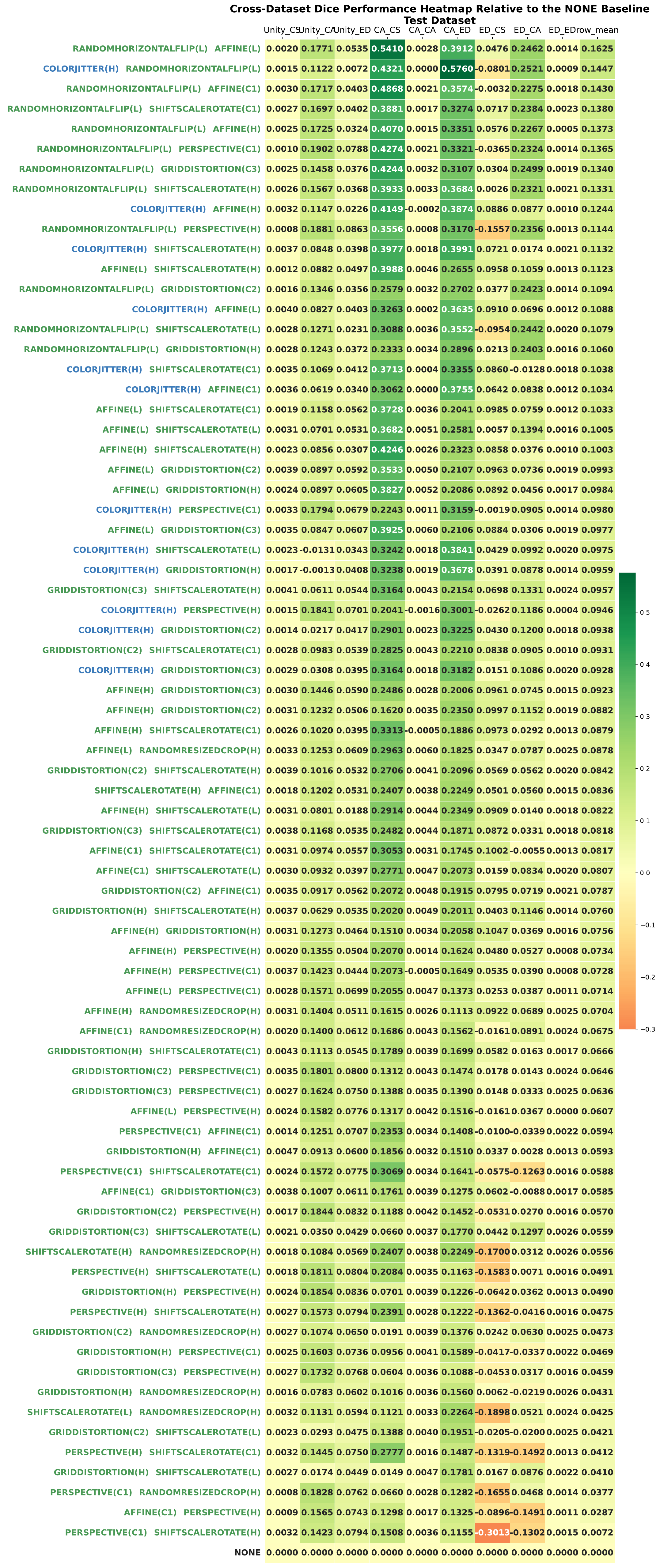}
        \caption{Relative to Baseline}
        \label{fig:right_plot}
    \end{subfigure}

    \caption{Cross-Dataset Dice Performance and Relative Gains Over the NONE for Pairwise Augmentations}
    \label{fig:comparison}
\end{figure}

\begin{figure}[H]
    \centering
    
    \begin{subfigure}[b]{0.49\textwidth}
        \centering
        \includegraphics[
            height=0.95\textheight,
            keepaspectratio
        ]{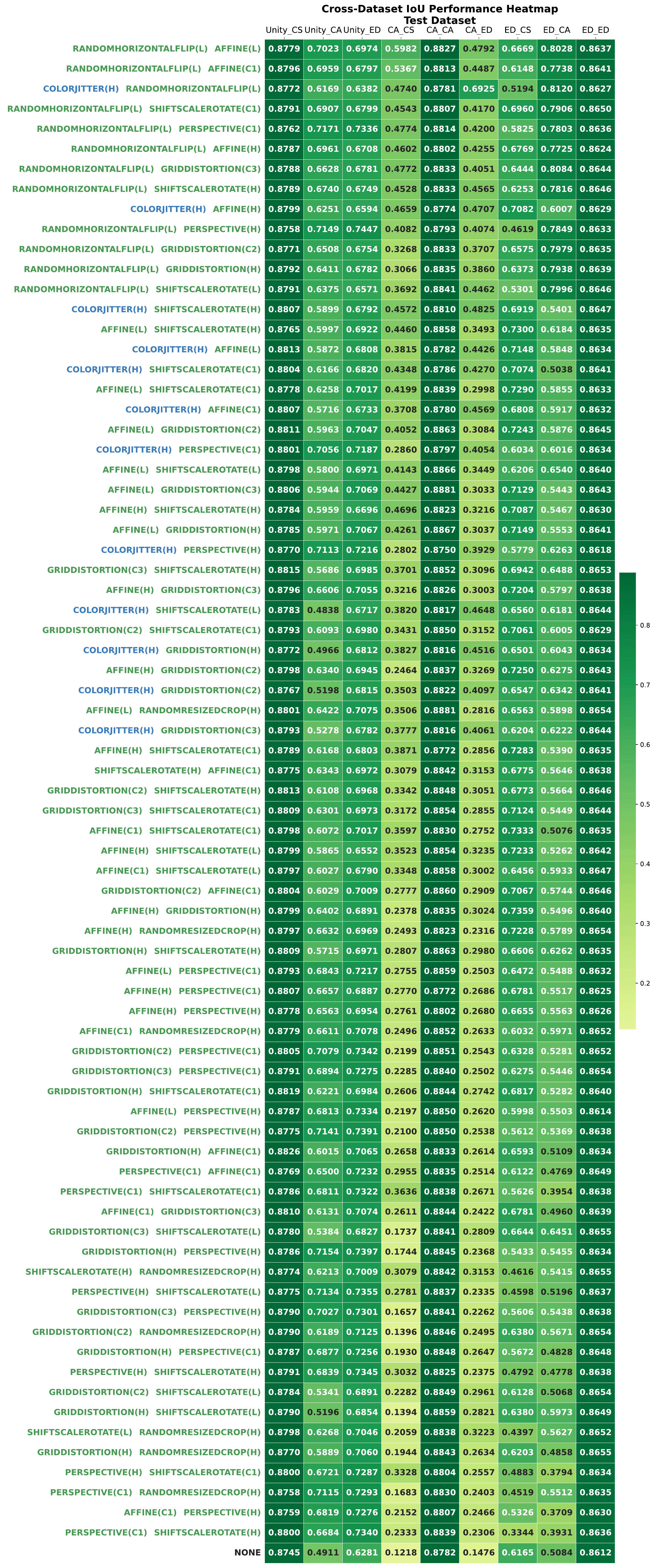} 
        \caption{Raw}
        \label{fig:left_plot}
    \end{subfigure}
    \hfill 
    \begin{subfigure}[b]{0.49\textwidth}
        \centering
        \includegraphics[
            height=0.95\textheight,
            keepaspectratio
        ]{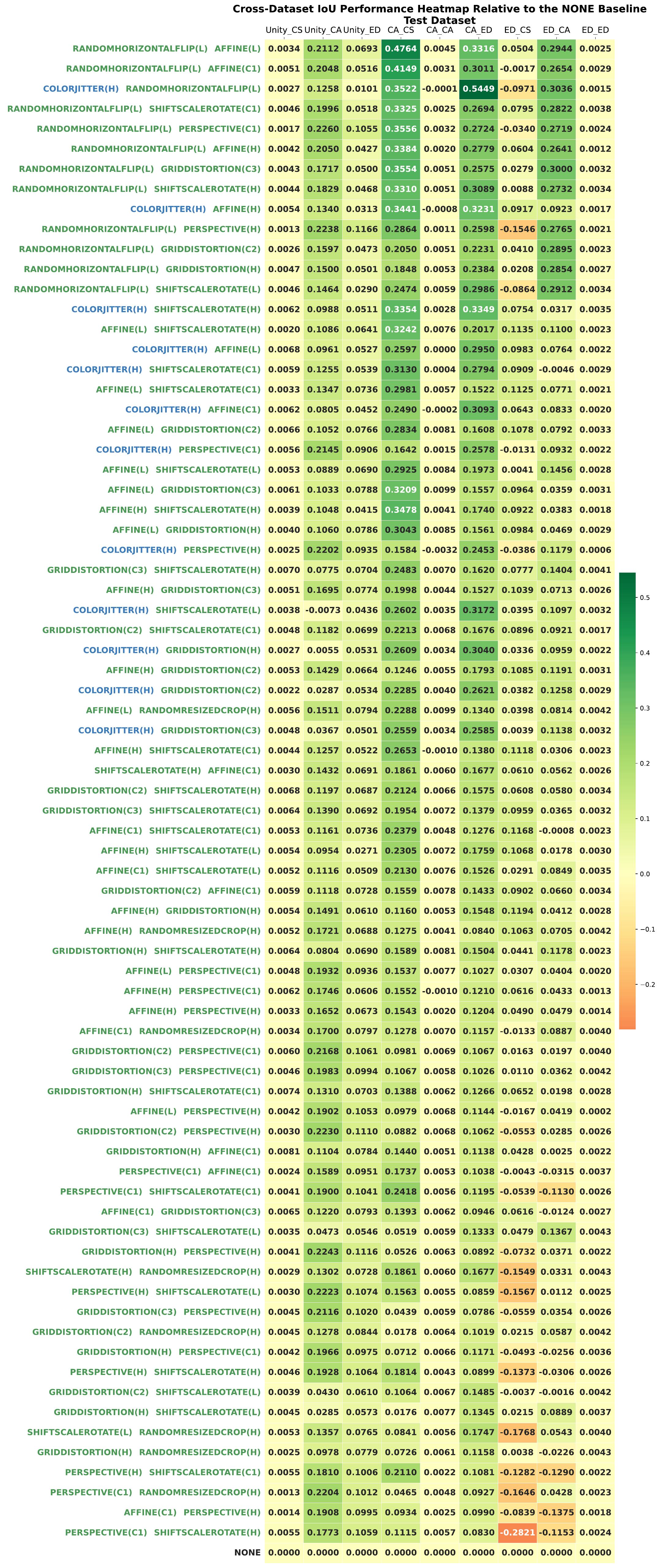}
        \caption{Relative to Baseline}
        \label{fig:right_plot}
    \end{subfigure}

    \caption{Cross-Dataset IoU Performance and Relative Gains Over the NONE for Pairwise Augmentations}
    \label{fig:comparison}
\end{figure}

\newgeometry{left=1cm,right=1cm,top=1.5cm,bottom=1.5cm}

\section{Augmentation Settings}
This appendix provides a concise overview of the augmentation settings used in this study. Each setting is denoted by a predefined code (L, H, C1, C2, C3, CH) to enable clear and consistent reference throughout the manuscript. Representative examples for each configuration are illustrated using images from the Unity dataset. Figures 19-111 present the visual effects of these augmentation settings.
\subsection{Affine}

\begin{figure}[H]
\centering
\includegraphics[width=\linewidth]{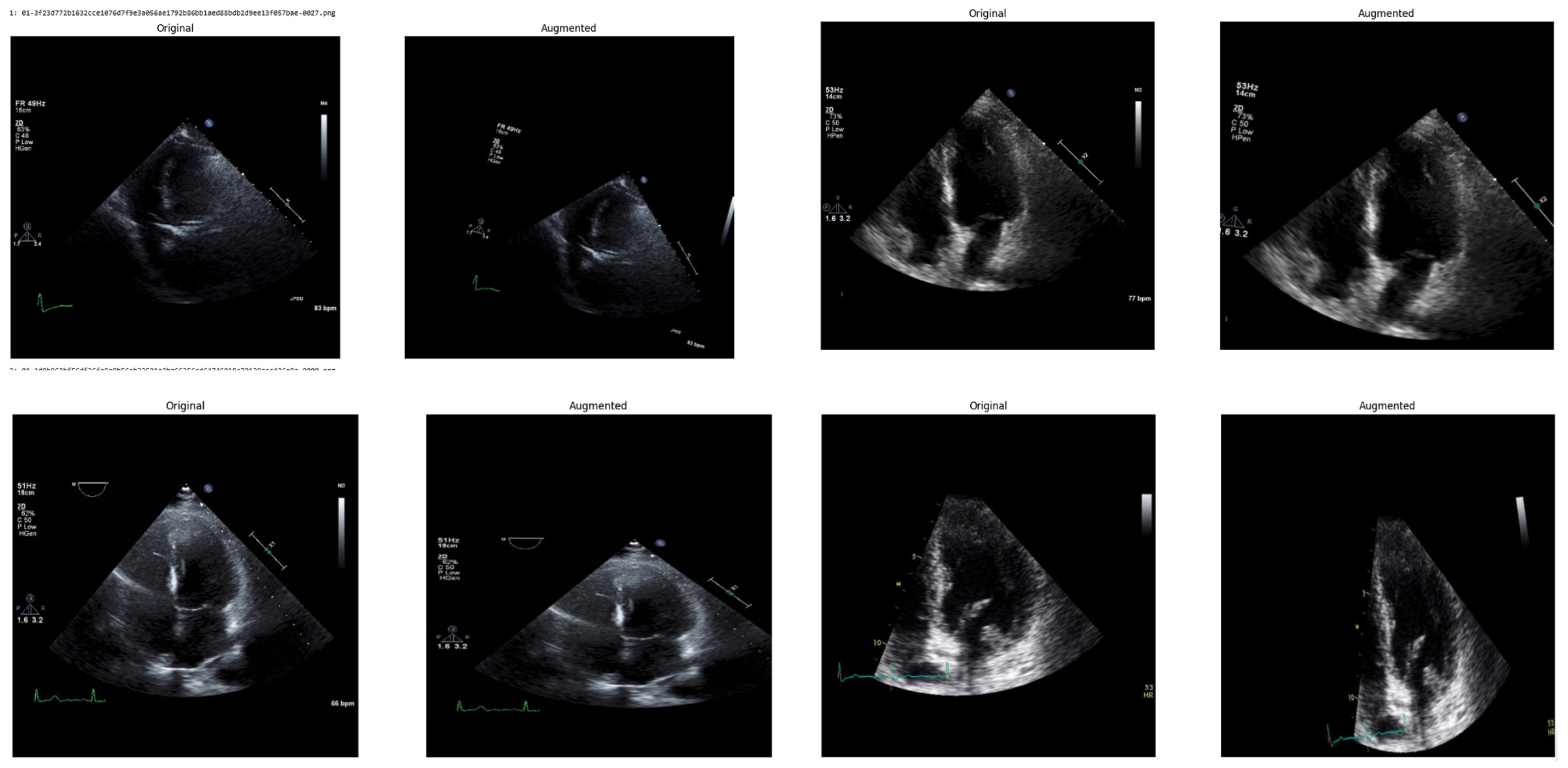}
\caption{
L: A.Affine(rotate=(-15, 15), translate\_percent=(0.1, 0.1), scale=(0.7, 1.3), p=1.0)
}
\label{fig:example}
\end{figure}

\begin{figure}[H]
\centering
\includegraphics[width=\linewidth]{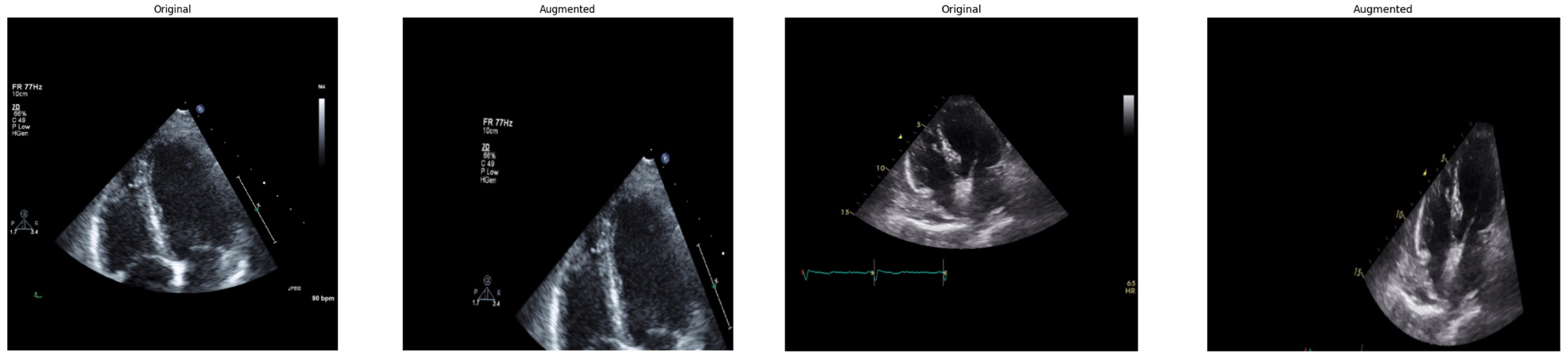}
\caption{
H: A.Affine(rotate=(-30, 30), translate\_percent=(0.2, 0.2), scale=(0.6, 1.5), p=0.7)
}
\label{fig:example}
\end{figure}

\begin{figure}[H]
\centering
\includegraphics[width=\linewidth]{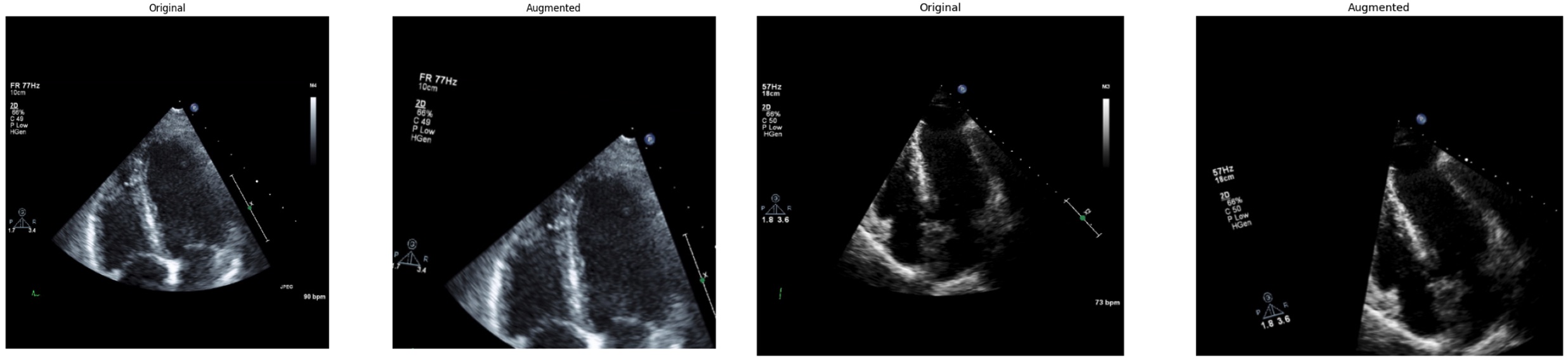}
\caption{
C1: A.Affine(rotate=(-25, 25), translate\_percent=(0.15, 0.15), scale=(0.75, 1.35), p=0.85)
}
\label{fig:example}
\end{figure}

\subsection{CLAHE}

\begin{figure}[H]
\centering
\includegraphics[width=\linewidth]{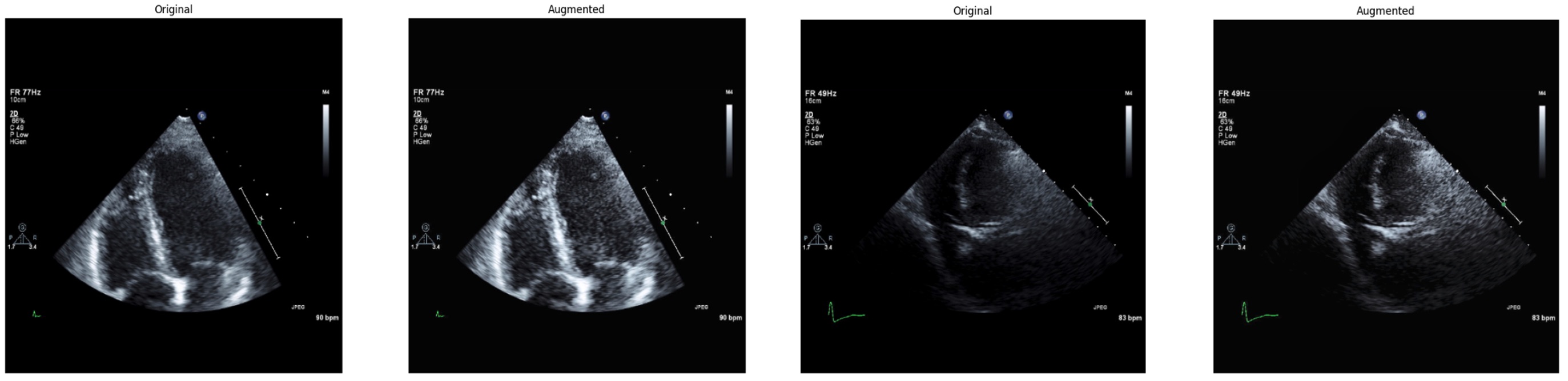}
\caption{
L: A.CLAHE(clip\_limit=2.0, tile\_grid\_size=(8, 8), p=0.2)
}
\label{fig:example}
\end{figure}

\begin{figure}[H]
\centering
\includegraphics[width=\linewidth]{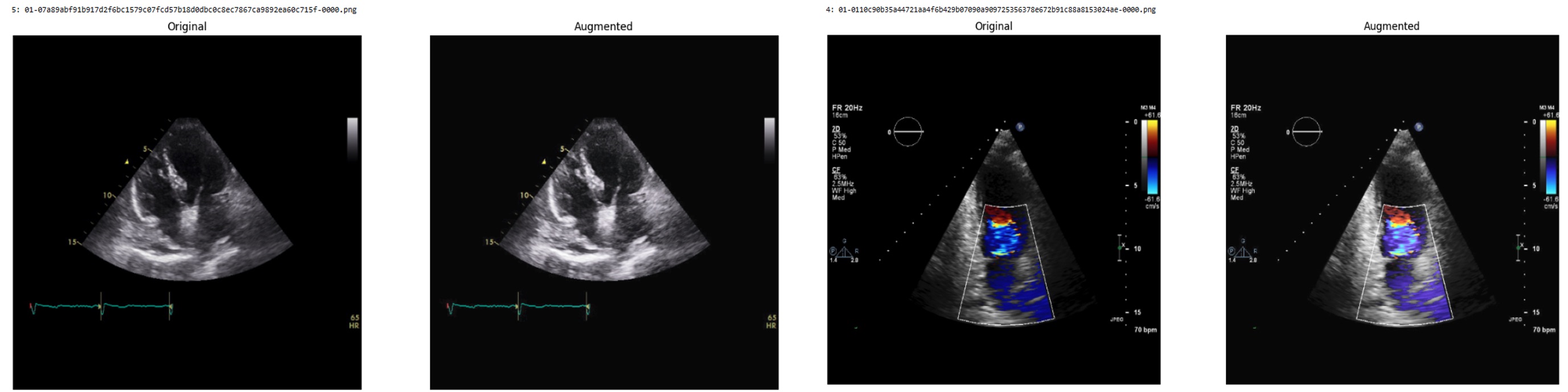}
\caption{
H: A.CLAHE(clip\_limit=4.0, tile\_grid\_size=(4, 4), p=0.4)
}
\label{fig:example}
\end{figure}

\begin{figure}[H]
\centering
\includegraphics[width=\linewidth]{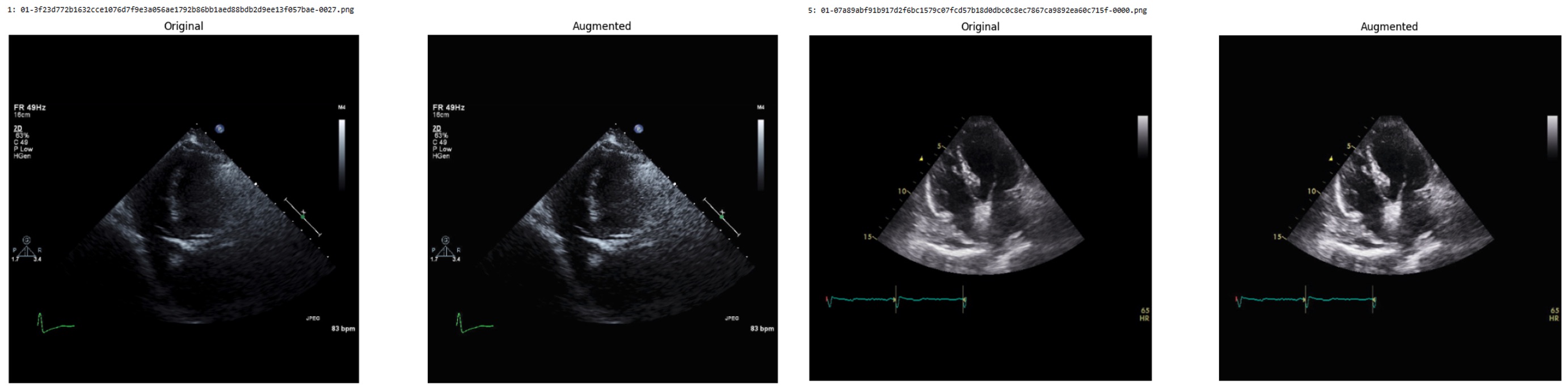}
\caption{
C1: A.CLAHE(clip\_limit=1.8, tile\_grid\_size=(8, 8), p=0.25)
}
\label{fig:example}
\end{figure}

\subsection{ColorJitter}

\begin{figure}[H]
\centering
\includegraphics[width=\linewidth]{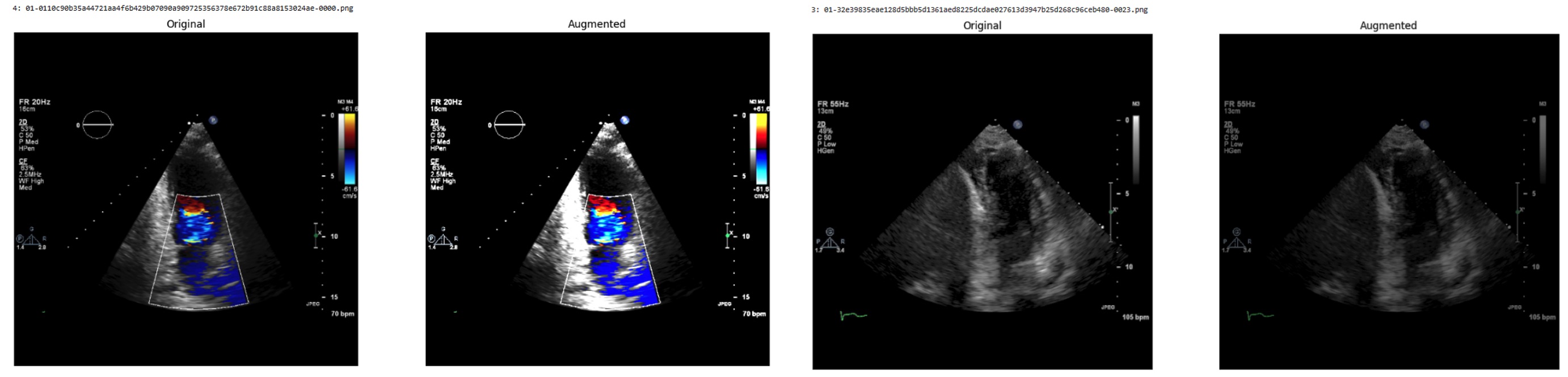}
\caption{
L: ColorJitter(brightness=0.8, contrast=0.8, saturation=0.0, hue=0.0, p=0.8)
}
\label{fig:example}
\end{figure}

\begin{figure}[H]
\centering
\includegraphics[width=\linewidth]{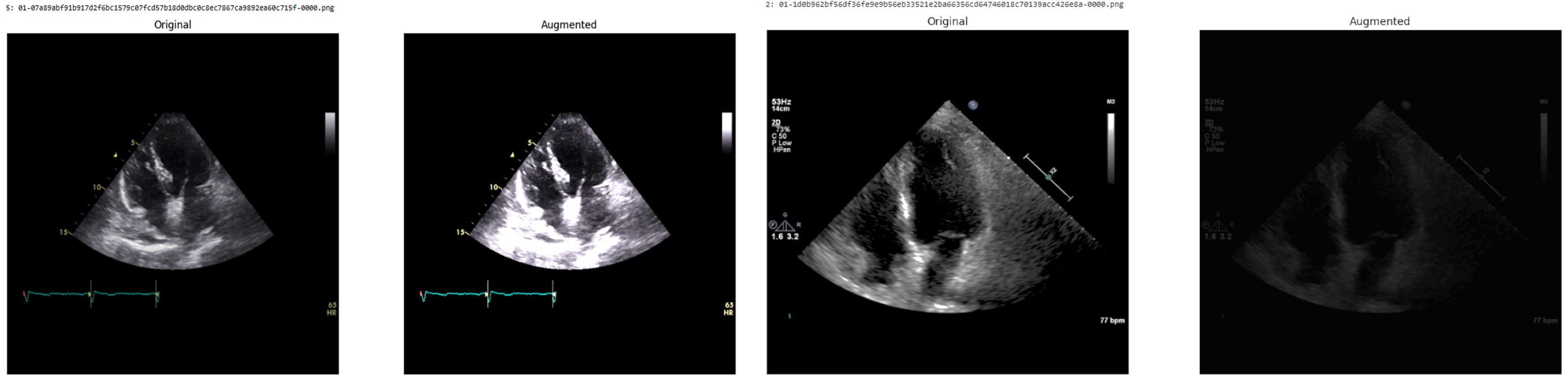}
\caption{
H: A.ColorJitter(brightness=1.0, contrast=1.0, saturation=0.0, hue=0.0, p=0.8))
}
\label{fig:example}
\end{figure}

\begin{figure}[H]
\centering
\includegraphics[width=\linewidth]{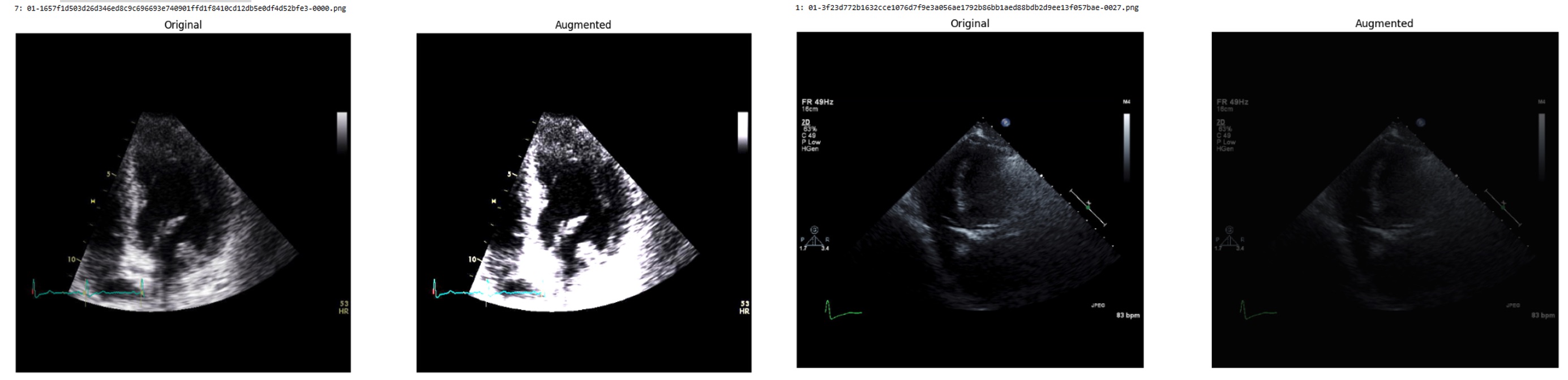}
\caption{
C1: A.ColorJitter(brightness=1.5, contrast=1.5, saturation=0.0, hue=0.0, p=0.35)
}
\label{fig:example}
\end{figure}

\begin{figure}[H]
\centering
\includegraphics[width=\linewidth]{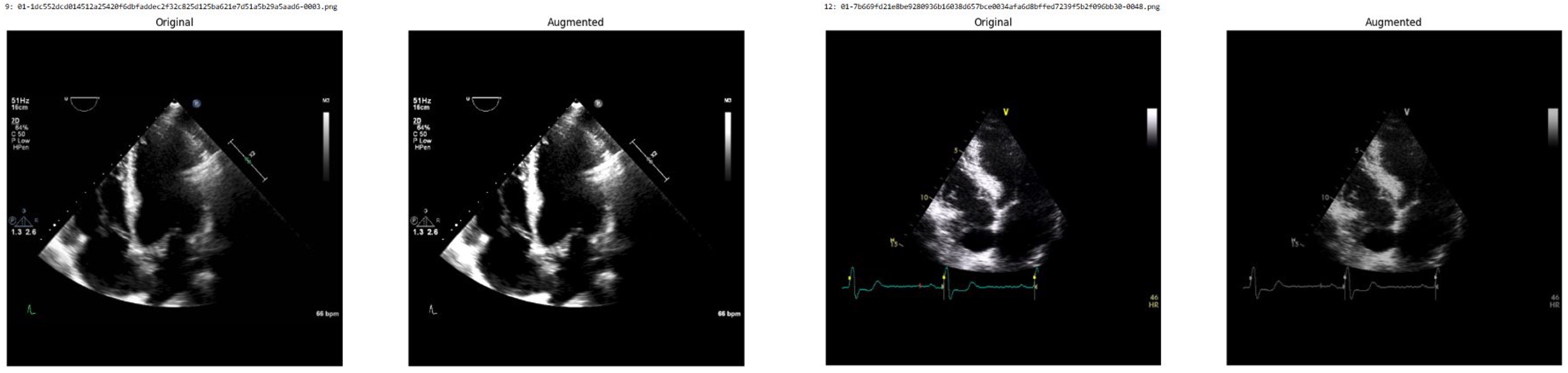}
\caption{
C2: A.ColorJitter(brightness=0.2, contrast=0.2, saturation=0.0, hue=0.0, p=0.5)
}
\label{fig:example}
\end{figure}

\subsection{ElasticTransform}

\begin{figure}[H]
\centering
\includegraphics[width=\linewidth]{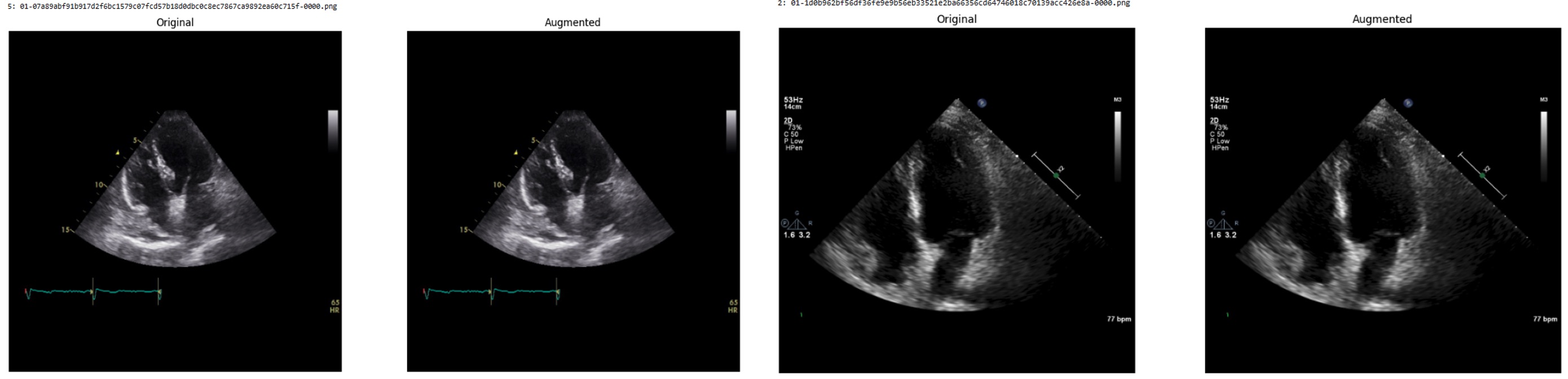}
\caption{
L: A.ElasticTransform(alpha=5.0, sigma=10.0, alpha\_affine=2.0, p=0.15)
}
\label{fig:example}
\end{figure}

\begin{figure}[H]
\centering
\includegraphics[width=\linewidth]{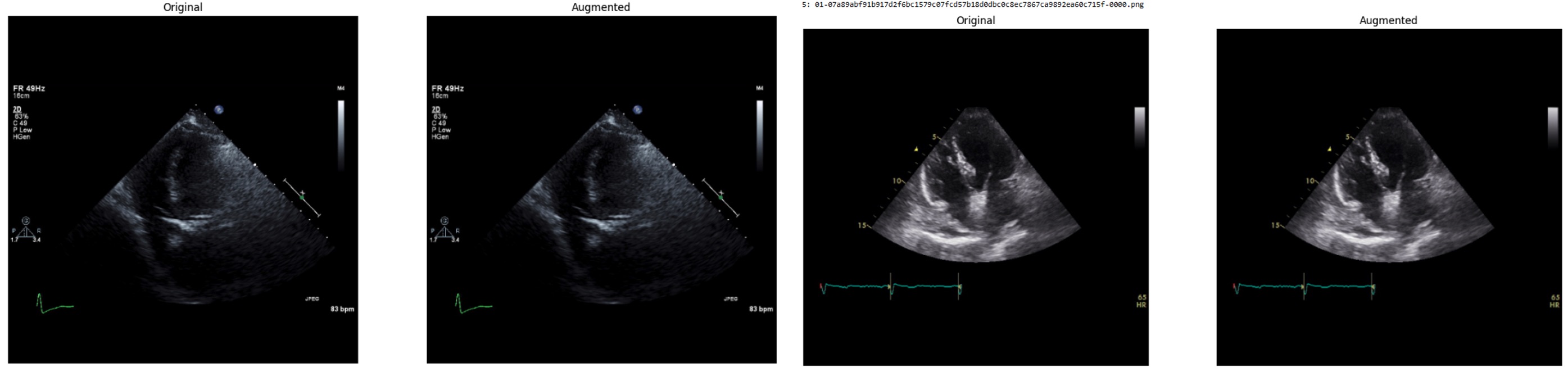}
\caption{
H: A.ElasticTransform( alpha=30.0, sigma=20.0, alpha\_affine=10.0, p=0.25 )p=0.15)
}
\label{fig:example}
\end{figure}

\begin{figure}[H]
\centering
\includegraphics[width=\linewidth]{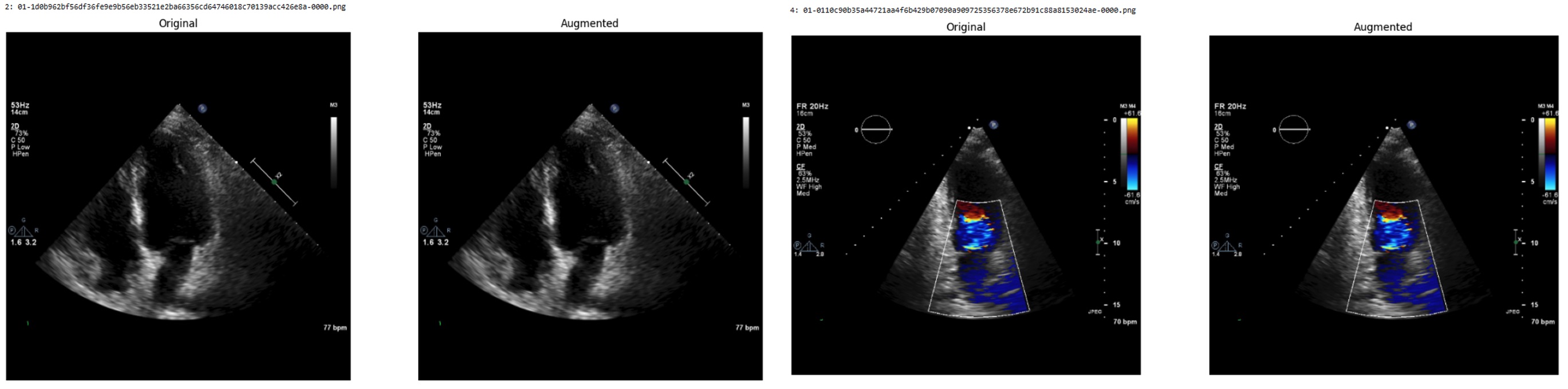}
\caption{
C1: A.ElasticTransform( alpha=7.5, sigma=13.0, alpha\_affine=2.0, p=0.15 )
}
\label{fig:example}
\end{figure}

\subsection{GaussianBlur}

\begin{figure}[H]
\centering
\includegraphics[width=\linewidth]{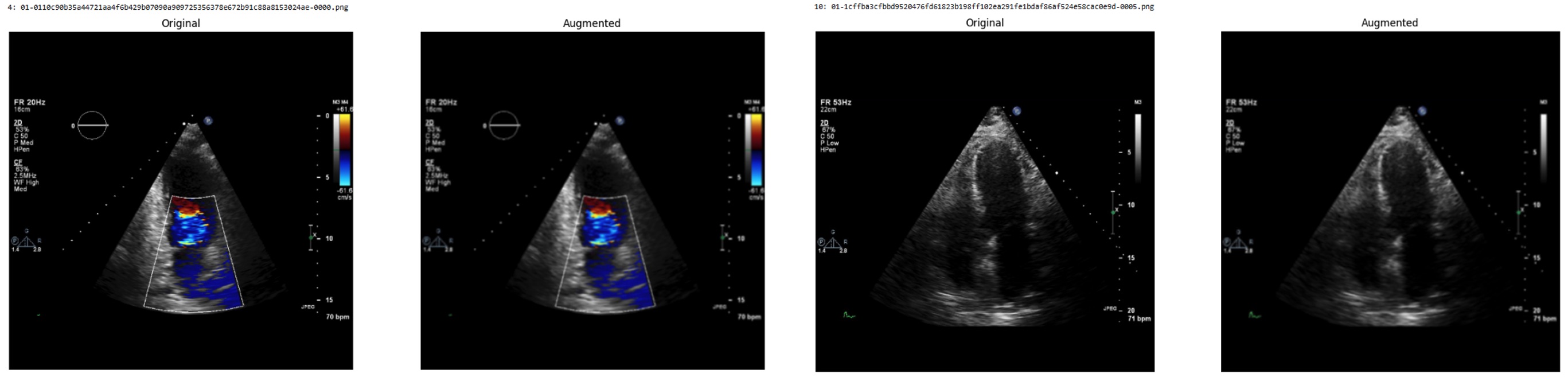}
\caption{
L: A.GaussianBlur(blur\_limit=(3, 3), p=0.15)
}
\label{fig:example}
\end{figure}

\begin{figure}[H]
\centering
\includegraphics[width=\linewidth]{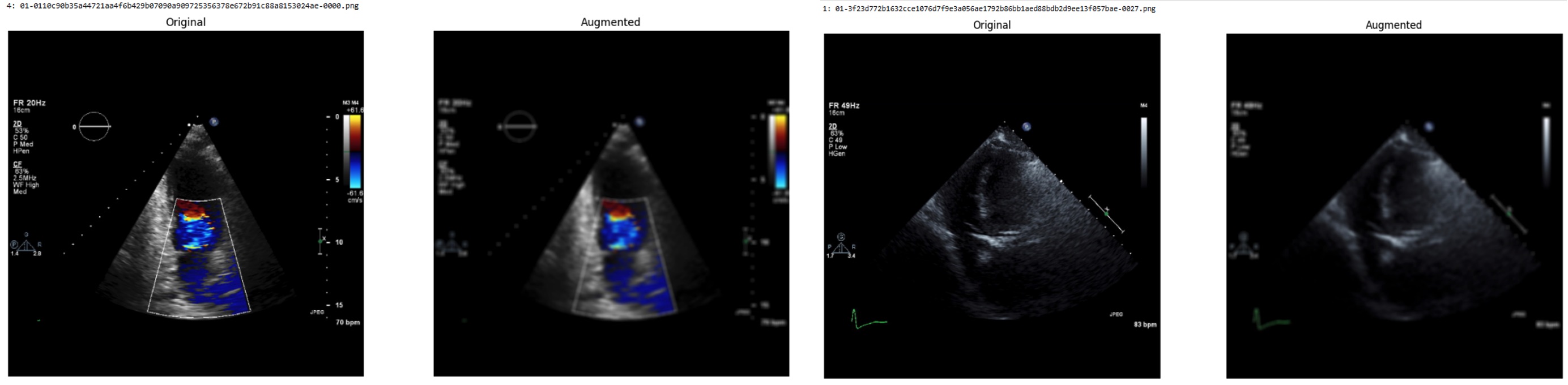}
\caption{
H: A.GaussianBlur(blur\_limit=(5, 9), p=0.35)
}
\label{fig:example}
\end{figure}

\begin{figure}[H]
\centering
\includegraphics[width=\linewidth]{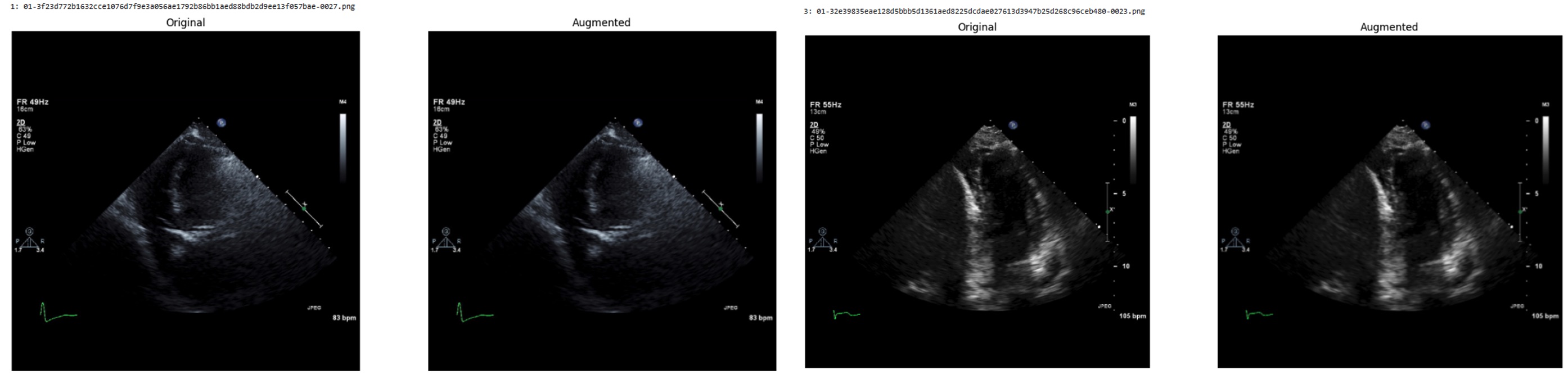}
\caption{
C1: A.GaussianBlur(blur\_limit=(3, 5),sigma\_limit=(0.1, 0.6), p=0.15)
}
\label{fig:example}
\end{figure}

\subsection{GaussNoise}

\begin{figure}[H]
\centering
\includegraphics[width=\linewidth]{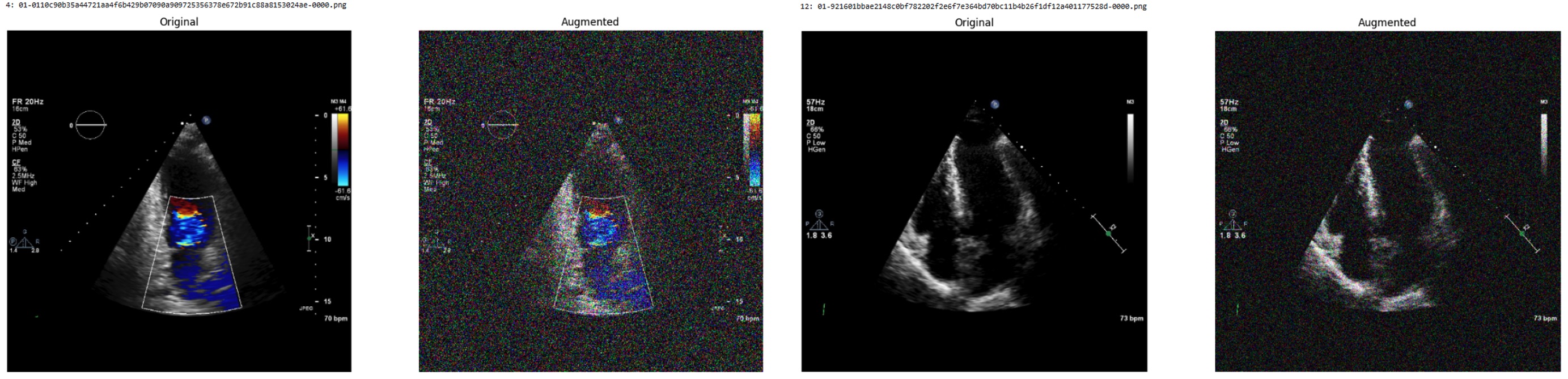}
\caption{
L: A.GaussNoise(var\_limit=(5.0, 20.0), p=0.4)
}
\label{fig:example}
\end{figure}

\begin{figure}[H]
\centering
\includegraphics[width=\linewidth]{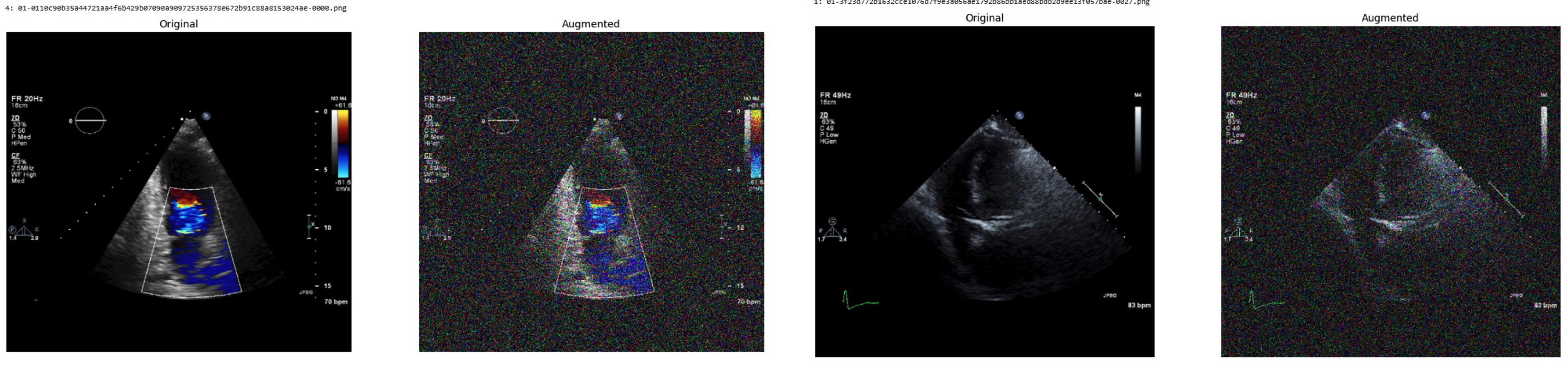}
\caption{
H: A.GaussNoise(var\_limit=(15.0, 50.0), p=0.5)
}
\label{fig:example}
\end{figure}

\begin{figure}[H]
\centering
\includegraphics[width=\linewidth]{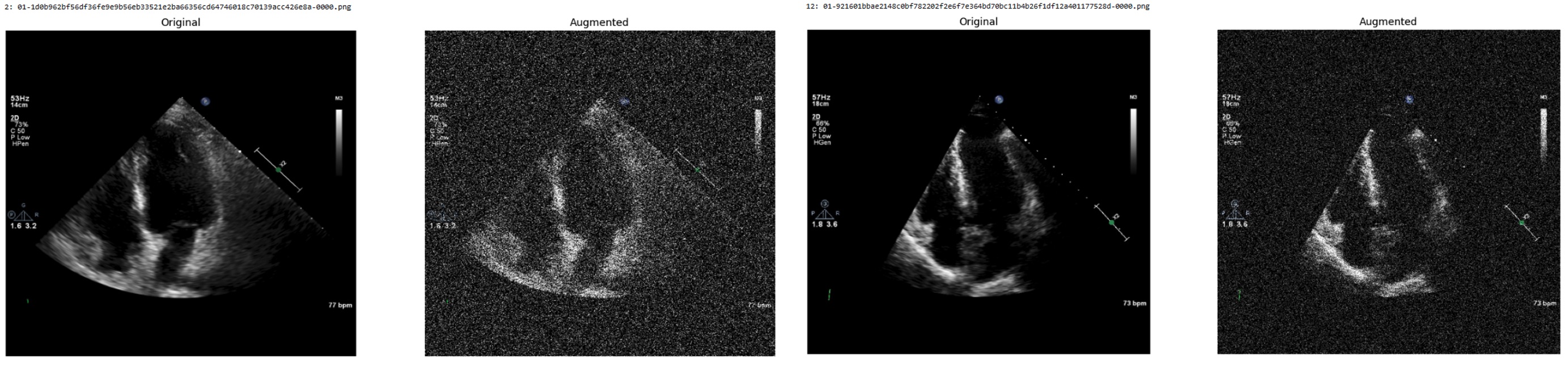}
\caption{
C1: A.GaussNoise(var\_limit=(3.0, 15.0), p=0.35, per\_channel=False))
}
\label{fig:example}
\end{figure}

\begin{figure}[H]
\centering
\includegraphics[width=\linewidth]{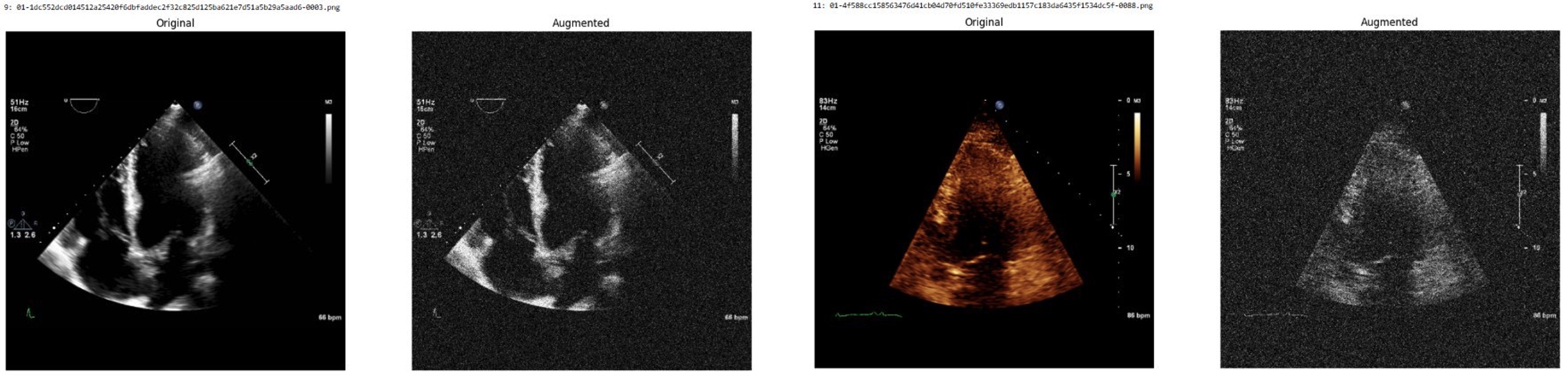}
\caption{
C2: A.GaussNoise(var\_limit=(0.001, 0.001),  p=0.3)
}
\label{fig:example}
\end{figure}

\subsection{MotionBlur}

\begin{figure}[H]
\centering
\includegraphics[width=\linewidth]{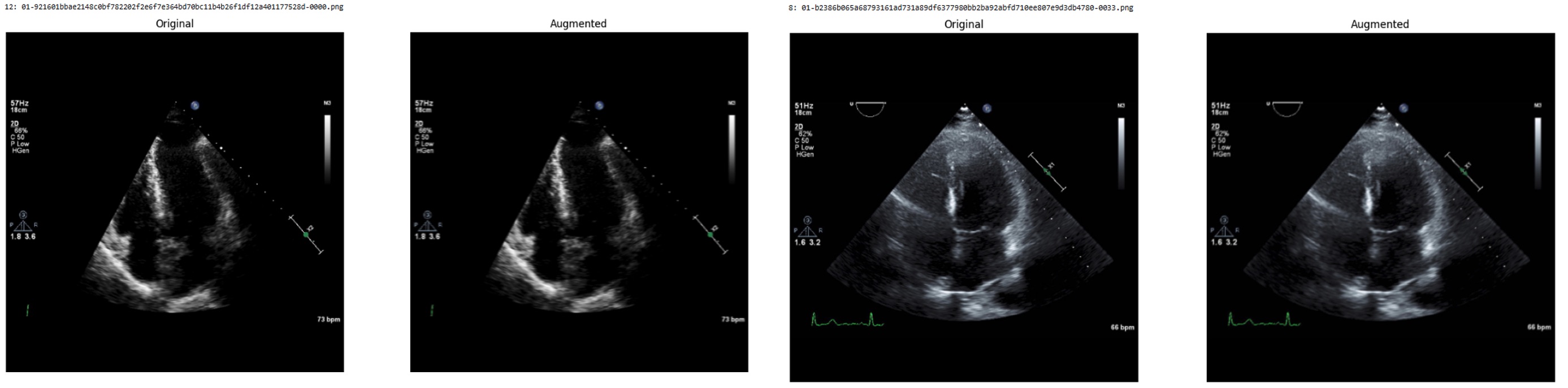}
\caption{
L: A.MotionBlur(blur\_limit=3, p=0.2)
}
\label{fig:example}
\end{figure}

\begin{figure}[H]
\centering
\includegraphics[width=\linewidth]{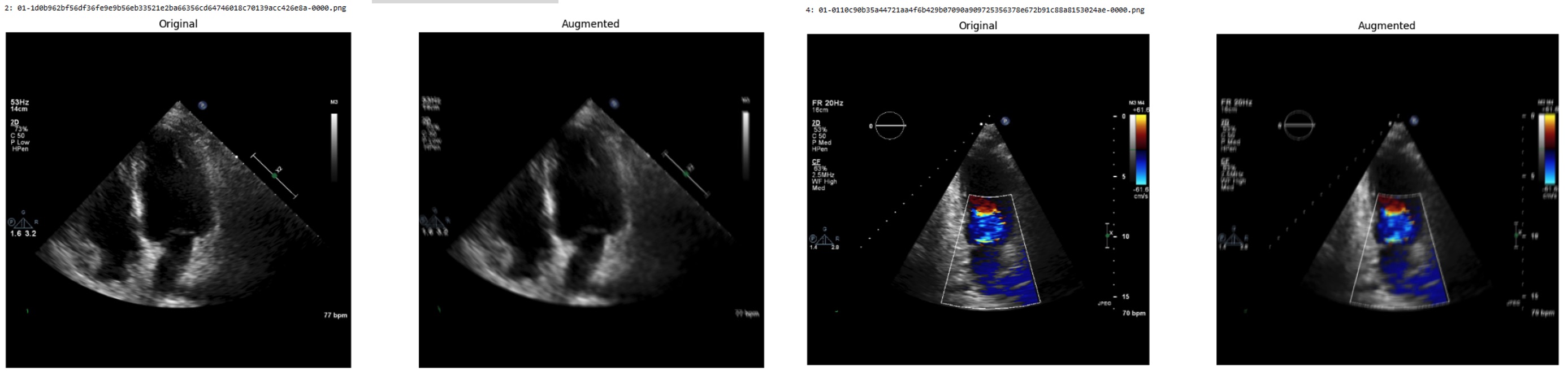}
\caption{
H: A.MotionBlur(blur\_limit=(5, 10), p=0.4))
}
\label{fig:example}
\end{figure}

\begin{figure}[H]
\centering
\includegraphics[width=\linewidth]{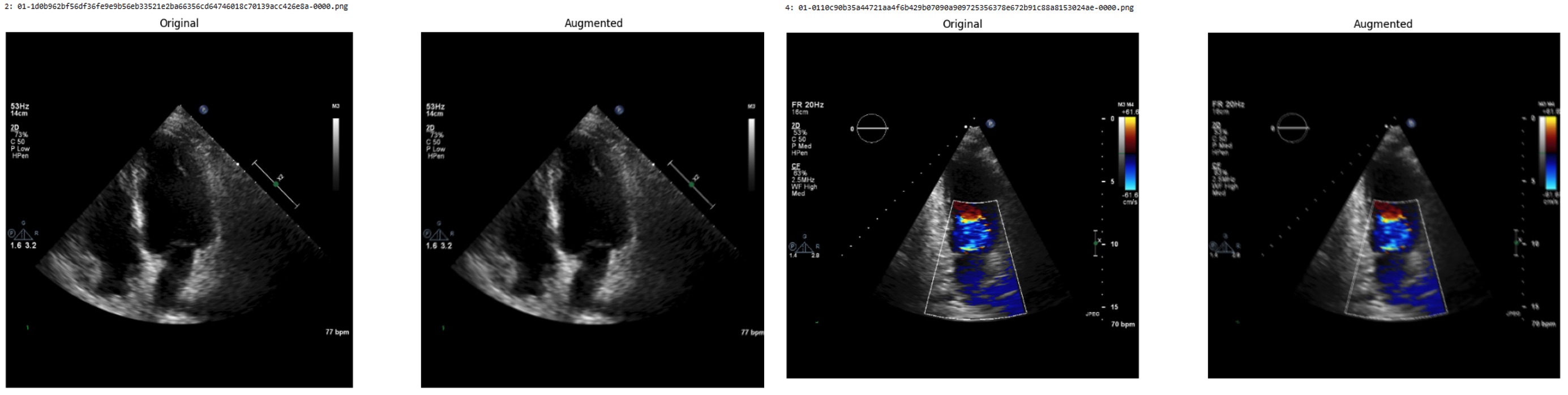}
\caption{
C1: A.MotionBlur(blur\_limit=4, p=0.30)
}
\label{fig:example}
\end{figure}

\subsection{MultiplicativeNoise}

\begin{figure}[H]
\centering
\includegraphics[width=\linewidth]{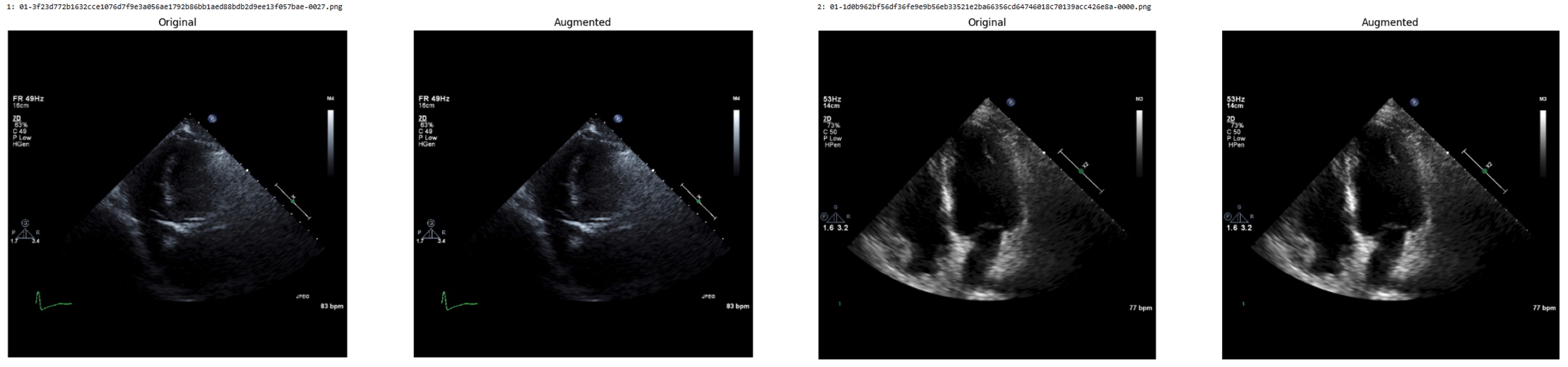}
\caption{
L: A.MultiplicativeNoise(multiplier=(0.9, 1.1), per\_channel=False, p=0.5)
}
\label{fig:example}
\end{figure}

\begin{figure}[H]
\centering
\includegraphics[width=\linewidth]{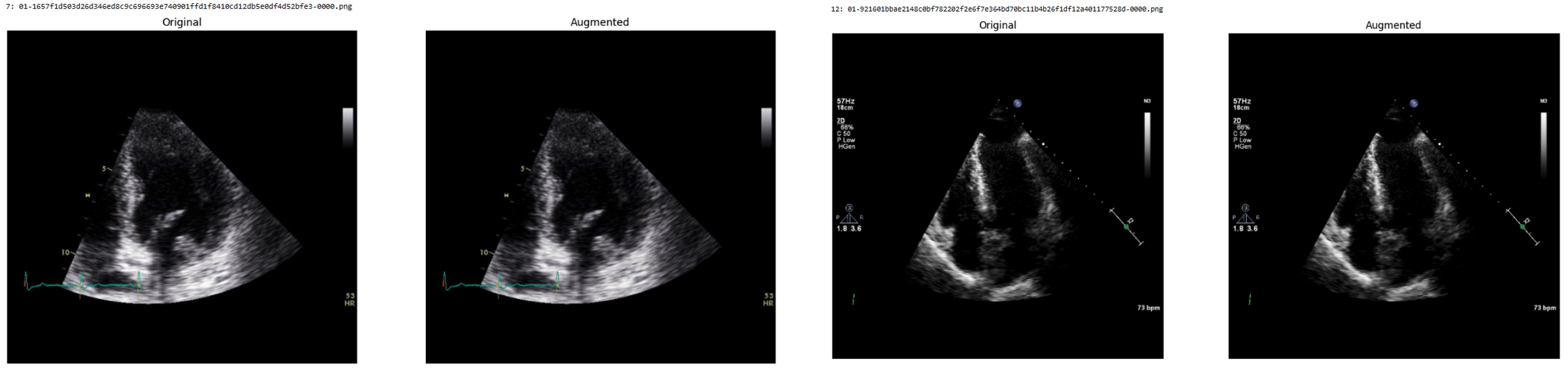}
\caption{
H: A.MultiplicativeNoise(multiplier=(0.8, 1.2), per\_channel=False, p=0.6)
}
\label{fig:example}
\end{figure}

\begin{figure}[H]
\centering
\includegraphics[width=\linewidth]{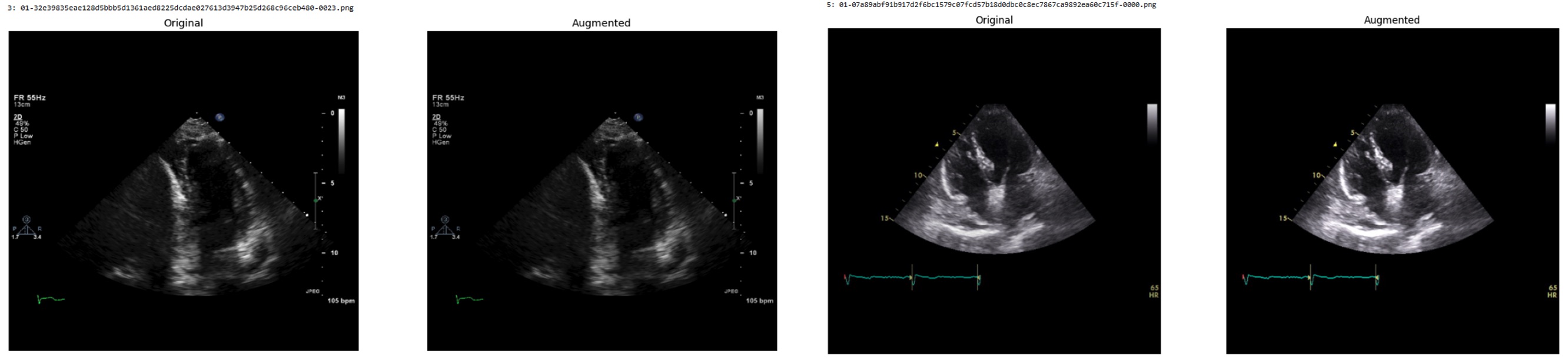}
\caption{
C1: A.MultiplicativeNoise(multiplier=(0.70, 1.3)), per\_channel=False, p=0.7)
}
\label{fig:example}
\end{figure}

\subsection{Perspective}

\begin{figure}[H]
\centering
\includegraphics[width=\linewidth]{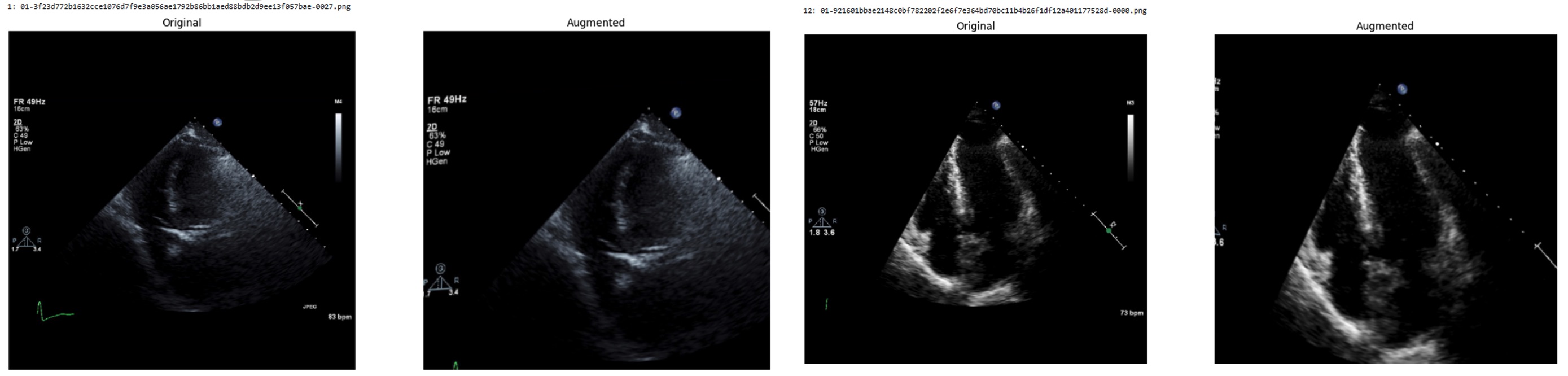}
\caption{
L: A.Perspective(scale=(0.05, 0.1), p=1.0)
}
\label{fig:example}
\end{figure}

\begin{figure}[H]
\centering
\includegraphics[width=\linewidth]{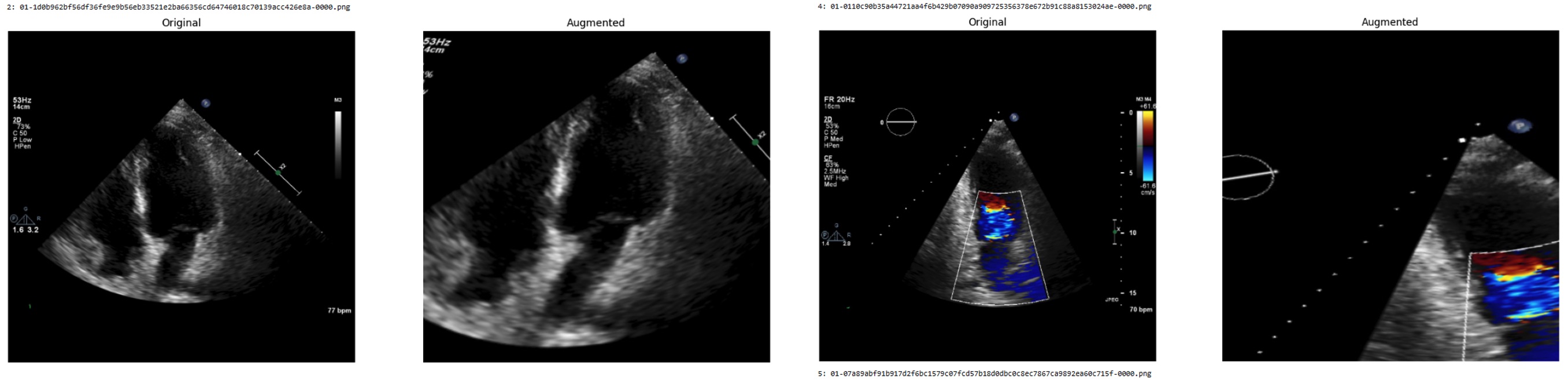}
\caption{
H: A.Perspective(scale=(0.08, 0.20), p=0.7)
}
\label{fig:example}
\end{figure}

\begin{figure}[H]
\centering
\includegraphics[width=\linewidth]{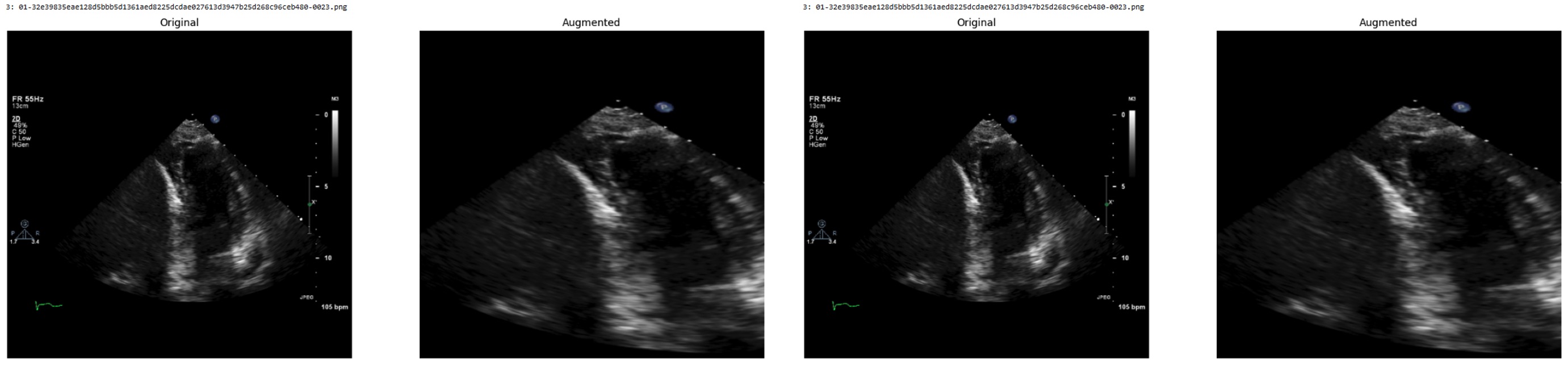}
\caption{
C1: A.Perspective(scale=(0.07, 0.15), p=0.6)
}
\label{fig:example}
\end{figure}

\subsection{RandomBrightnessContrast}

\begin{figure}[H]
\centering
\includegraphics[width=\linewidth]{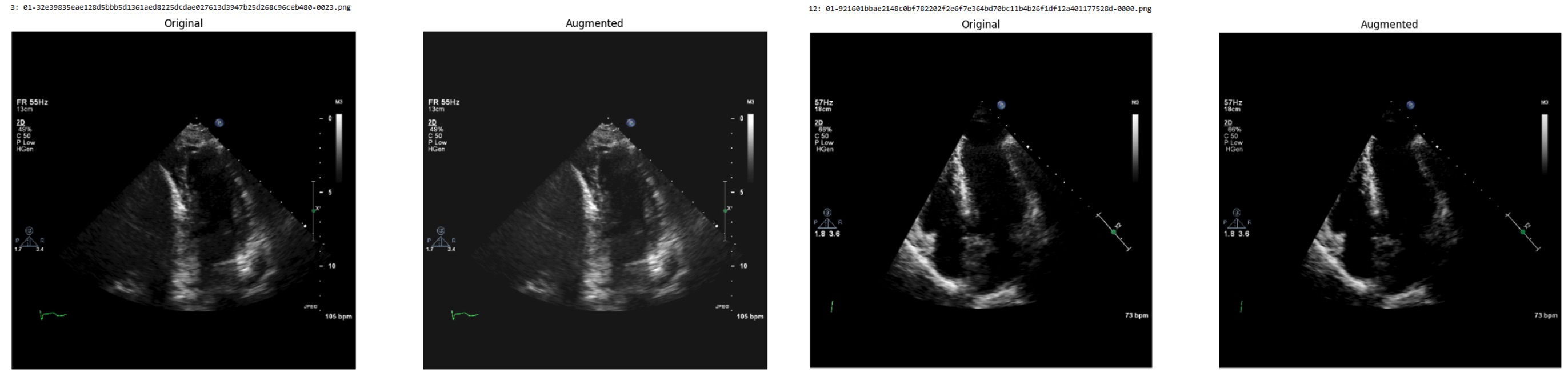}
\caption{
L: A.RandomBrightnessContrast(brightness\_limit=0.10, contrast\_limit=0.10, p=0.6)
}
\label{fig:example}
\end{figure}

\begin{figure}[H]
\centering
\includegraphics[width=\linewidth]{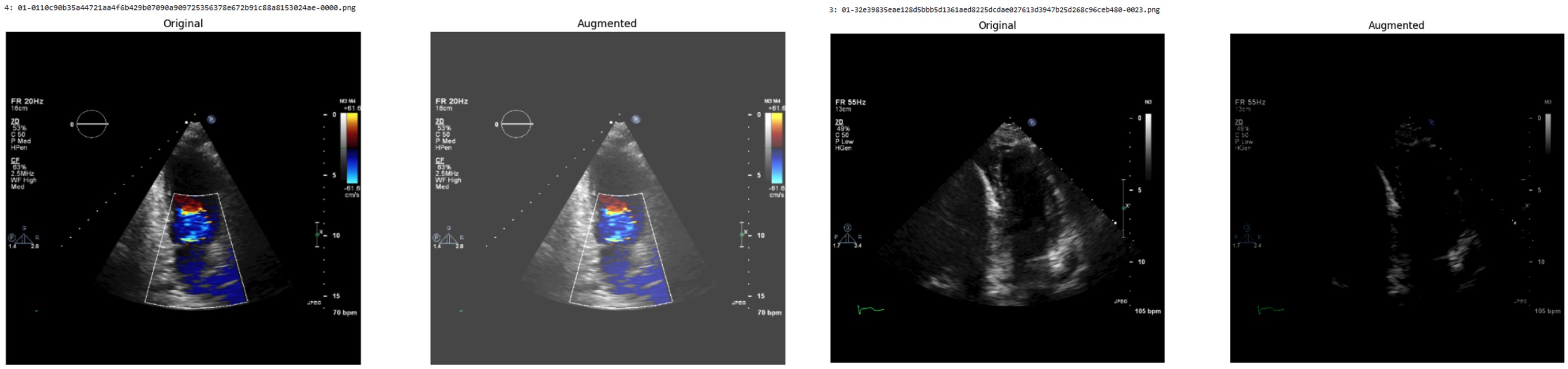}
\caption{
H: A.RandomBrightnessContrast(brightness\_limit=0.4, contrast\_limit=0.4, p=0.7)
}
\label{fig:example}
\end{figure}

\begin{figure}[H]
\centering
\includegraphics[width=\linewidth]{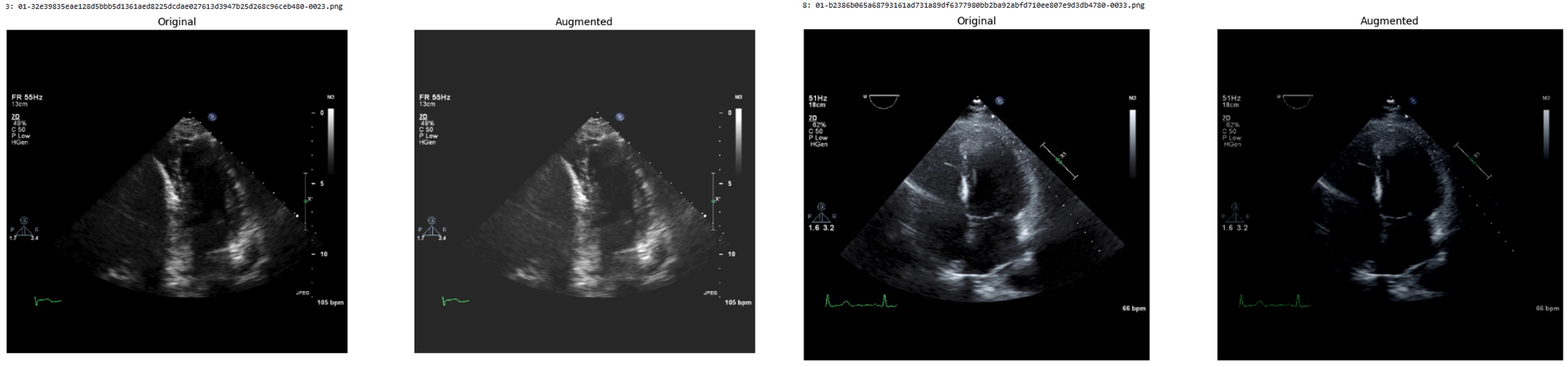}
\caption{\centering
C1: A.RandomBrightnessContrast(brightness\_limit=(-0.25, 0.35), contrast\_limit=(-0.08, 0.12),brightness\_by\_max=True, p=0.45)
}
\label{fig:example}
\end{figure}

\subsection{RandomGamma}

\begin{figure}[H]
\centering
\includegraphics[width=\linewidth]{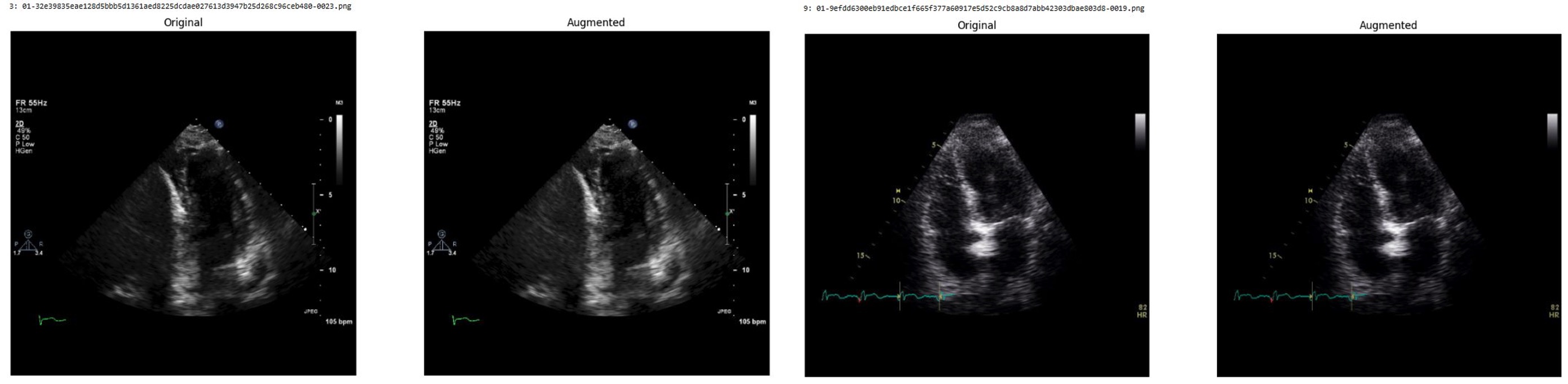}
\caption{
L: A.RandomGamma(gamma\_limit=(80, 120), p=0.4)
}
\label{fig:example}
\end{figure}

\begin{figure}[H]
\centering
\includegraphics[width=\linewidth]{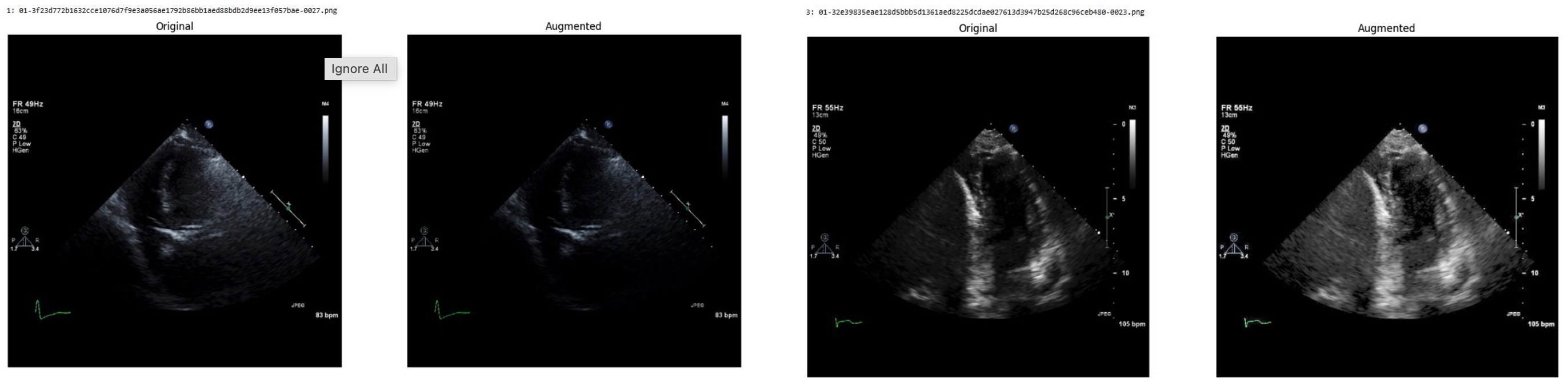}
\caption{
H: A.RandomGamma(gamma\_limit=(40, 160), p=0.5)
}
\label{fig:example}
\end{figure}

\begin{figure}[H]
\centering
\includegraphics[width=\linewidth]{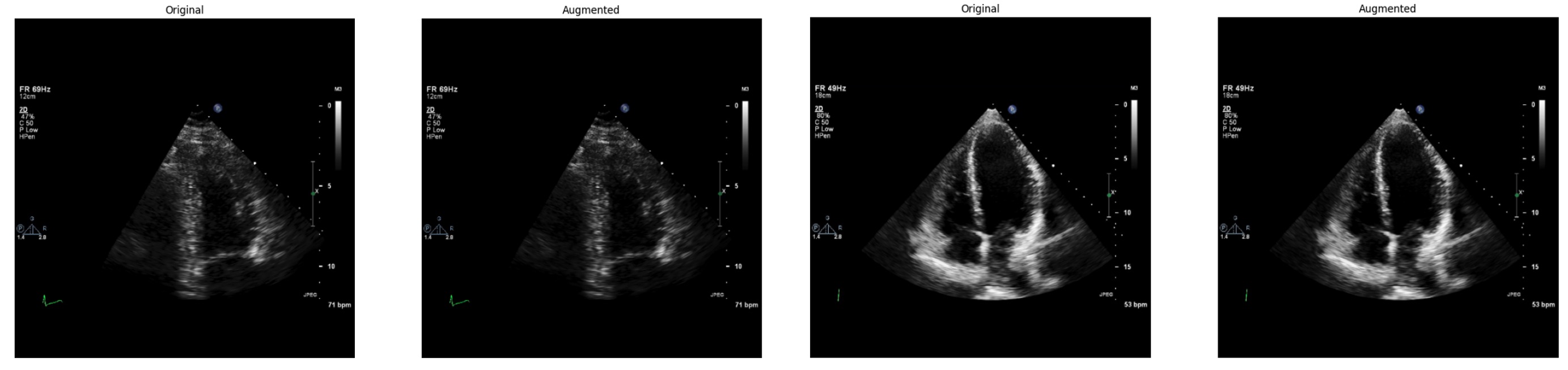}
\caption{
C1: A.RandomGamma(gamma\_limit=(90, 110), p=0.30)
}
\label{fig:example}
\end{figure}

\subsection{HorizontalFlip}

\begin{figure}[H]
\centering
\includegraphics[width=\linewidth]{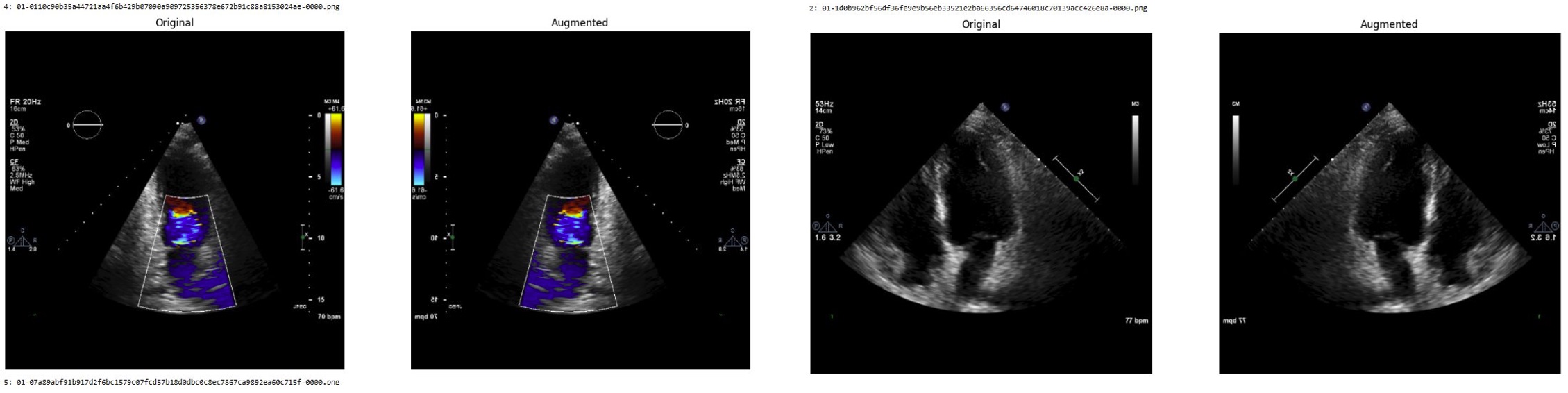}
\caption{
L: A.HorizontalFlip(p=0.5), H: A.HorizontalFlip(p=0.75), A.HorizontalFlip(p=0.30)
}
\label{fig:example}
\end{figure}

\subsection{RandomResizedCrop}

\begin{figure}[H]
\centering
\includegraphics[width=\linewidth]{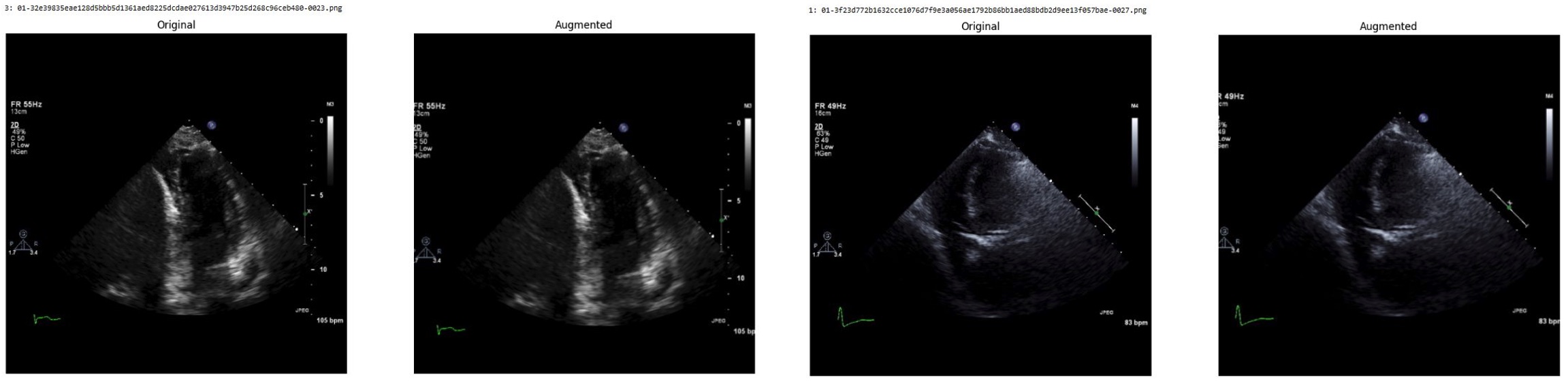}
\caption{
L: A.RandomResizedCrop(size=(512, 512), scale=(0.9, 1.0), ratio=(0.95, 1.05), p=1.0)
}
\label{fig:example}
\end{figure}

\begin{figure}[H]
\centering
\includegraphics[width=\linewidth]{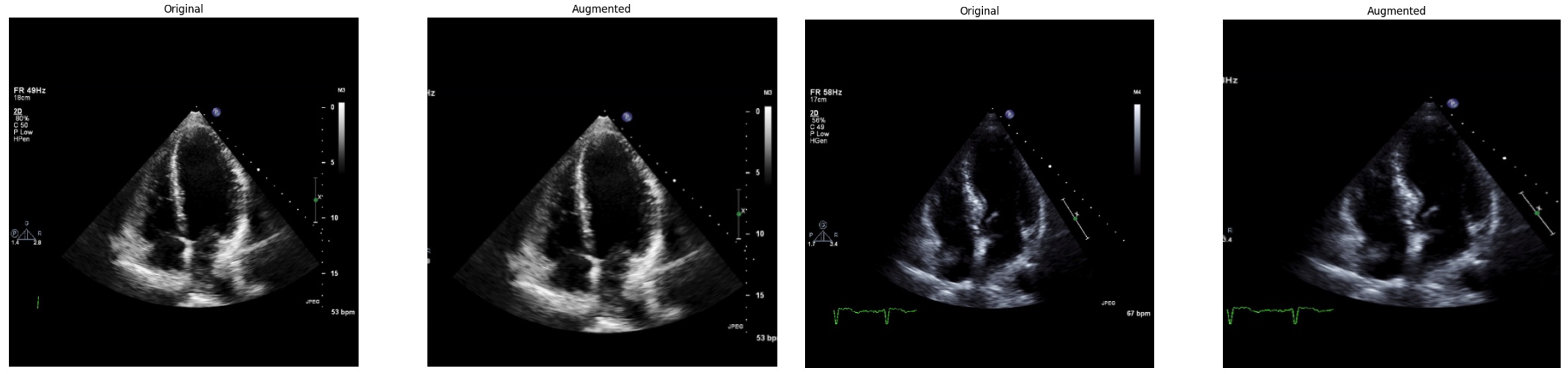}
\caption{
H: A.RandomResizedCrop(size=(512, 512), scale=(0.6, 1.0), ratio=(0.8, 1.2), p=0.7)
}
\label{fig:example}
\end{figure}

\begin{figure}[H]
\centering
\includegraphics[width=\linewidth]{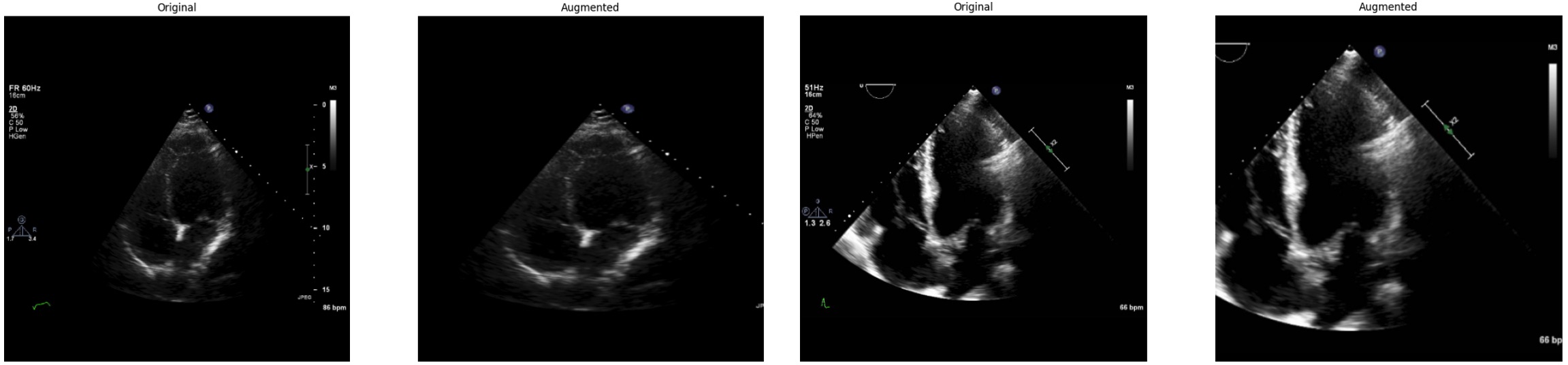}
\caption{
C1: A.RandomResizedCrop(size=(512, 512),scale=(0.50, 0.9),ratio=(0.7, 1.3),p=0.9)
}
\label{fig:example}
\end{figure}

\subsection{Sharpen}

\begin{figure}[H]
\centering
\includegraphics[width=\linewidth]{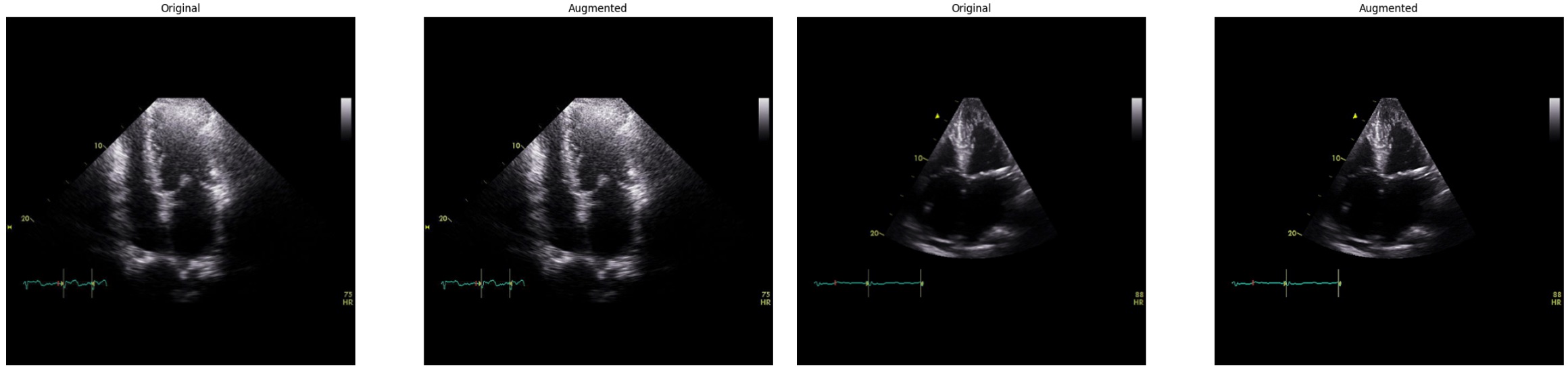}
\caption{
L: A.Sharpen(alpha=(0.05, 0.10), lightness=(1.0, 1.0), p=0.15)
}
\label{fig:example}
\end{figure}

\begin{figure}[H]
\centering
\includegraphics[width=\linewidth]{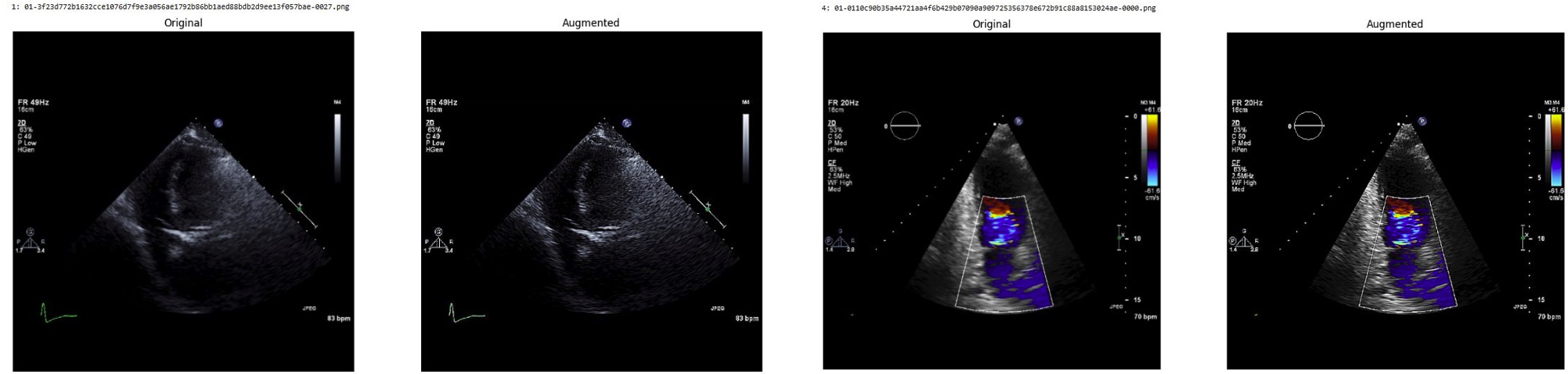}
\caption{
H: A.Sharpen(alpha=(0.15, 0.35), lightness=(0.8, 1.2), p=0.4)
}
\label{fig:example}
\end{figure}

\begin{figure}[H]
\centering
\includegraphics[width=\linewidth]{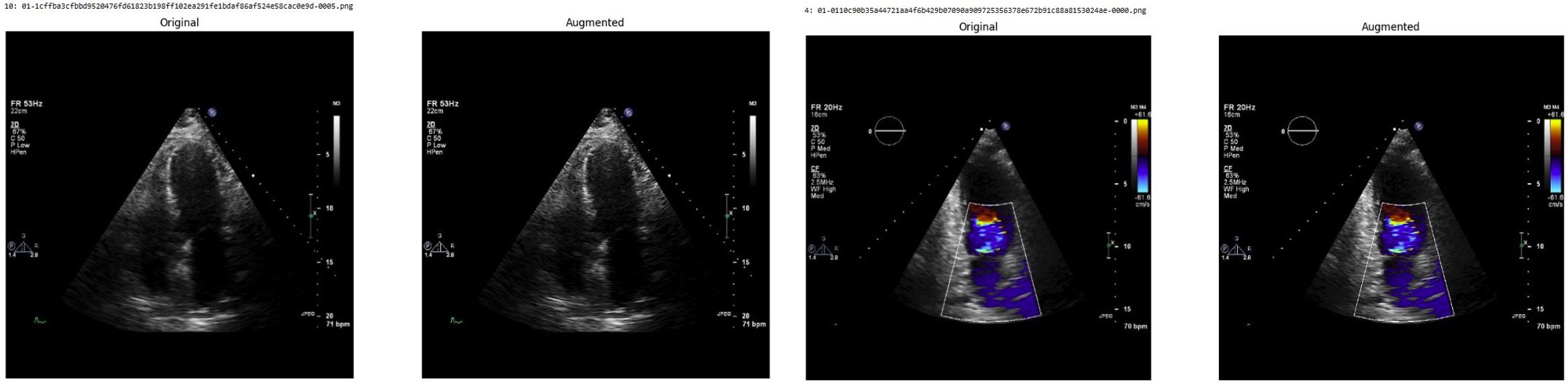}
\caption{
C1: A.Sharpen(alpha=(0.06, 0.12), lightness=(1.0, 1.0), p=0.25)
}
\label{fig:example}
\end{figure}

\subsection{ShiftScaleRotate}

\begin{figure}[H]
\centering
\includegraphics[width=\linewidth]{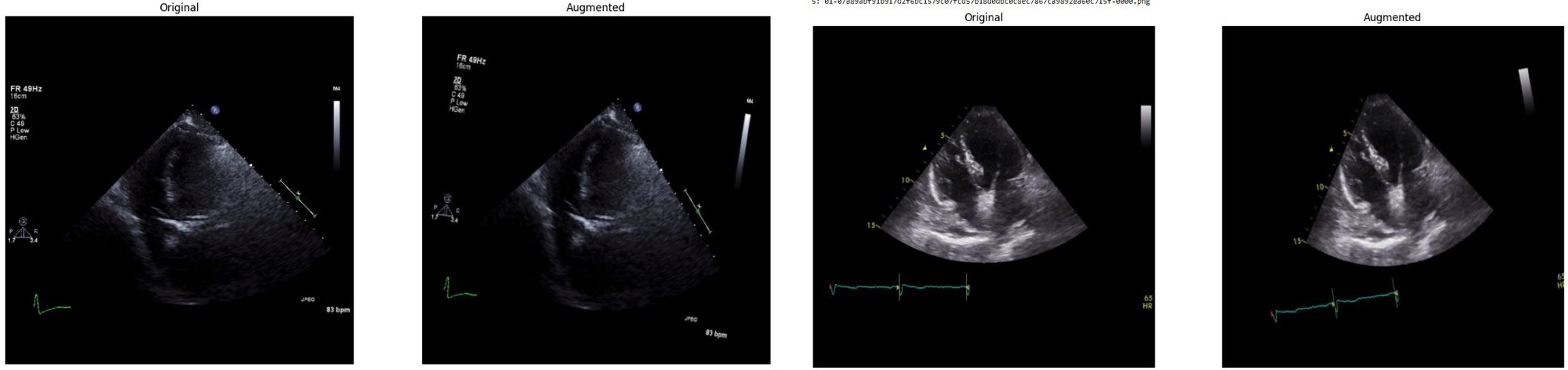}
\caption{\centering
L: A.ShiftScaleRotate( shift\_limit=0.05, scale\_limit=0.10, rotate\_limit=10, border\_mode=0,value=0, mask\_value=0, p=0.9)
}
\label{fig:example}
\end{figure}

\begin{figure}[H]
\centering
\includegraphics[width=\linewidth]{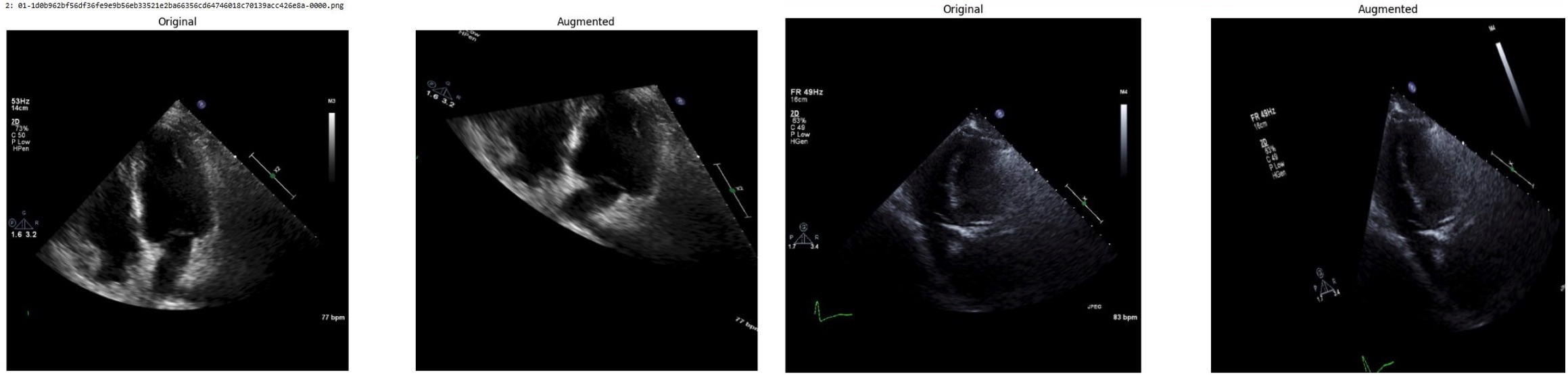}
\caption{\centering
H: A.ShiftScaleRotate(shift\_limit=0.15, scale\_limit=0.25, rotate\_limit=30, border\_mode=0,value=0, mask\_value=0, p=0.7)
}
\label{fig:example}
\end{figure}

\begin{figure}[H]
\centering
\includegraphics[width=\linewidth]{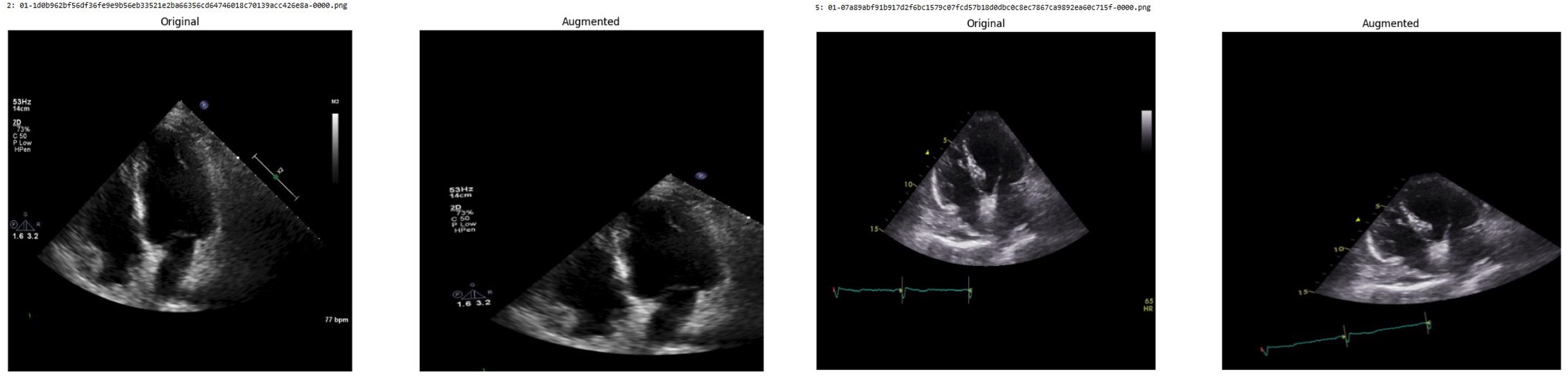}
\caption{\centering
C1: A.ShiftScaleRotate(shift\_limit=0.25, scale\_limit=0.35, rotate\_limit=35, border\_mode=0,value=0, mask\_value=0, p=0.6)
}
\label{fig:example}
\end{figure}

\subsection{CenterCrop}

\begin{figure}[H]
\centering
\includegraphics[width=\linewidth]{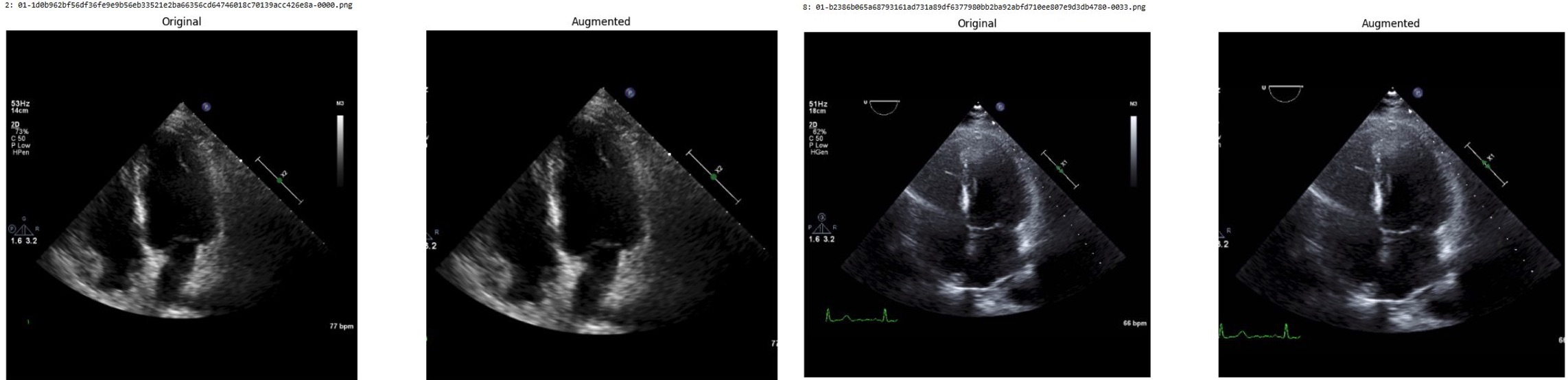}
\caption{
L: A.CenterCrop(448, 448, p=1.0)
}
\label{fig:example}
\end{figure}

\begin{figure}[H]
\centering
\includegraphics[width=\linewidth]{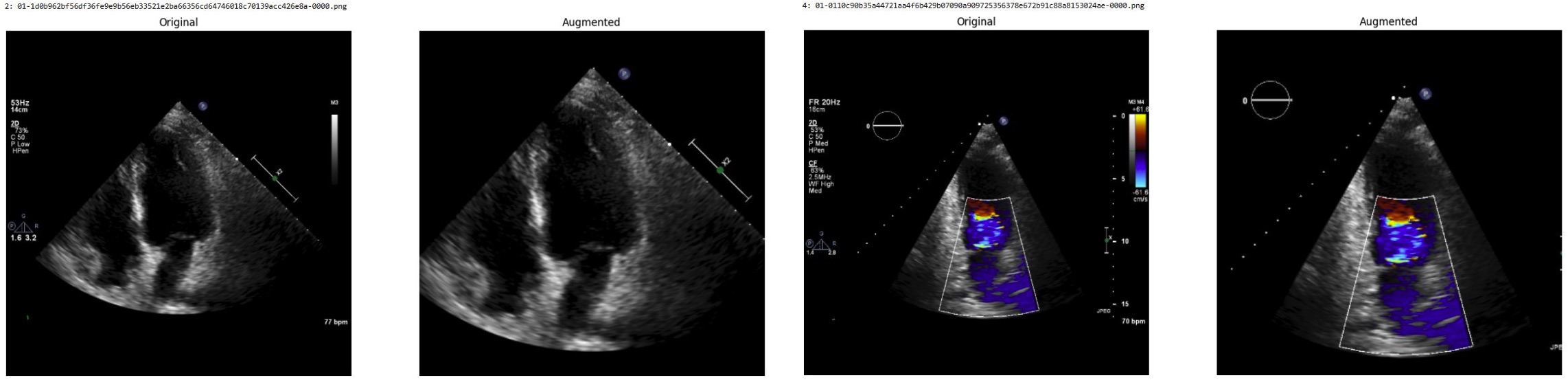}
\caption{
H: A.CenterCrop(384, 384, p=1.0)
}
\label{fig:example}
\end{figure}

\begin{figure}[H]
\centering
\includegraphics[width=\linewidth]{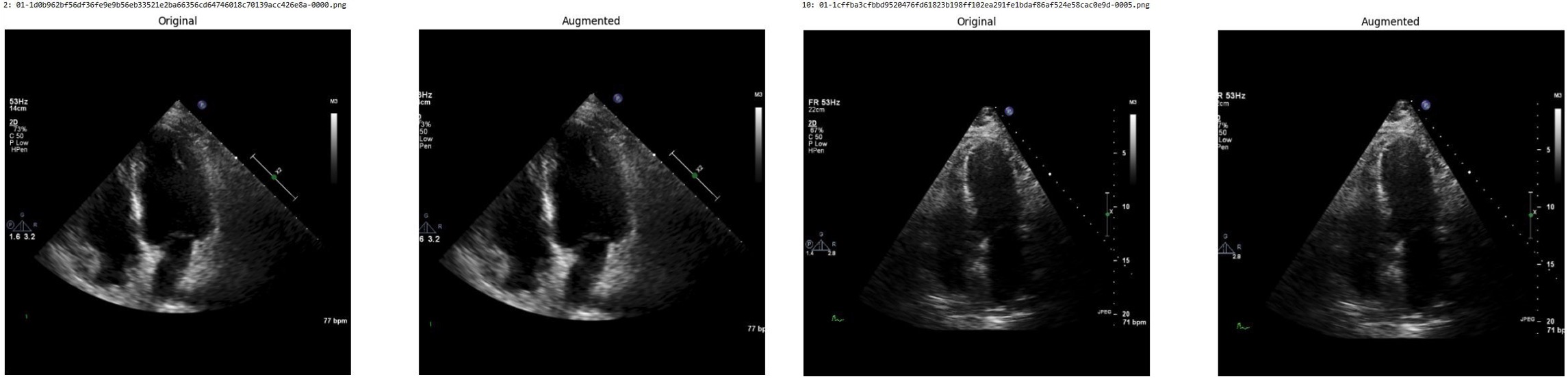}
\caption{
C1: A.CenterCrop(480, 480, p=1.0)
}
\label{fig:example}
\end{figure}

\subsection{CoarseDropout}

\begin{figure}[H]
\centering
\includegraphics[width=\linewidth]{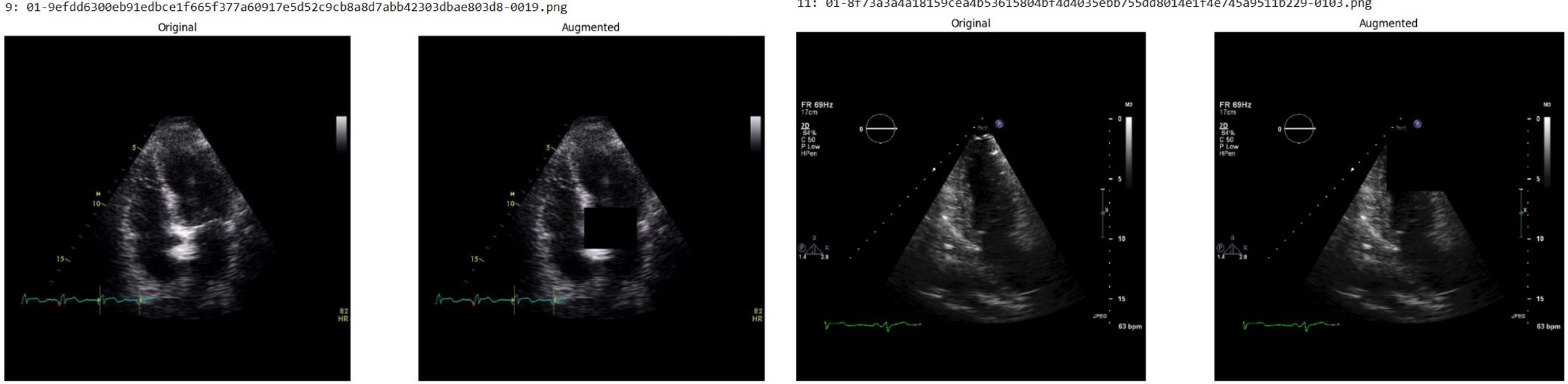}
\caption{\centering
L: A.CoarseDropout( max\_holes=2, max\_height=int(0.08*H),max\_width=int(0.08*W), min\_holes=1, min\_height=int(0.03*H),min\_width=int(0.03*W), fill\_value=0, p=0.15 )
}
\label{fig:example}
\end{figure}

\begin{figure}[H]
\centering
\includegraphics[width=\linewidth]{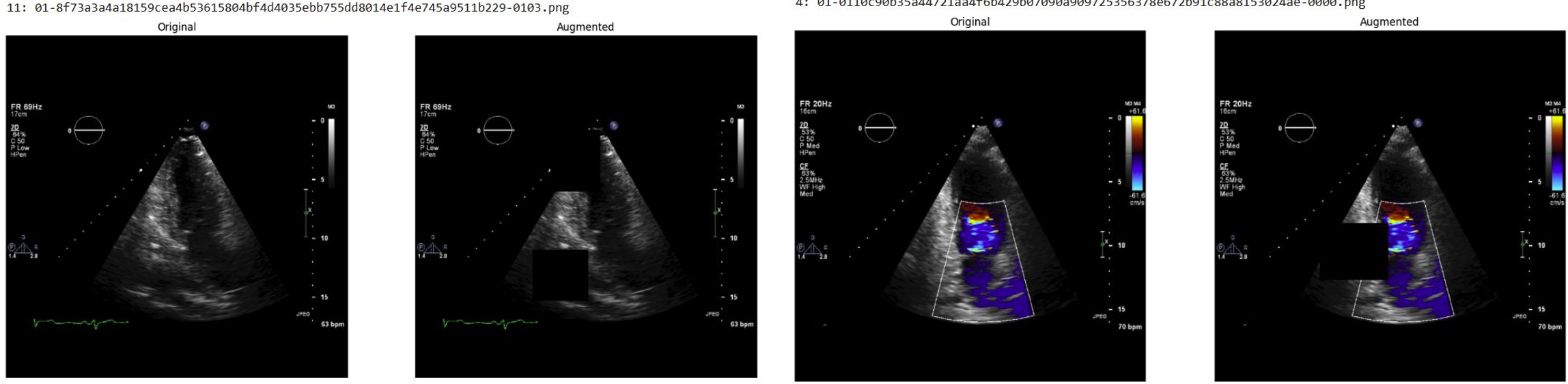}
\caption{
C1: A.CLAHE(clip\_limit=1.8, tile\_grid\_size=(8, 8), p=0.25)
}
\label{fig:example}
\end{figure}

\begin{figure}[H]
\centering
\includegraphics[width=\linewidth]{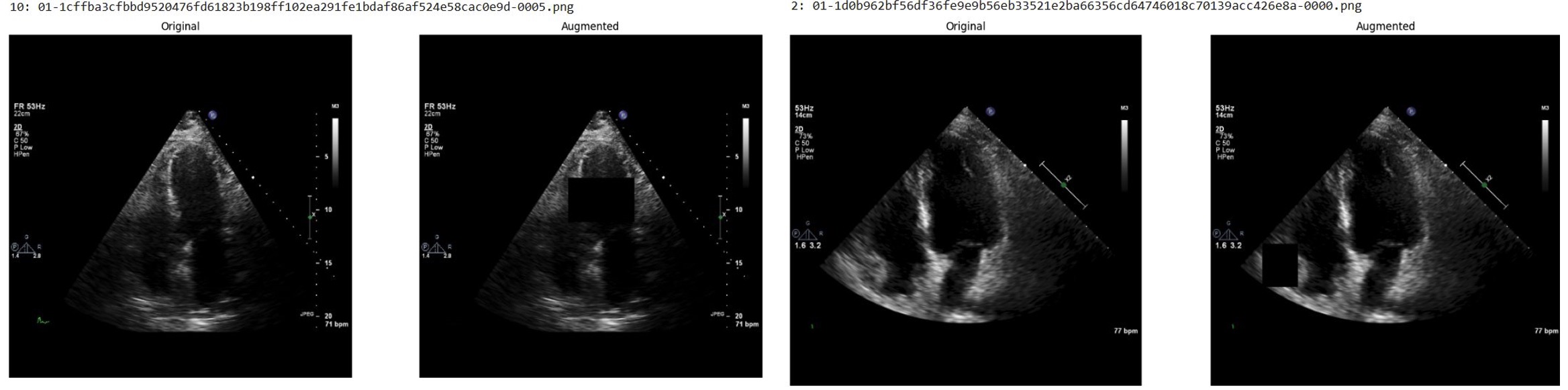}
\caption{\centering
H: A.CoarseDropout(max\_holes=4, max\_height=int(0.15 * H), max\_width=int(0.15 * W), min\_holes=2, min\_height=int(0.05 * H), min\_width=int(0.05 * W), fill\_value=0, p=0.25)
}
\label{fig:example}
\end{figure}

\begin{figure}[H]
\centering
\includegraphics[width=\linewidth]{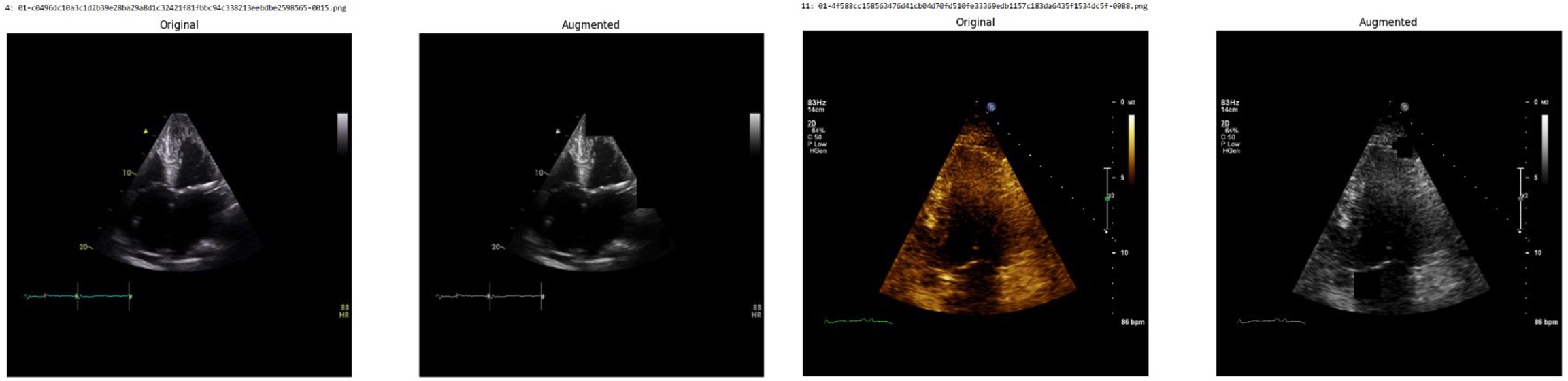}
\caption{\centering
C1: A.CoarseDropout(max\_holes=1, min\_holes=1, min\_height=int(0.02 * H), min\_width=int(0.02 * W), max\_height=int(0.04 * H), max\_width=int(0.04 * W), fill\_value=0.2, p=0.05)
}
\label{fig:example}
\end{figure}

\subsection{CropNonEmptyMaskIfExists}

\begin{figure}[H]
\centering
\includegraphics[width=\linewidth]{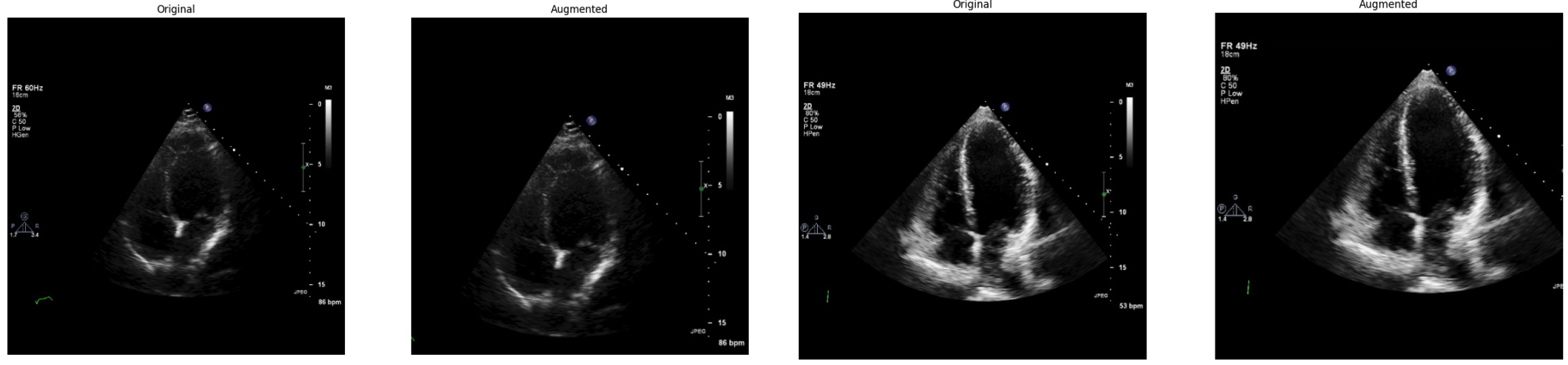}
\caption{
L: A.CropNonEmptyMaskIfExists(height=448, width=448,p=1.0)
}
\label{fig:example}
\end{figure}

\begin{figure}[H]
\centering
\includegraphics[width=\linewidth]{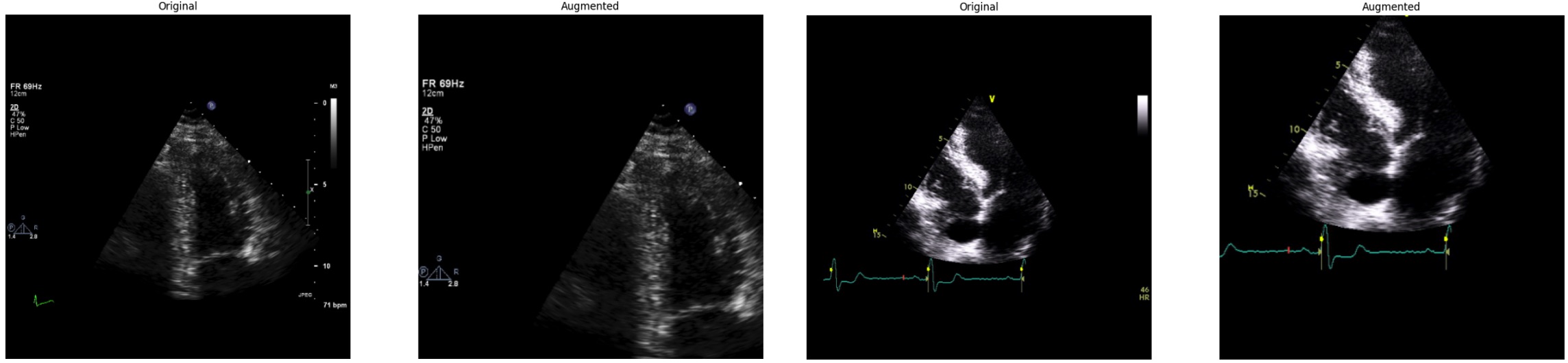}
\caption{
H1: A.CropNonEmptyMaskIfExists(height=384, width=384, p=1.0)
}
\label{fig:example}
\end{figure}

\begin{figure}[H]
\centering
\includegraphics[width=\linewidth]{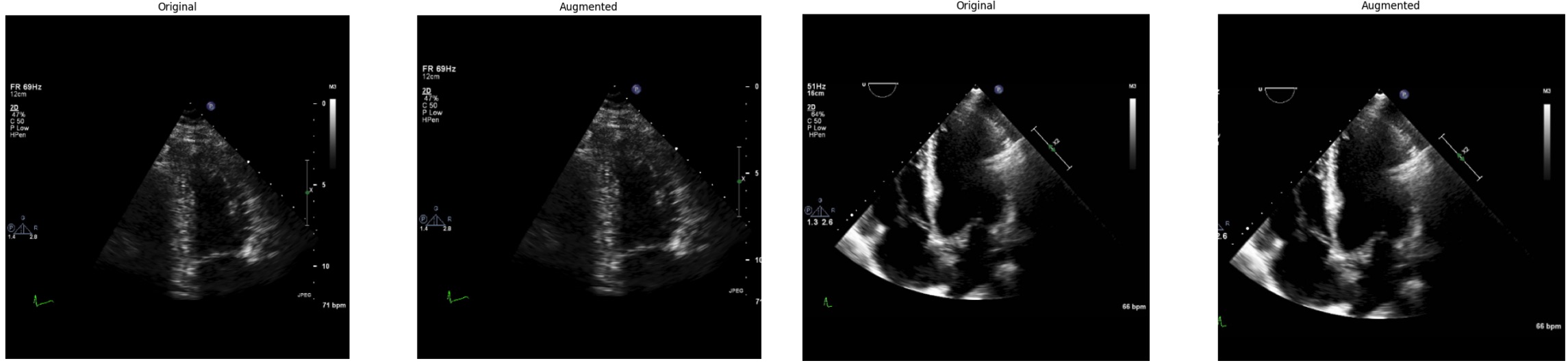}
\caption{
C1: A.CropNonEmptyMaskIfExists(height=480, width=480,p=0.7)
}
\label{fig:example}
\end{figure}

\subsection{Downscale}

\begin{figure}[H]
\centering
\includegraphics[width=\linewidth]{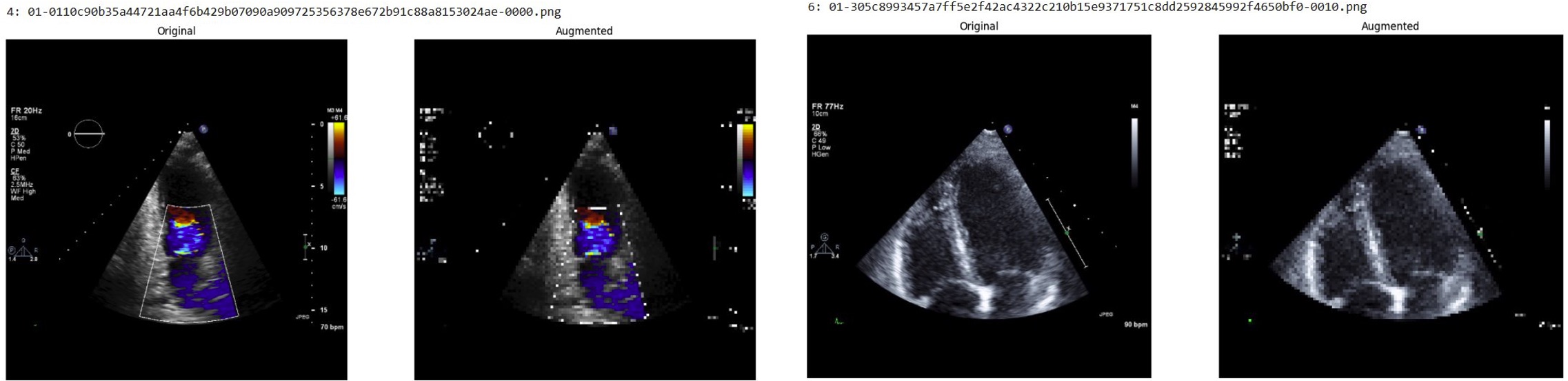}
\caption{
L: A.Downscale(scale\_min=0.65, scale\_max=0.85, interpolation=1, p=0.25)
}
\label{fig:example}
\end{figure}

\begin{figure}[H]
\centering
\includegraphics[width=\linewidth]{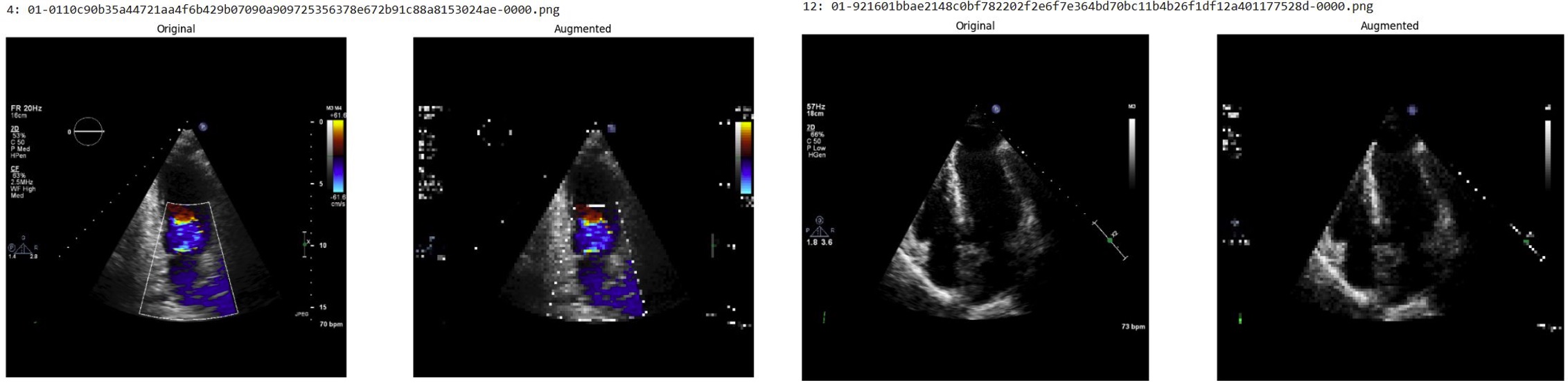}
\caption{
H: A.Downscale(scale\_min=0.4, scale\_max=0.8, interpolation=1, p=0.35)
}
\label{fig:example}
\end{figure}

\begin{figure}[H]
\centering
\includegraphics[width=\linewidth]{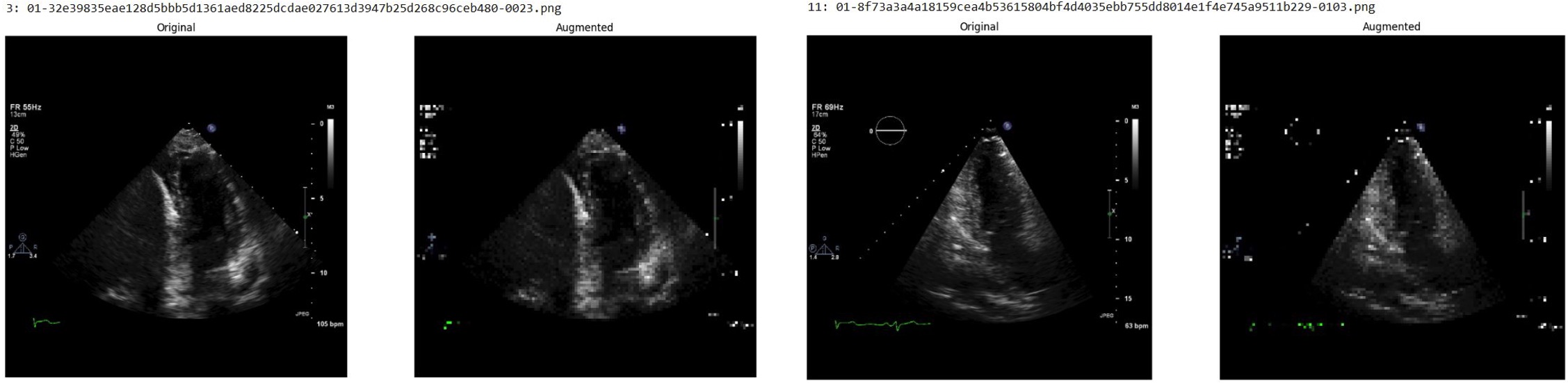}
\caption{
C1: A.Downscale(scale\_min=0.5, scale\_max=0.80, interpolation=1, p=0.45)
}
\label{fig:example}
\end{figure}

\begin{figure}[H]
\centering
\includegraphics[width=\linewidth]{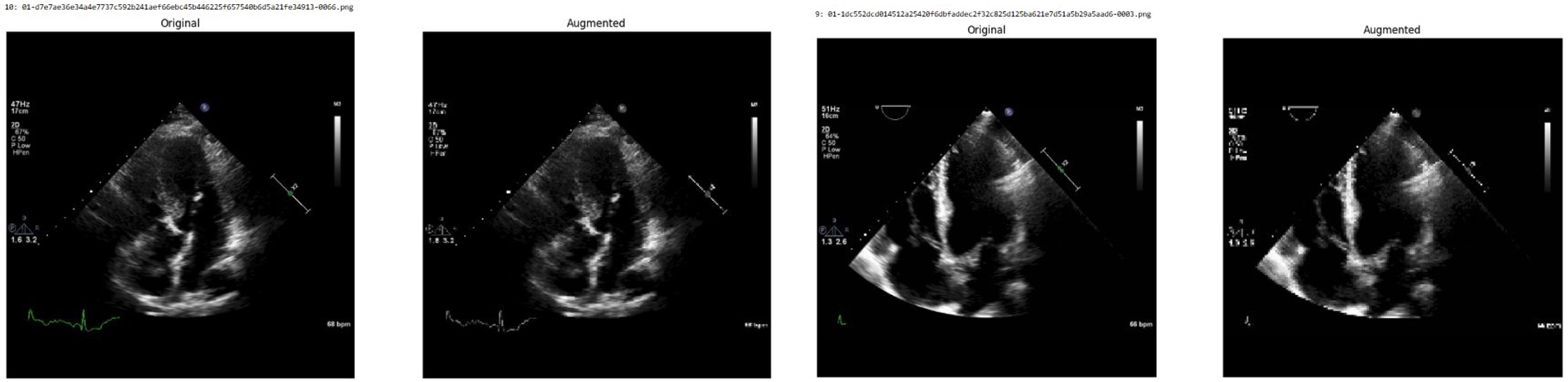}
\caption{
C2: A.Downscale(scale\_range=[0.35, 0.6], interpolation\_pair={"upscale":0,"downscale":0})
}
\label{fig:example}
\end{figure}

\subsection{ImageCompression}

\begin{figure}[H]
\centering
\includegraphics[width=\linewidth]{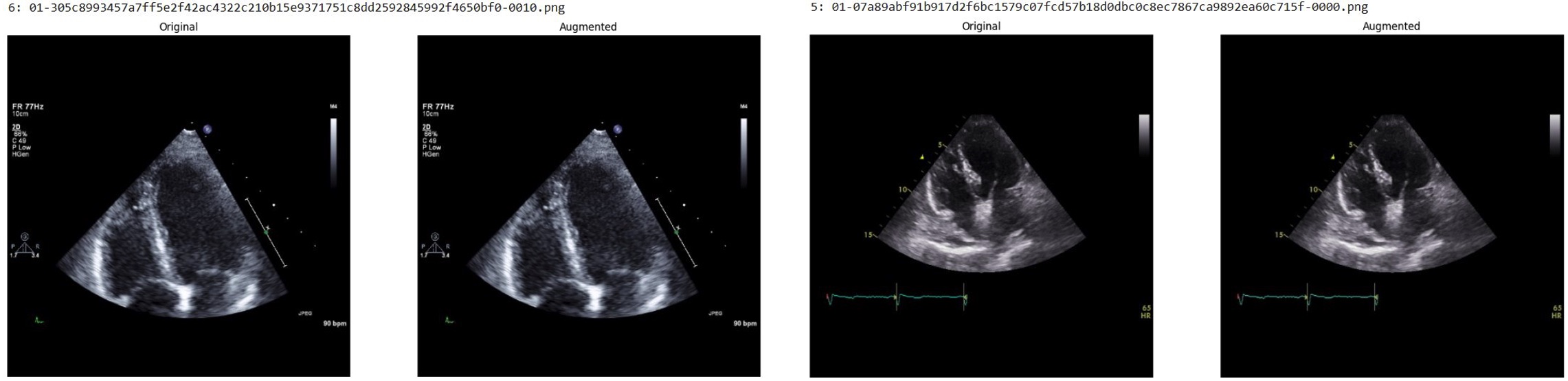}
\caption{
L: A.ImageCompression(quality\_lower=40, quality\_upper=70,p=0.30)
}
\label{fig:example}
\end{figure}

\begin{figure}[H]
\centering
\includegraphics[width=\linewidth]{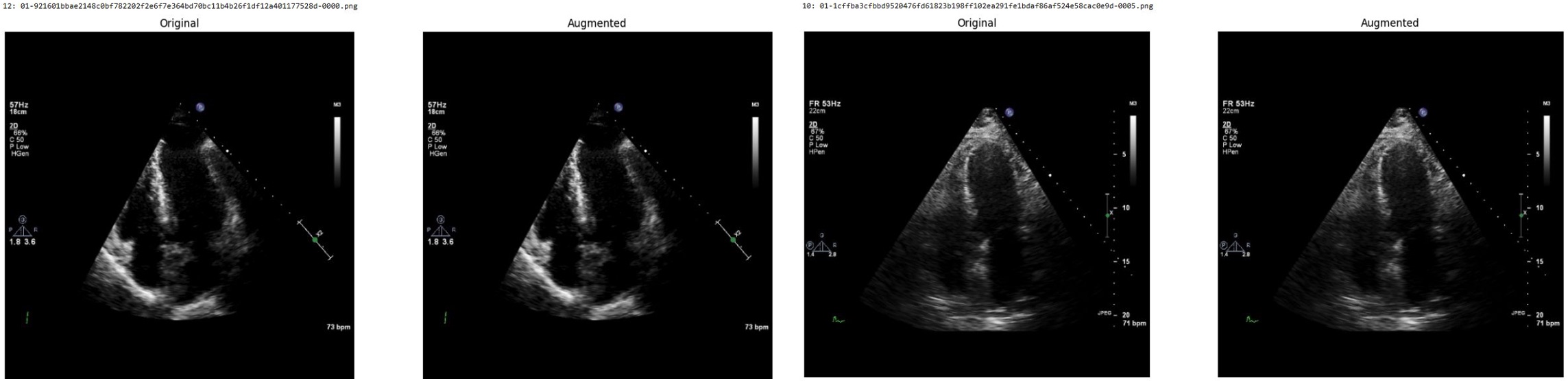}
\caption{
H: A.ImageCompression(quality\_lower=10, quality\_upper=50, p=0.40)
}
\label{fig:example}
\end{figure}

\begin{figure}[H]
\centering
\includegraphics[width=\linewidth]{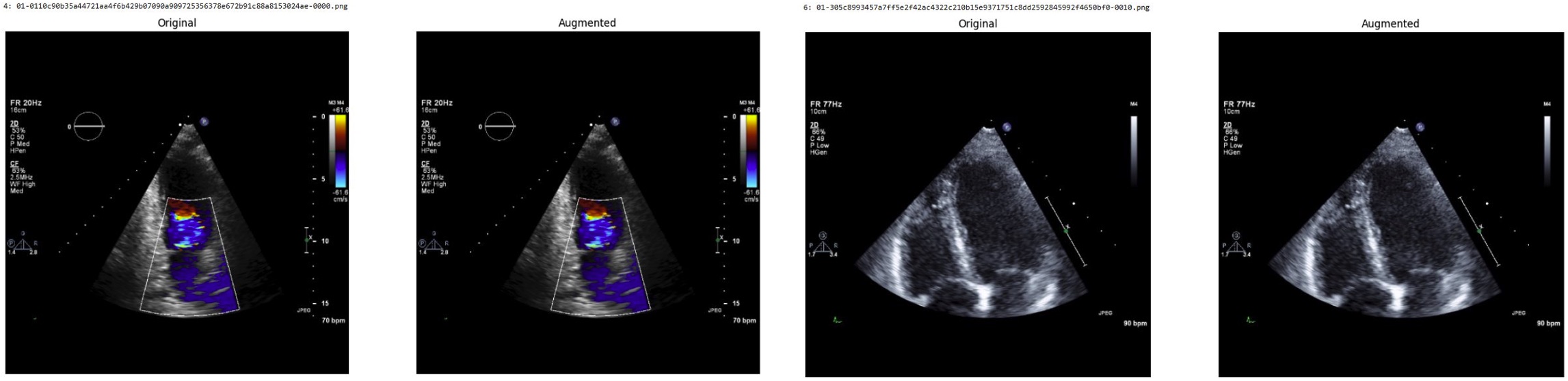}
\caption{
C1: A.ImageCompression(quality\_lower=30, quality\_upper=80, p=0.35)
}
\label{fig:example}
\end{figure}

\subsection{IntensityWindowing}

\begin{figure}[H]
\centering
\includegraphics[width=\linewidth]{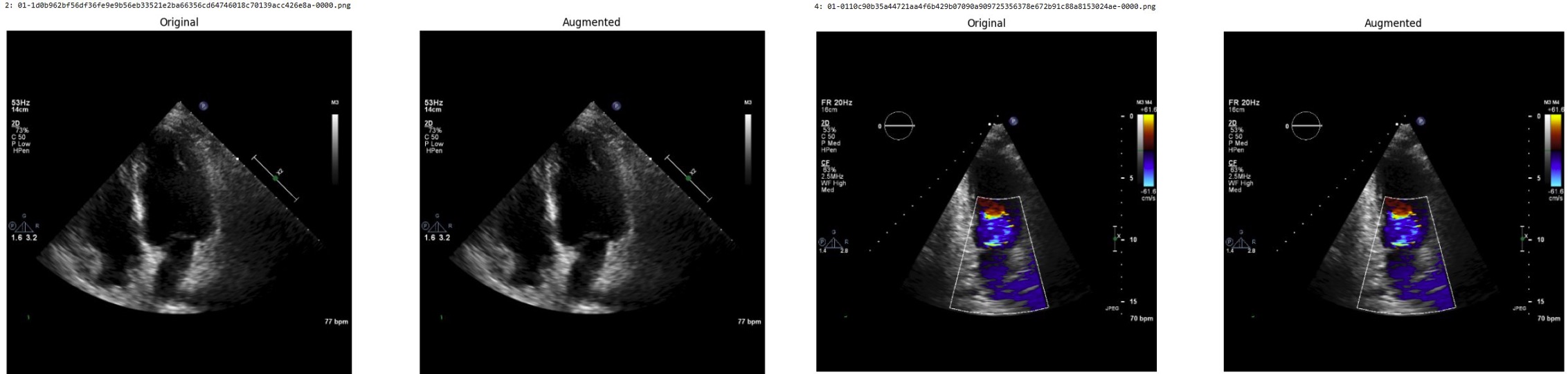}
\caption{
L: IntensityWindowing( window\_center=(0.35, 0.65),  window\_width=(0.25,0.55),  p=0.5 )
}
\label{fig:example}
\end{figure}

\begin{figure}[H]
\centering
\includegraphics[width=\linewidth]{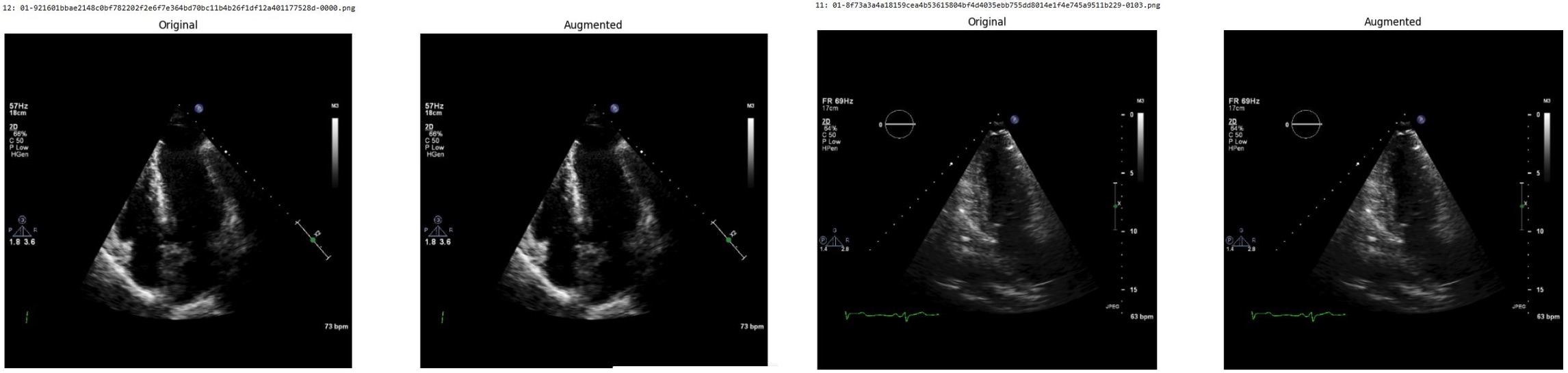}
\caption{
H: IntensityWindowing(window\_center=(-0.20, 1.20),window\_width=(0.08, 1.40),p=0.5)
}
\label{fig:example}
\end{figure}

\begin{figure}[H]
\centering
\includegraphics[width=\linewidth]{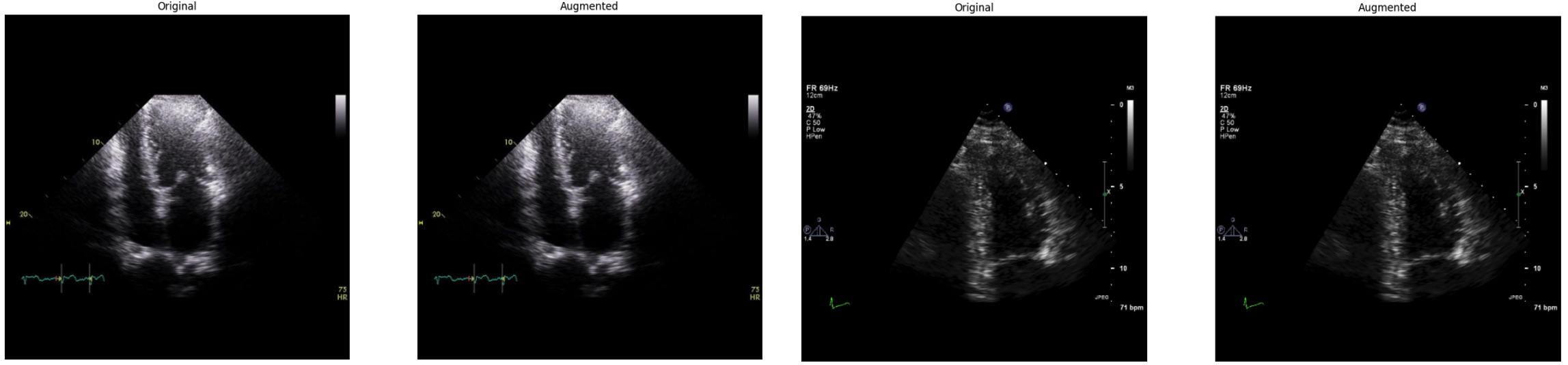}
\caption{
C1: IntensityWindowing(window\_center=(0.48,0.55), window\_width=(0.38, 0.48),p=0.15)
}
\label{fig:example}
\end{figure}

\subsection{UnsharpMask}

\begin{figure}[H]
\centering
\includegraphics[width=\linewidth]{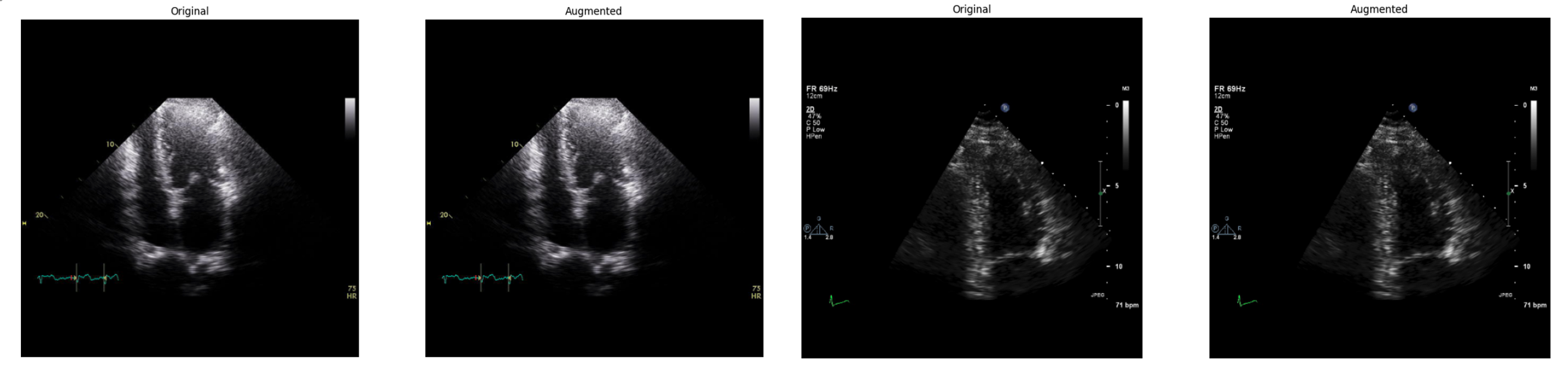}
\caption{
L: A.UnsharpMask(blur\_limit=(3, 5), alpha=(0.10, 0.25),p=0.20)
}
\label{fig:example}
\end{figure}

\begin{figure}[H]
\centering
\includegraphics[width=\linewidth]{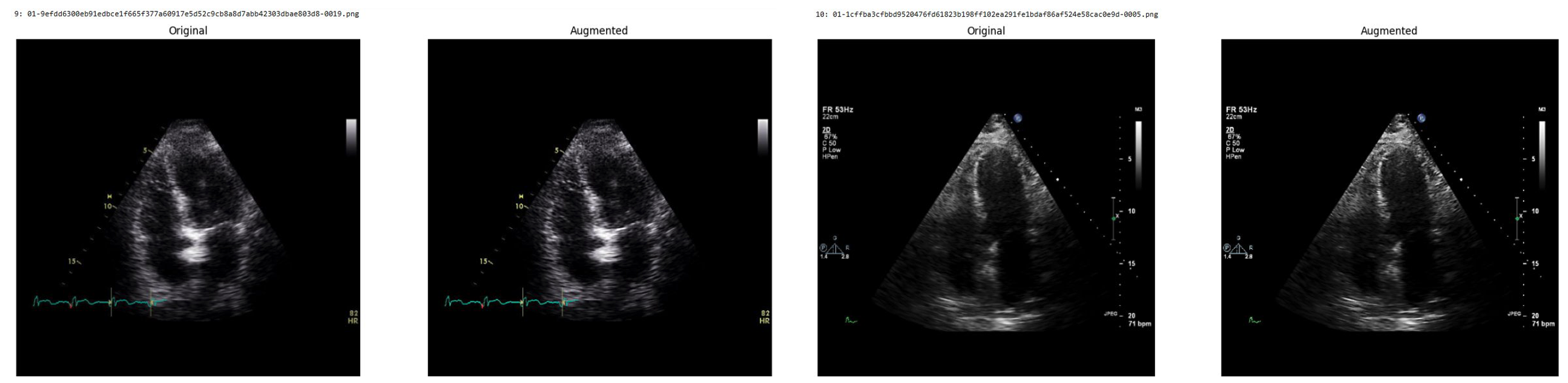}
\caption{
H: A.UnsharpMask(blur\_limit=(5, 13), alpha=(0.30, 0.70), p=0.60)
}
\label{fig:example}
\end{figure}

\begin{figure}[H]
\centering
\includegraphics[width=\linewidth]{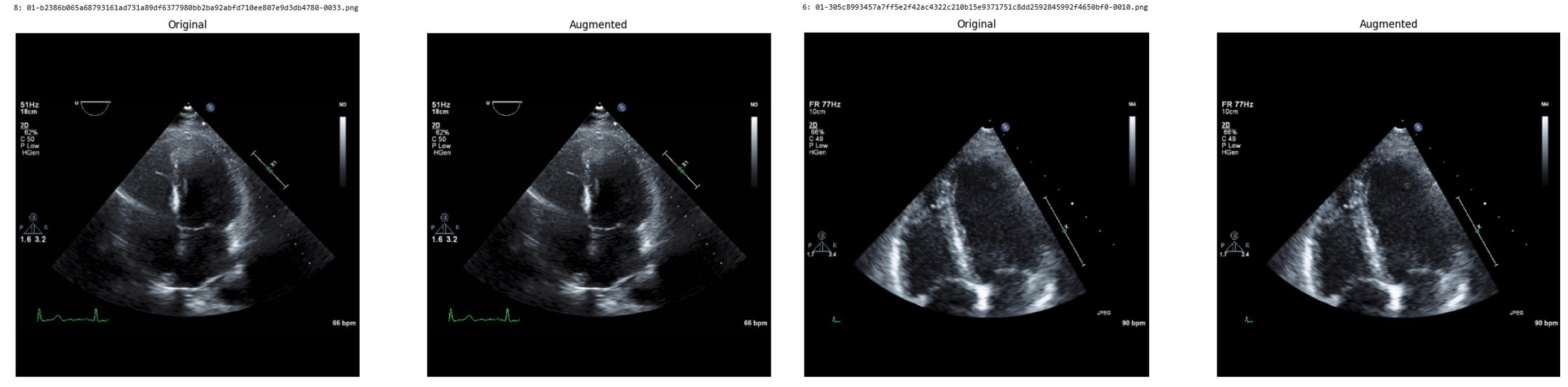}
\caption{
C1: A.UnsharpMask(blur\_limit=(3,7), alpha=(0.15,0.40), p=0.30)
}
\label{fig:example}
\end{figure}

\begin{figure}[H]
\centering
\includegraphics[width=\linewidth]{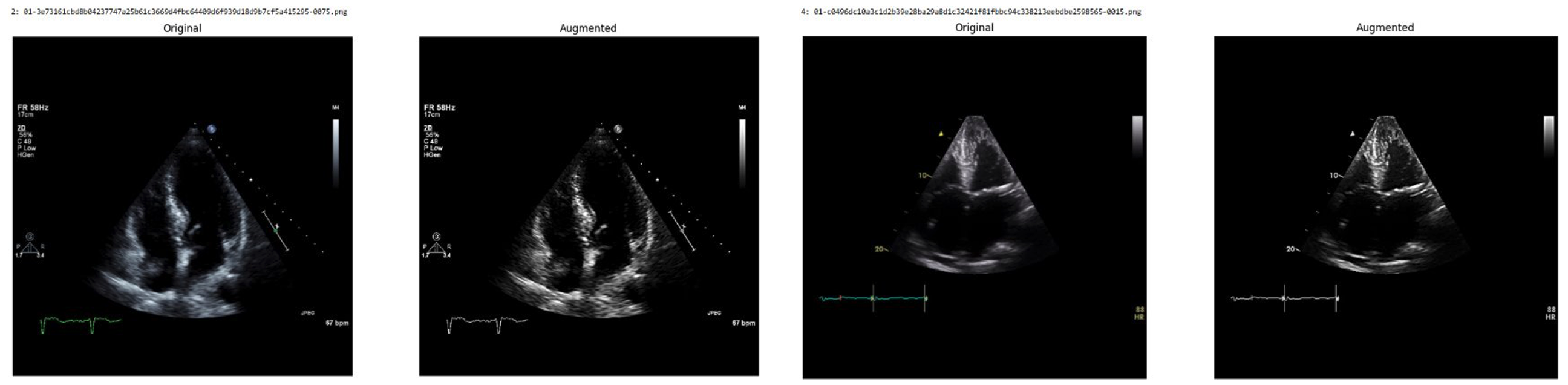}
\caption{
C2: A.UnsharpMask(blur\_limit=(3,7), alpha=( 0.15,0.40), p=0.30)
}
\label{fig:example}
\end{figure}

\begin{figure}[H]
\centering
\includegraphics[width=\linewidth]{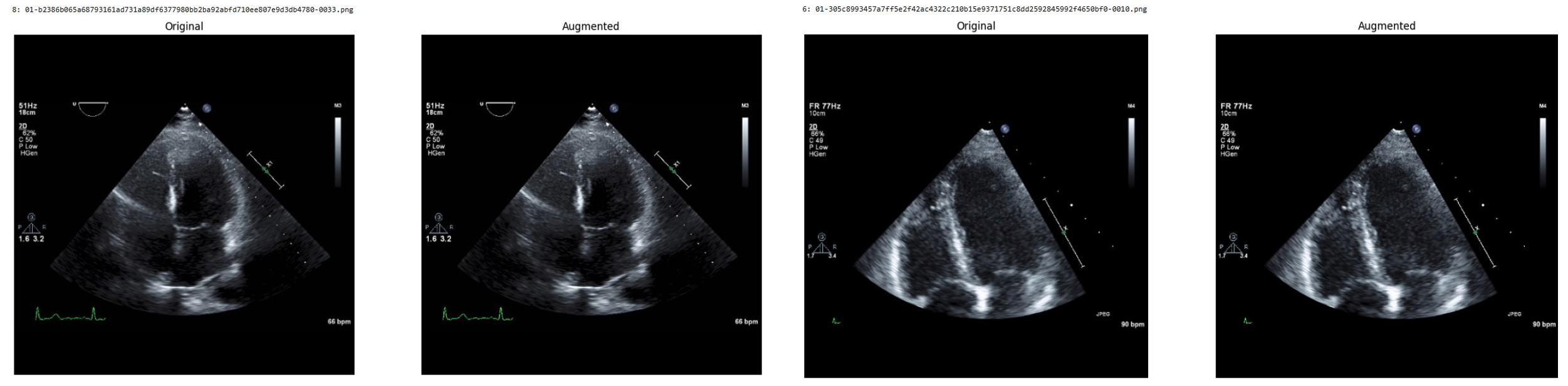}
\caption{
CH: A.UnsharpMask(blur\_limit=(10,20), sigma\_limit=10, alpha=(1,1), threshold=1, p=0.30)
}
\label{fig:example}
\end{figure}

\subsection{GridDistortion}

\begin{figure}[H]
\centering
\includegraphics[width=\linewidth]{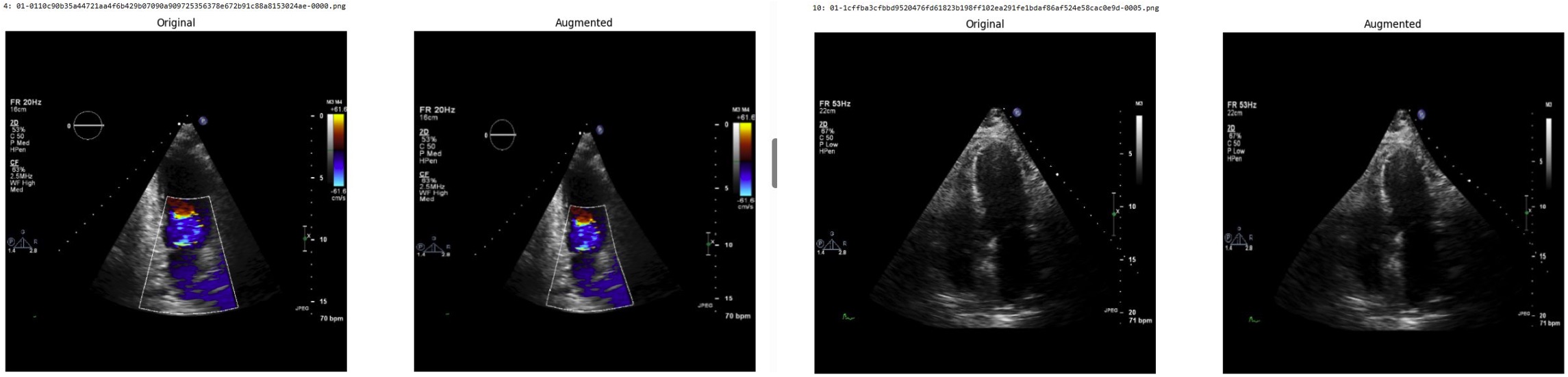}
\caption{
L: A.GridDistortion(num\_steps=5, distort\_limit=0.2, p=0.4)
}
\label{fig:example}
\end{figure}

\begin{figure}[H]
\centering
\includegraphics[width=\linewidth]{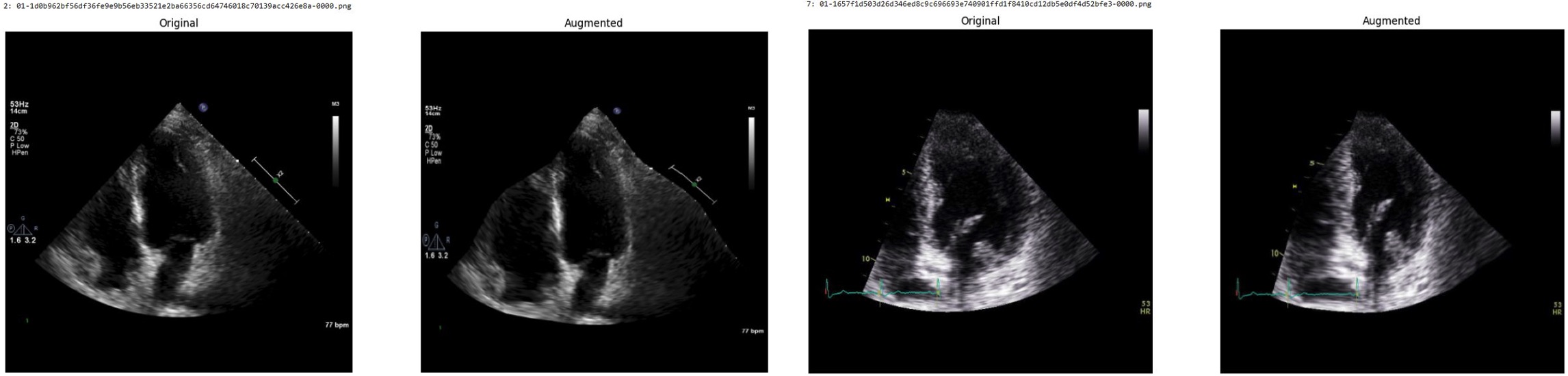}
\caption{
H: A.GridDistortion(num\_steps=8, distort\_limit=0.35, p=0.6)
}
\label{fig:example}
\end{figure}

\begin{figure}[H]
\centering
\includegraphics[width=\linewidth]{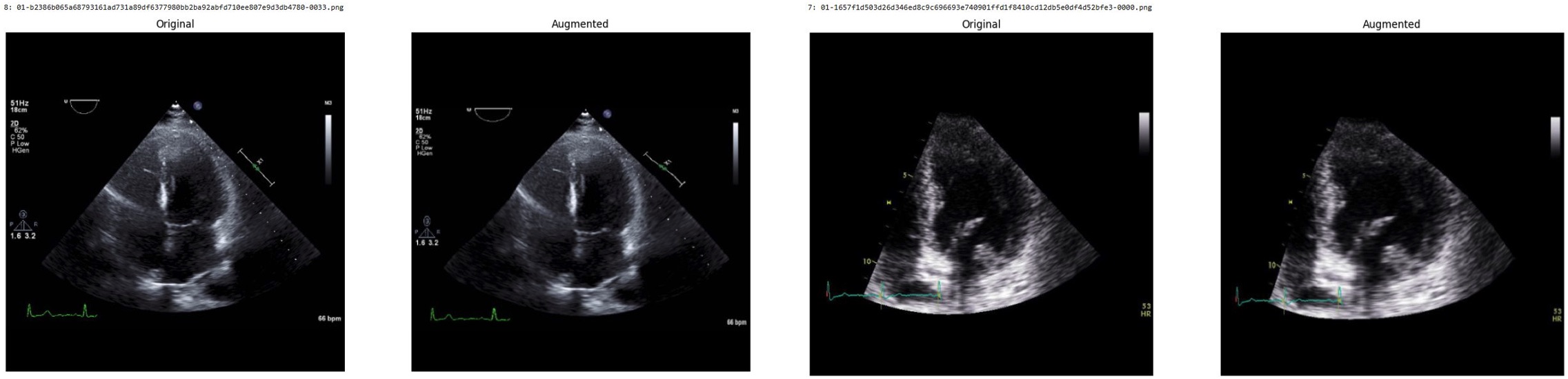}
\caption{
C1: A.GridDistortion(num\_steps=7, distort\_limit=0.25,p=0.35) 
}
\label{fig:example}
\end{figure}

\begin{figure}[H]
\centering
\includegraphics[width=\linewidth]{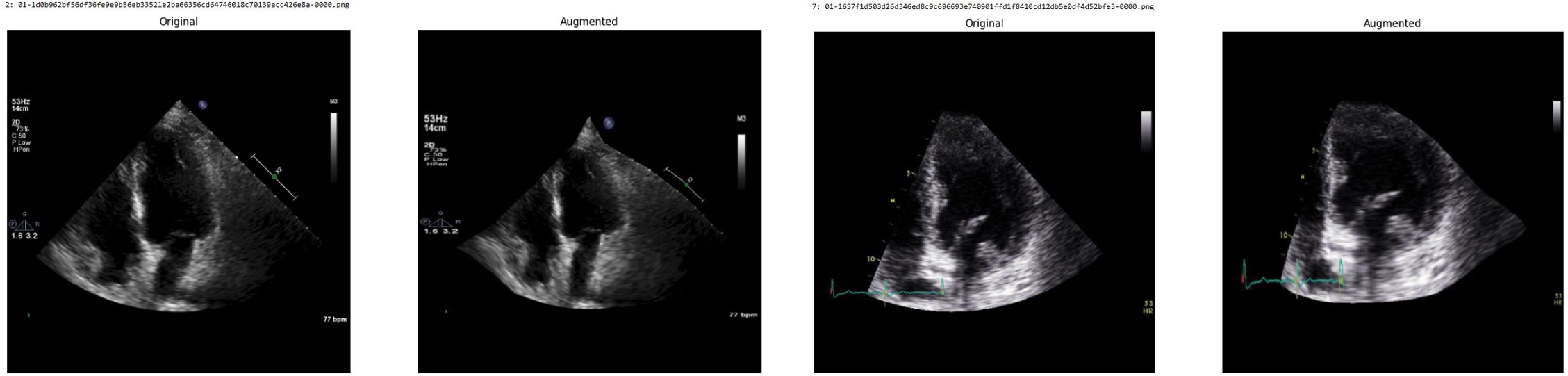}
\caption{
C2: GridDistortion(num\_steps=9, distort\_limit=0.38, p=0.55)
}
\label{fig:example}
\end{figure}

\begin{figure}[H]
\centering
\includegraphics[width=\linewidth]{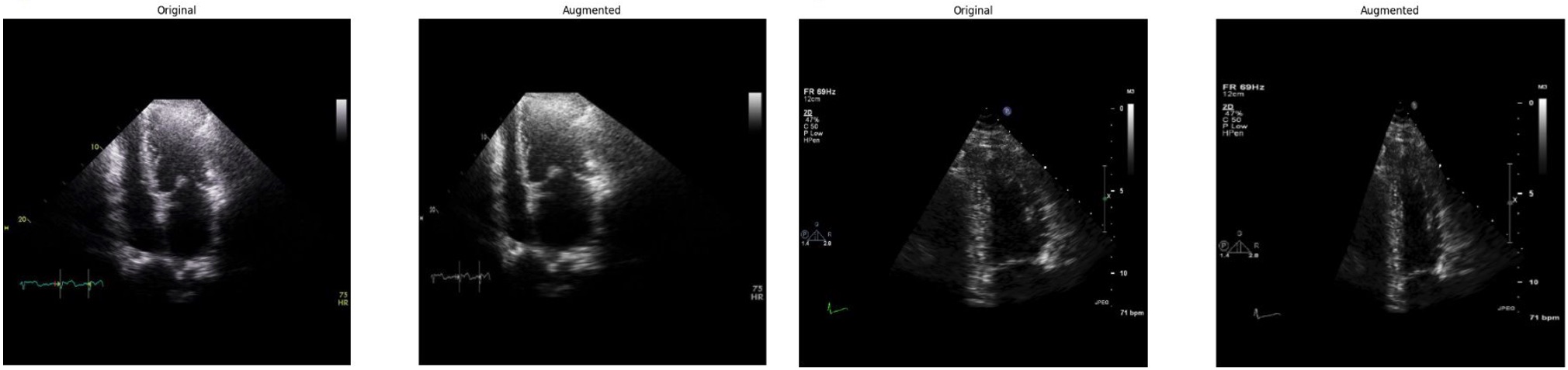}
\caption{\centering
C3: GridDistortion(num\_steps=3, distort\_limit=[-0.4, 0.4], interpolation=cv2.INTER\_AREA, normalized=True, mask\_interpolation=cv2.INTER\_AREA, keypoint\_remapping\_method="mask", border\_mode=cv2.BORDER\_REPLICATE, fill=0,  fill\_mask=0)
}
\label{fig:example}
\end{figure}

\subsection{RandomErasing}

\begin{figure}[H]
\centering
\includegraphics[width=\linewidth]{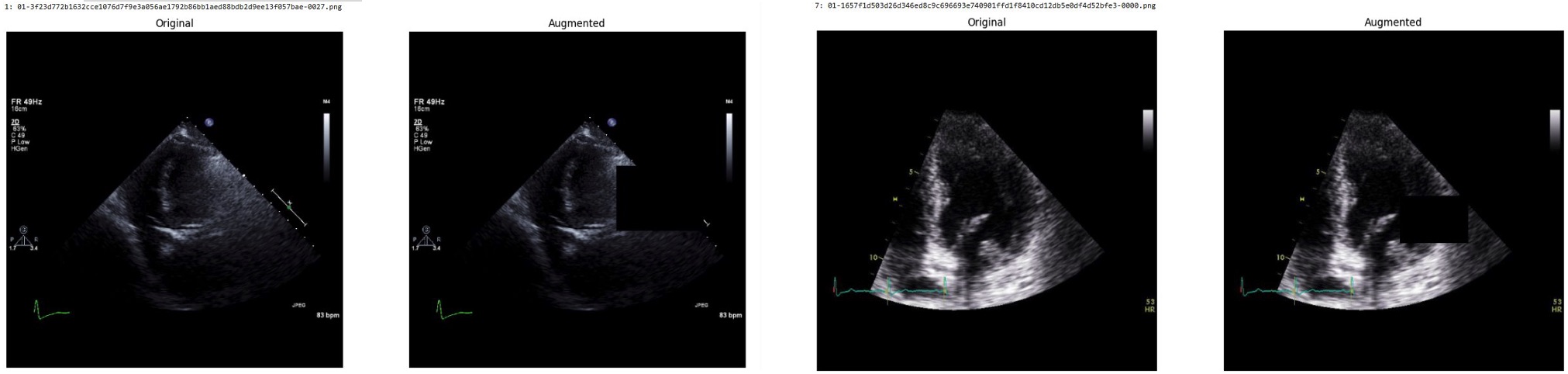}
\caption{
L: T.RandomErasing( p=1.0, scale=(0.02, 0.06), ratio=(0.6, 1.6), value=0.0,             inplace=False, p=0.25)
}
\label{fig:example}
\end{figure}

\begin{figure}[H]
\centering
\includegraphics[width=\linewidth]{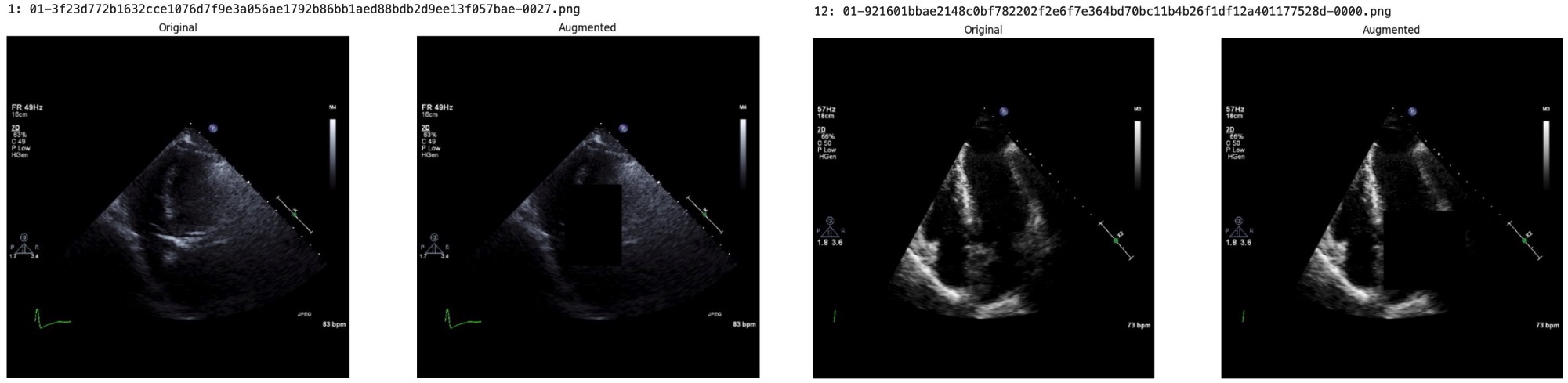}
\caption{
H: T.RandomErasing( p=1.0, scale=(0.02, 0.06), ratio=(0.6, 1.6), value=0.0, inplace=False, p=0.45)             
}
\label{fig:example}
\end{figure}

\begin{figure}[H]
\centering
\includegraphics[width=\linewidth]{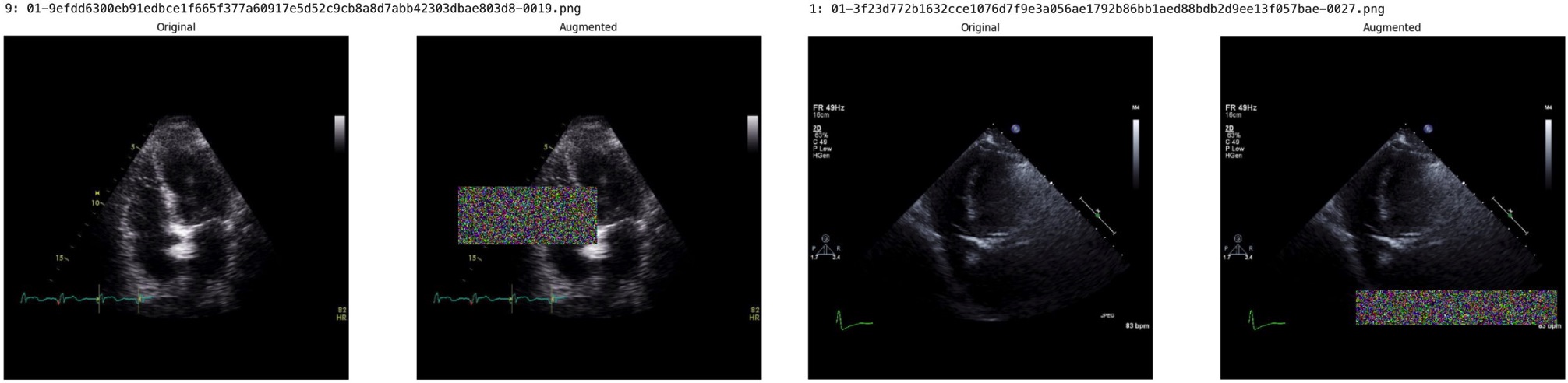}
\caption{
C1: T.RandomErasing( p=0.30, scale=(0.01, 0.07), ratio=(0.15, 4.5), value="random",inplace=False, p=0.45)           
}
\label{fig:example}
\end{figure}

\begin{figure}[H]
\centering
\includegraphics[width=\linewidth]{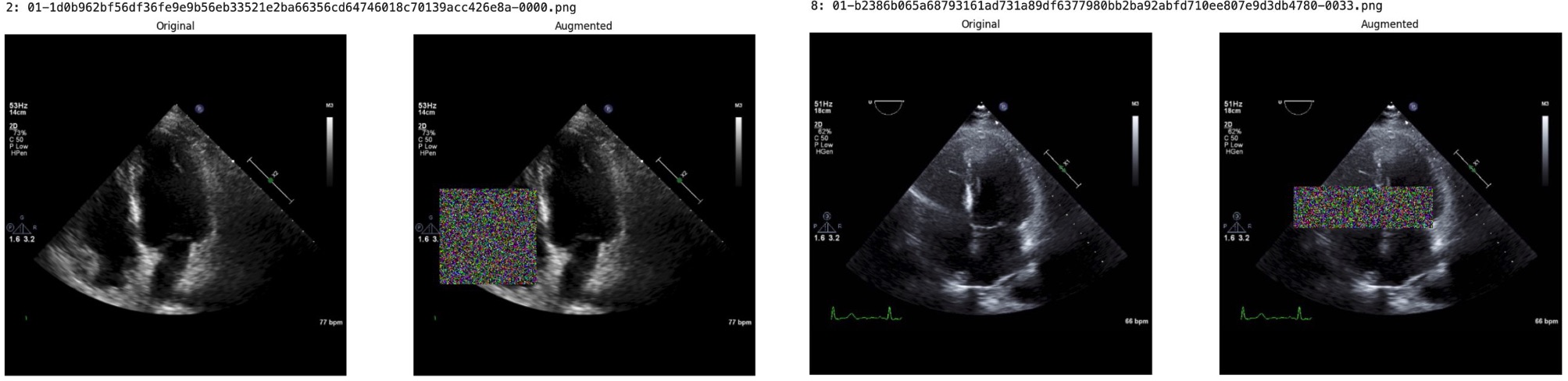}
\caption{
C2: T.RandomErasing( p=0.35, scale=(0.01, 0.08), ratio=(0.2, 3.0), value="random",inplace=False, p=0.25)            
}
\label{fig:example}
\end{figure}

\subsection{salt-and-pepper}

\begin{figure}[H]
\centering
\includegraphics[width=\linewidth]{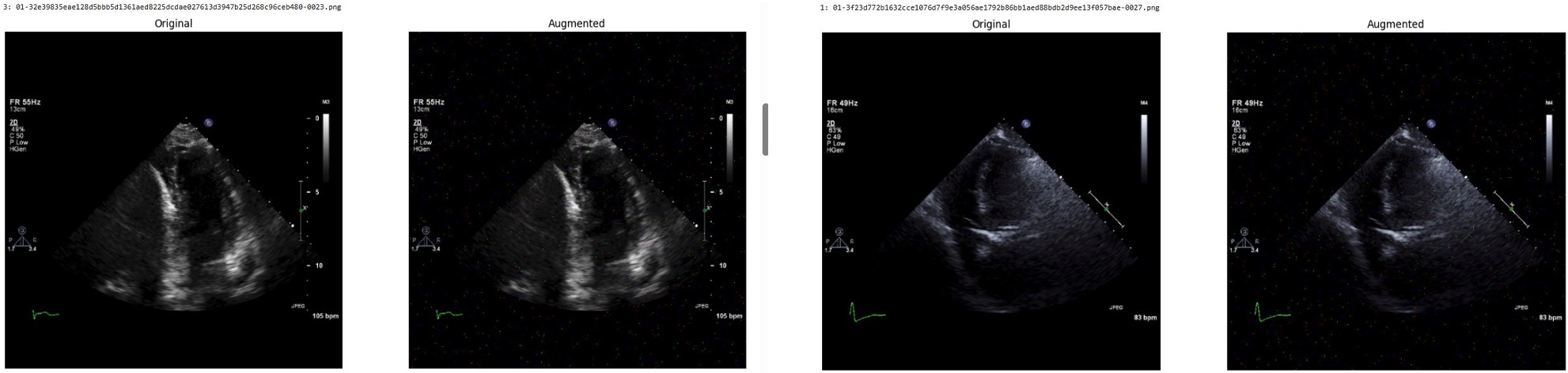}
\caption{
L: \_sp\_noise\_uint8(img, amount=0.003,ratio=0.5, p=0.25)
}
\label{fig:example}
\end{figure}

\begin{figure}[H]
\centering
\includegraphics[width=\linewidth]{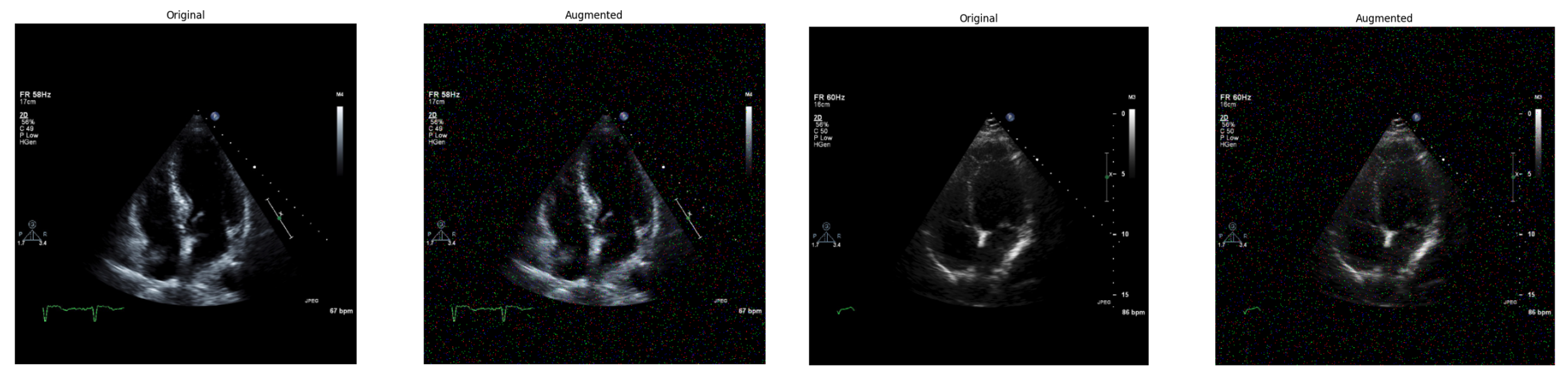}
\caption{\centering
H: \_sp\_noise\_uint8(img, amount=float(np.random.uniform(0.008, 0.030)), ratio=float(np.random.uniform(0.35, 0.65)), p=0.5)          
}
\label{fig:example}
\end{figure}

\begin{figure}[H]
\centering
\includegraphics[width=\linewidth]{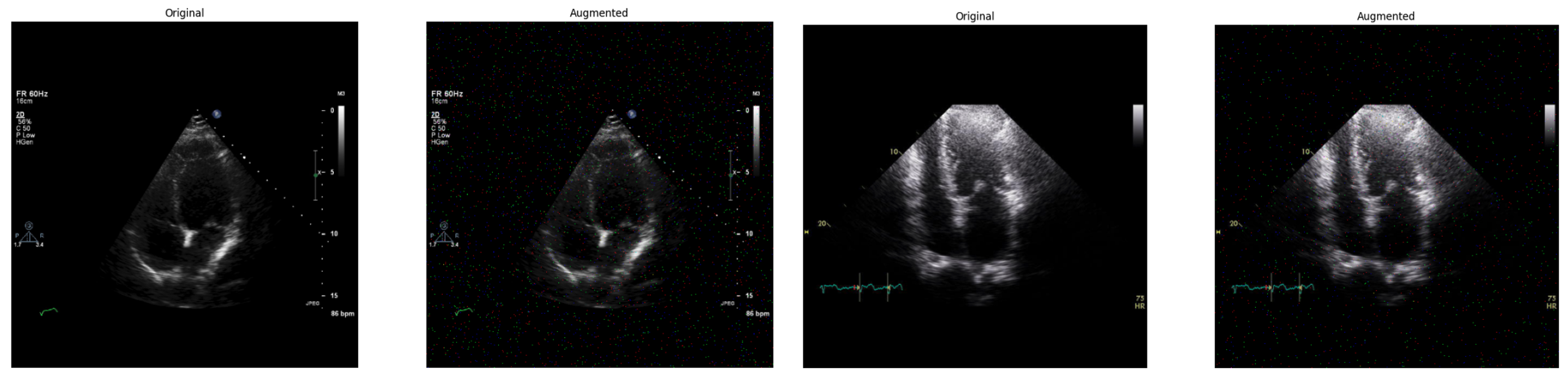}
\caption{\centering
C1: \_sp\_noise\_uint8(img, amount=float(np.random.uniform(0.004, 0.012)), ratio=float(np.random.uniform(0.45, 0.55)) , p=0.20)
}
\label{fig:example}
\end{figure}

\begin{figure}[H]
\centering
\includegraphics[width=\linewidth]{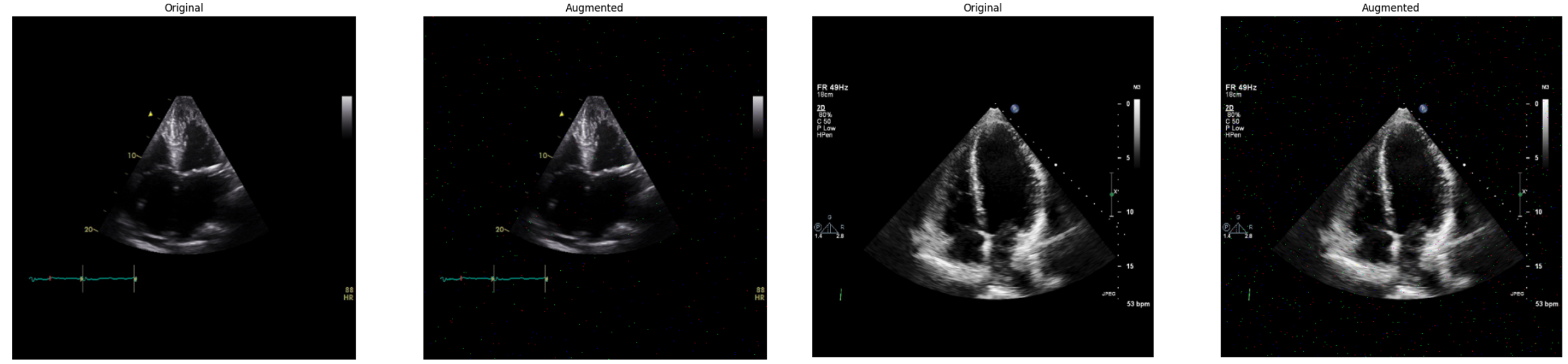}
\caption{\centering
C2: \_sp\_noise\_uint8(img, amount=float(np.random.uniform(0.001, 0.004)), ratio=float(np.random.uniform(0.48, 0.52)) , p=0.20)
}
\label{fig:example}
\end{figure}

\begin{figure}[H]
\centering
\includegraphics[width=\linewidth]{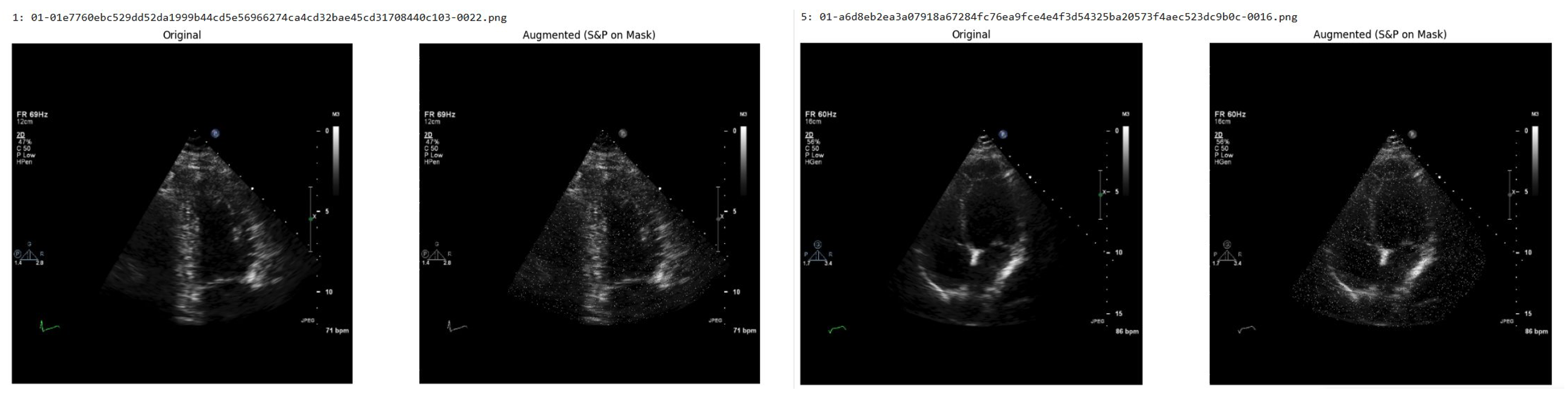}
\caption{
C3: SaltAndPepperOnMask(amount=(0.03, 0.06), ratio=(0.48, 0.52), p=1.0) {mask-based}
}
\label{fig:example}
\end{figure}

\subsection{DepthAttenuation}

\begin{figure}[H]
\centering
\includegraphics[width=\linewidth]{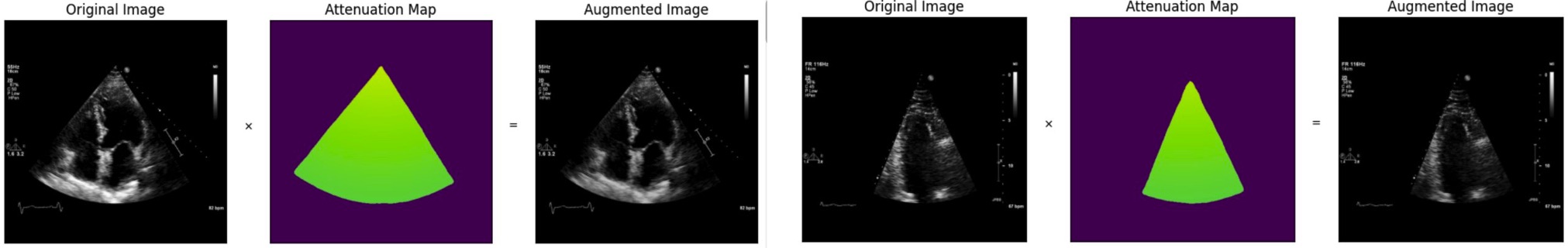}
\caption{
L: DepthAttenuation(attenuation\_rate=1.0,max\_attenuation=0.6, p=0.3)
}
\label{fig:example}
\end{figure}

\begin{figure}[H]
\centering
\includegraphics[width=\linewidth]{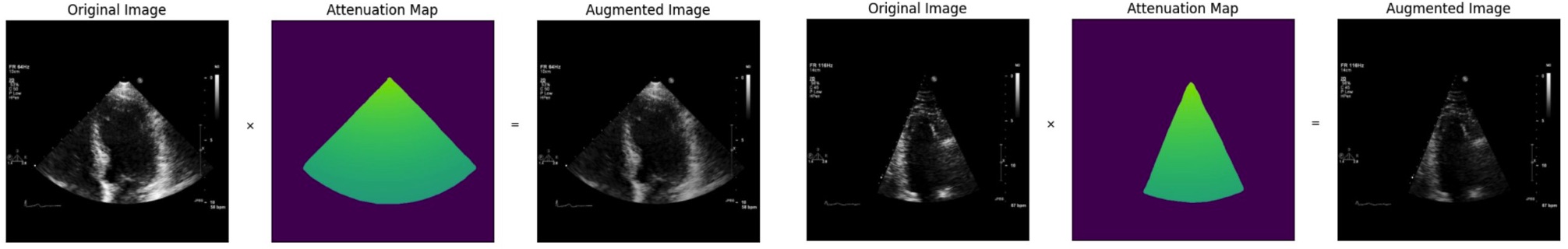}
\caption{
C1: DepthAttenuation(attenuation\_rate=(0.25, 0.9),max\_attenuation=0.08, p=0.2)
}
\label{fig:example}
\end{figure}

\begin{figure}[H]
\centering
\includegraphics[width=\linewidth]{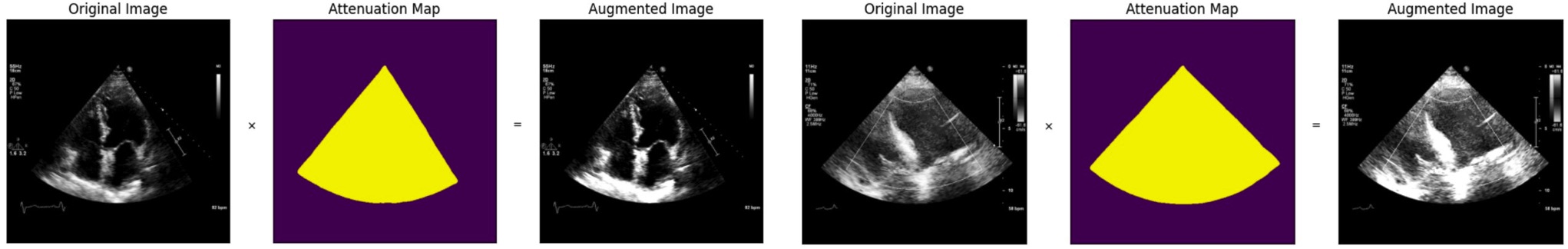}
\caption{
C2: DepthAttenuation(attenuation\_rate=1,max\_attenuation=2, p=0.3)
}
\label{fig:example}
\end{figure}

\subsection{GaussianShadow}

\begin{figure}[H]
\centering
\includegraphics[width=\linewidth]{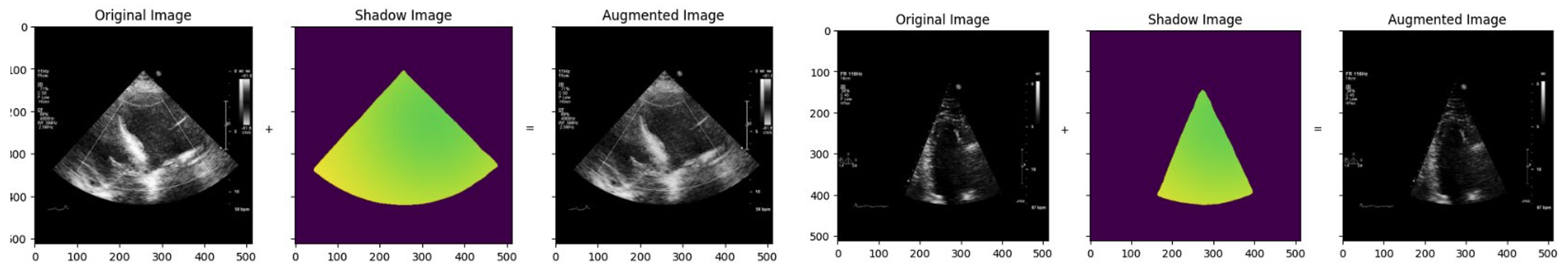}
\caption{
100 L: GaussianShadow(strength=0.2, sigma\_x=0.3,sigma\_y=0.3, p=0.3)
}
\label{fig:example}
\end{figure}

\begin{figure}[H]
\centering
\includegraphics[width=\linewidth]{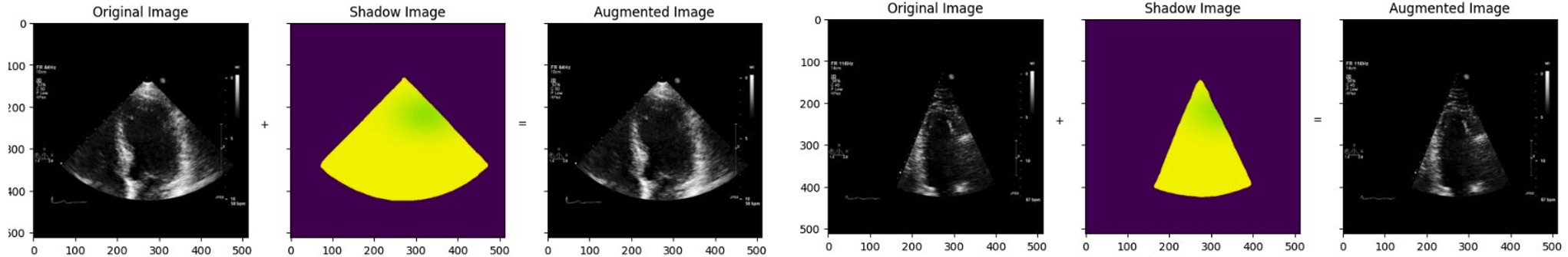}
\caption{
C1: GaussianShadow(strength=(0.12, 0.28), sigma\_x=(0.05, 0.12), sigma\_y=(0.05, 0.12), p=0.1)
}
\label{fig:example}
\end{figure}

\subsection{HazeArtifact}

\begin{figure}[H]
\centering
\includegraphics[width=\linewidth]{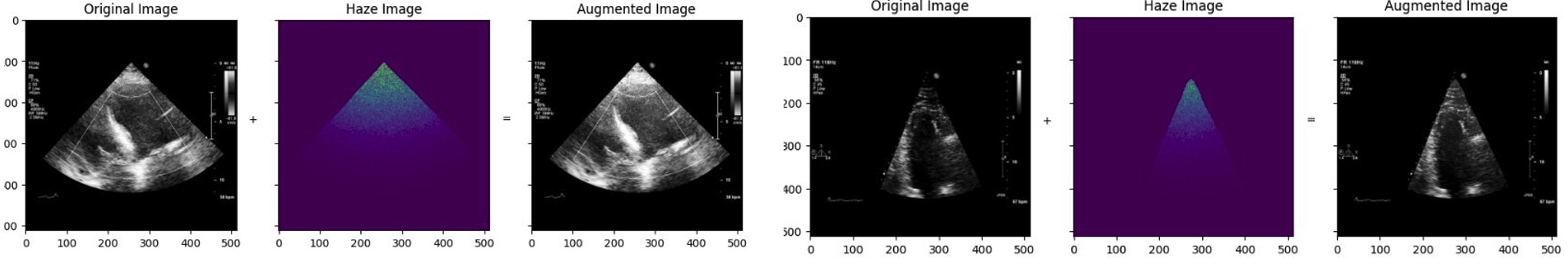}
\caption{
L: HazeArtifact(radius=0.2, sigma=0.2,p=0.2)
}
\label{fig:example}
\end{figure}

\begin{figure}[H]
\centering
\includegraphics[width=\linewidth]{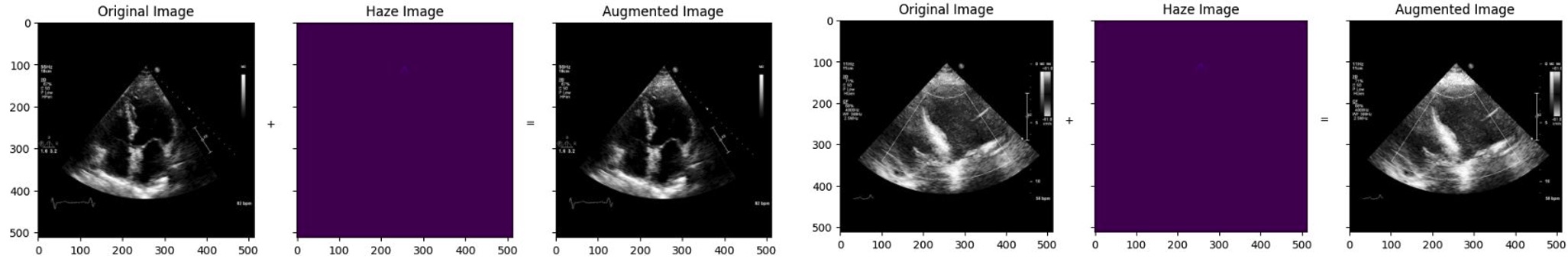}
\caption{
C1: HazeArtifact(radius=(0.15, 0.45), sigma=(0.03, 0.06), p=0.15)
}
\label{fig:example}
\end{figure}

\subsection{SpeckleReduction}

\begin{figure}[H]
\centering
\includegraphics[width=\linewidth]{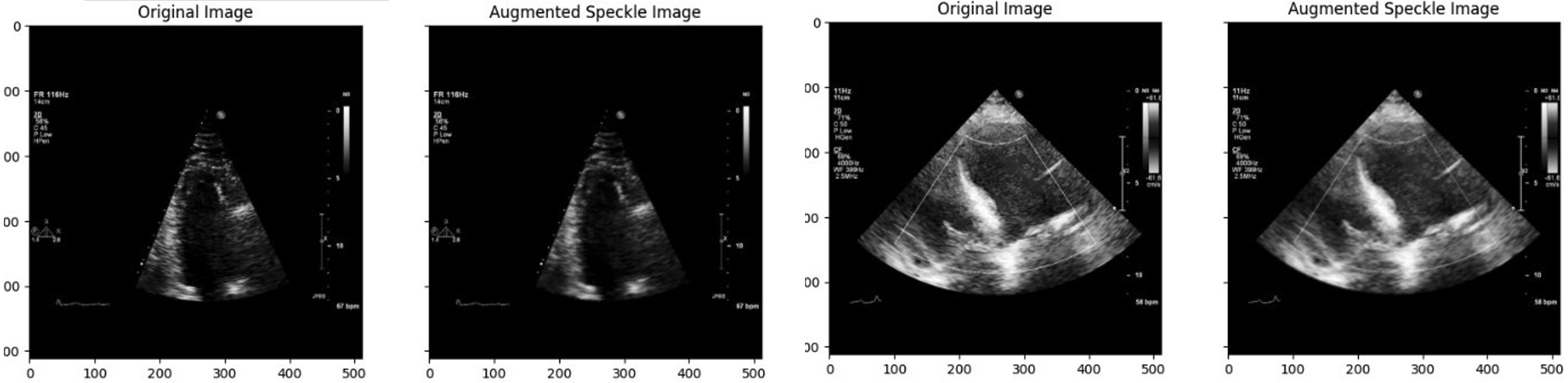}
\caption{
C1:SpeckleReduction(sigma\_spatial=0.2, sigma\_color=0.2, window\_size=5, p=0.3)
}
\label{fig:example}
\end{figure}

\begin{figure}[H]
\centering
\includegraphics[width=\linewidth]{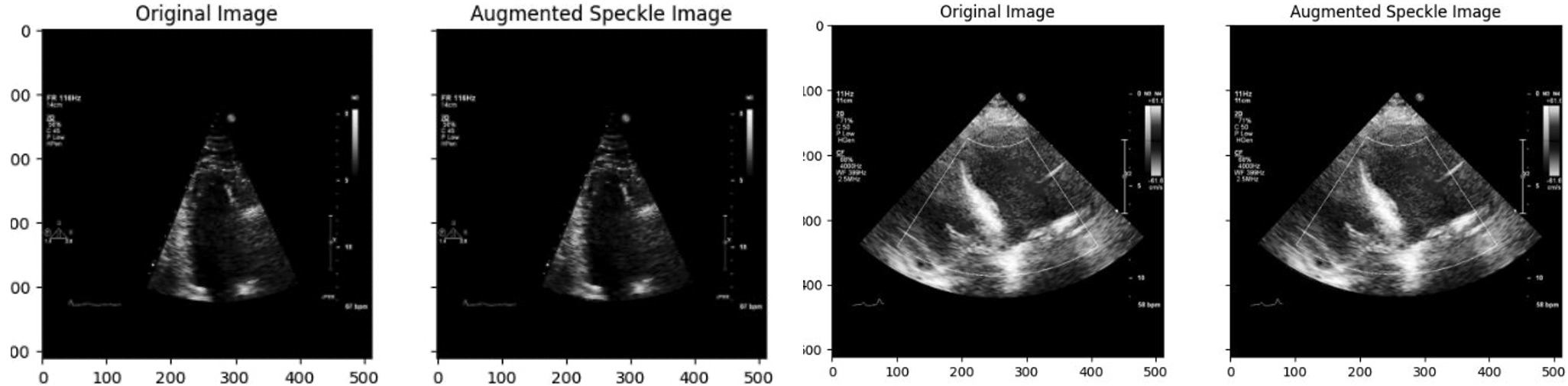}
\caption{
C1:SpeckleReduction(sigma\_spatial=0.2, sigma\_color=0.2, window\_size=5, p=0.3)
}
\label{fig:example}
\end{figure}

\section{Ablation Study of Padding Before Resizing}

The padding ablation showed that padding before resizing improved absolute cross-dataset performance, increasing the NONE baseline row mean from 0.6658 to 0.7236 for Dice and from 0.5697 to 0.6173 for IoU. The improvement was somewhat more noticeable for CAMUS, where direct resizing without padding may introduce mild distortion when adapting images to a square 512×512 format, whereas padding better preserved the anatomical shape. Despite this improvement, the augmentation ranking remained stable, with horizontal flipping as the strongest individual augmentation and its combination with affine transformation as the best pairwise strategy, achieving a Dice score of approximately 0.83. These findings indicate that padding reduces preprocessing-induced geometric distortion without altering the relative effectiveness of the evaluated augmentations, thereby supporting the robustness of the main conclusions and its retention in the preprocessing pipeline, as shown in Figures 112 to 116.

\begin{figure}[H]
\centering
\includegraphics[width=\linewidth]{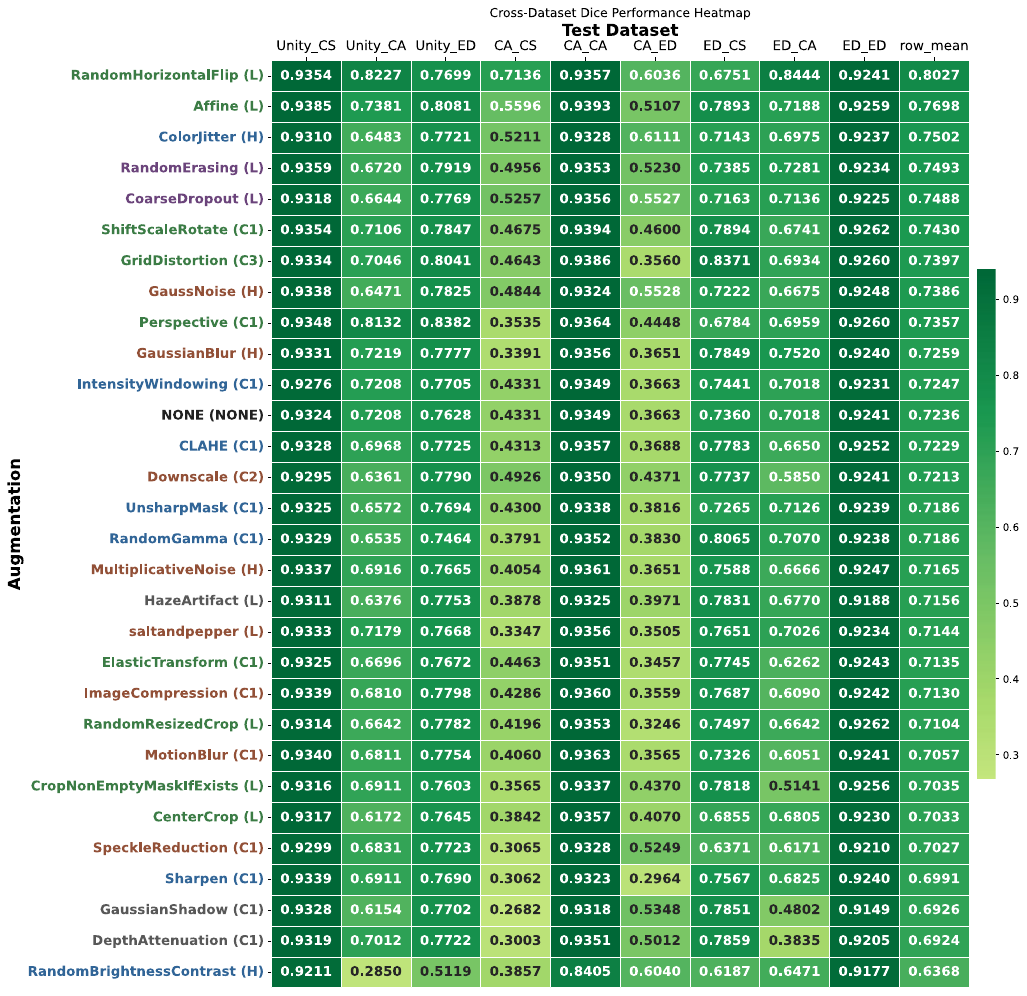}
\caption{
Dice performance heatmap for all augmentation strategies (with Padding)
}
\label{fig:example}
\end{figure}

\begin{figure}[H]
\centering
\includegraphics[width=0.8\linewidth]{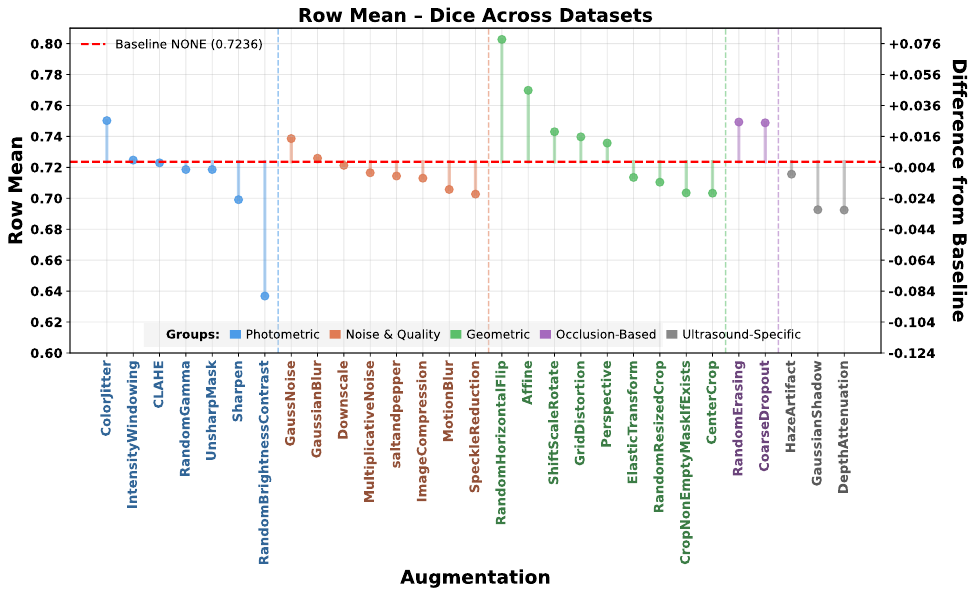
}
\caption{
Row-Mean Dice Performance Across Datasets and Relative Improvements Over the NONE (with Padding)
}
\label{fig:example}
\end{figure}

\begin{figure}[H]
\centering
\includegraphics[width=0.9\linewidth]{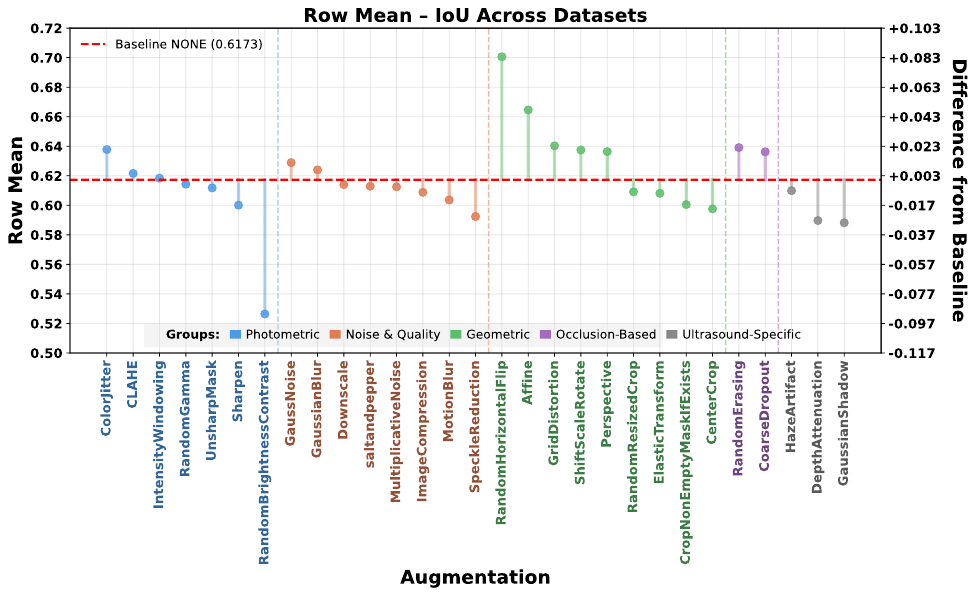}
\caption{
Row-Mean IoU Performance Across Datasets and Relative Improvements Over the NONE (with Padding)
}
\label{fig:example}
\end{figure}

\begin{figure}[H]
\centering
\includegraphics[width=0.62\linewidth]{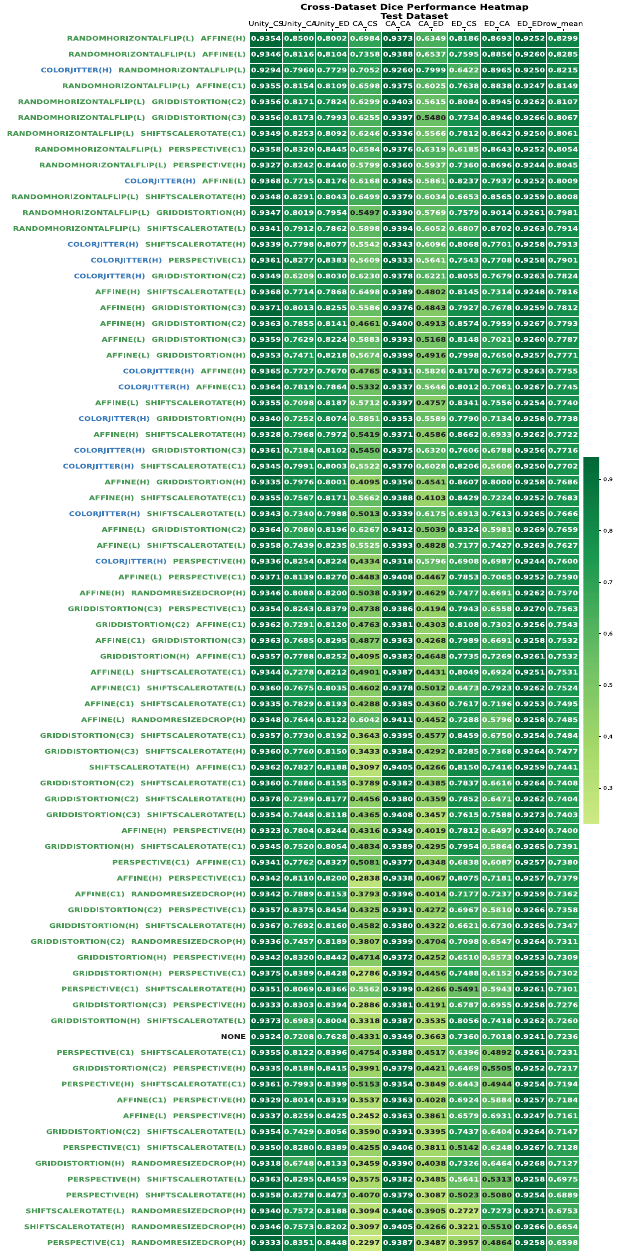}
\caption{
Cross-Dataset Dice Performance for Pairwise Augmentations (with Padding)
}
\label{fig:example}
\end{figure}

  \begin{figure}[htbp]
    \centering

    \begin{subfigure}[t]{0.80\textwidth}
        \centering
        \includegraphics[
            width=\textwidth,
            keepaspectratio
        ]{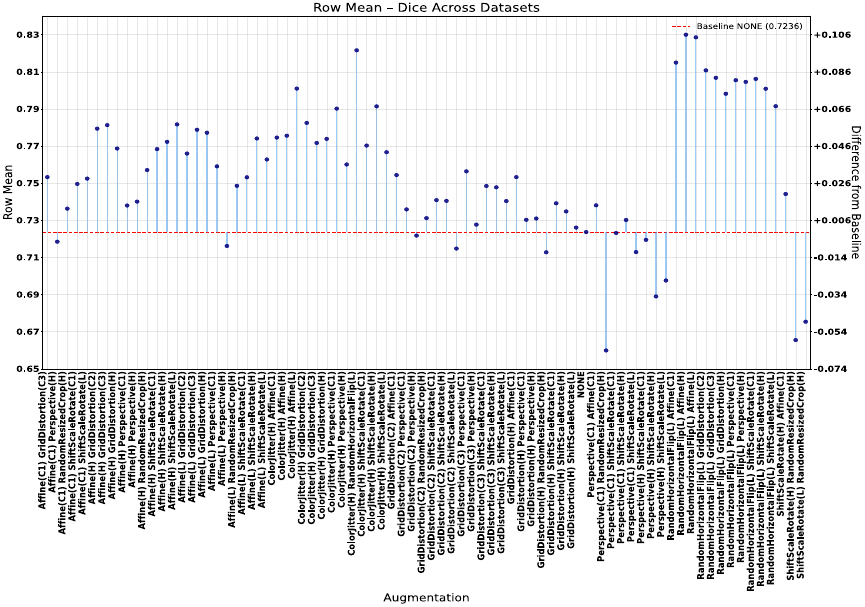}
        \caption{Row-Mean Dice }
        \label{fig:left_plot}
    \end{subfigure}

    \vspace{0.5cm}

    \begin{subfigure}[t]{0.80\textwidth}
        \centering
        \includegraphics[
            width=\textwidth,
            keepaspectratio
        ]{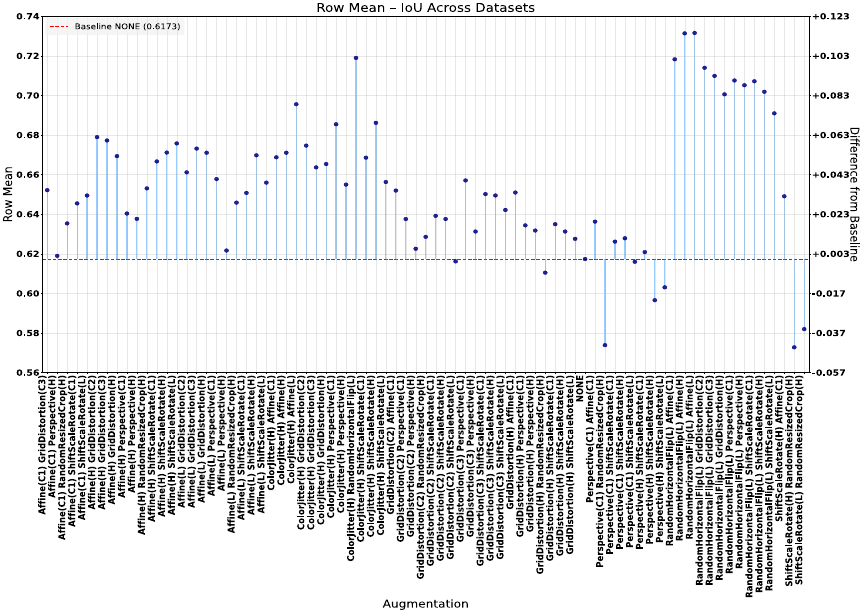}
        \caption{Row-Mean IoU}
        \label{fig:right_plot}
    \end{subfigure}

    \caption{
         performance across datasets and relative
        improvements over the NONE baseline for pairwise augmentations
        with padding.
    }
    \label{fig:comparison}
\end{figure}

\end{document}